%% file: ms.tex
\documentclass{article}

\pdfoutput=1

\usepackage{PRIMEarxiv}
\usepackage[utf8]{inputenc} 
\usepackage[T1]{fontenc}    
\usepackage{hyperref}       
\usepackage{url}            
\usepackage{booktabs}       
\usepackage{amsfonts}       
\usepackage{nicefrac}       
\usepackage{microtype}      
\usepackage{lipsum}
\usepackage{fancyhdr}       
\usepackage{graphicx}       

\usepackage{afterpage}

\title{Race Bias Analysis of Bona Fide Errors \\ in face anti-spoofing}

\author{Latifah Abduh \& Ioannis Ivrissimtzis\\ 
Department of Computer Science, Durham University, UK\\ 
{\tt\small e-mail:\{latifah.a.abduh, ioannis.ivrissimtzis\}@durham.ac.uk}
}

\begin{document}
\maketitle

\begin{abstract}
The study of bias in Machine Learning is receiving a lot of attention in recent years, however, few only papers deal explicitly with the problem of race bias in face anti-spoofing. In this paper, we present a systematic study of race bias in face anti-spoofing with three key characteristics: the focus is on analysing potential bias in the bona fide errors, where significant ethical and legal issues lie; the analysis is not restricted to the final binary outcomes of the classifier, but also covers the classifier's scalar responses and its latent space; the threshold determining the operating point of the classifier is considered a variable. 
We demonstrate the proposed bias analysis process on a VQ-VAE based face anti-spoofing algorithm, trained on the Replay Attack and the Spoof in the Wild (SiW) databases, and analysed for bias on the SiW and Racial Faces in the Wild (RFW), databases. The results demonstrate that race bias is not necessarily the result of different mean response values among the various populations. Instead, it can be better understood as the combined effect of several possible characteristics of the response distributions: different means; different variances; bimodal behaviour; existence of outliers. 
\end{abstract}



\input{sec1.tex}
\input{sec2.tex}

\input{sec3.tex}

\input{sec4.tex}

\input{sec5.tex}

\input{sec6.tex}

\bibliographystyle{unsrt}
\bibliography{ref}

\end{document}

%% file: sec1.tex
\section{Introduction}
\label{sec:introduction}

Face recognition is the method of choice behind some of the most widely deployed biometric authentication systems, currently supporting a range of applications, from passport control at airports, to mobile phone or laptop login. A key weaknesses of the technology, preventing it from being employed in security sensitive applications in uncontrolled environments, as for example ATM machines for money withdrawal, is its vulnerability to {\em presentation attacks}, where imposters attempt to gain wrongful access by presenting in front of the system’s camera a photo, or a video, or by wearing a mask resembling a registered person. As a solution to this problem, algorithms for presentation attack detection (PAD) are developed, that is, binary classifiers trained to distinguish between the bona fide samples coming from live subjects, and those coming from imposters.

The large variety in the types of possible presentation attacks, and the large variation in the environmental conditions under which they might take place, make PAD a particularly challenging problem. However, the current state-of-the-art, utilising the power of deep learning, comprises classifiers with excellent accuracy rates, and a satisfactory generalisation power to at least a limited number of previously unseen attacks. Cross-database generalisation is still problematic, however, it is debatable if this is a real obstacle to the deployment of PAD algorithms in practical applications, since such algorithms as usually embedded in specific face recognition systems, with given camera specifications and configurations.

Here, we deal with the problem of race bias in face anti-spoofing algorithms. It is a topic that has attracted considerably less research interest than accuracy and generalisation power, despite the fact that it raises ethical, legal, and regulatory considerations, which, by their own, can prevent adoption in specific applications. Addressing this gap, the aim of this paper is to provide a framework for studying the question: {\em Does the classifier work equally well on people from all races?}.

The proposed race bias analysis process has three key characteristics. First, the focus is on the bona fide error, that is, on genuine people wrongly classified as imposters. Bias in this type of error has significant ethical, legal and regulatory ramifications, and as it has recently point out ``creates customer annoyance and inconvenience, and this is also where bias can occur in PAD systems'', \cite{idrd2022}. Secondly, we analyse various stages of the classification process. Not just the final binary outcome, but also the scalar responses of the network prior to thresholding, and before that the representation of the face image in the network's latent space. Thirdly, we treat the value of the threshold that determines the classifier's operating point on the ROC curve as a user-defined variable. We do not assume it is fixed by the vendor of the biometric verification system through a black-box process.

In the rest of the paper, we demonstrate the application of the proposed bias analysis approach on a face anti-spoofing algorithm based on the recently proposed Vector Quantized Variational Autoencoder (VQ-VAE) architecture, \cite{van2017neural}. The network is trained and validated on the Replay Attack and the SiW databases, and tested for racial bias on bona fide samples from the SiW and the RFW databases. Hypotheses are tested using the chi-squared test on the binary outcomes, the Mann–Whitney U test on the scalar responses, and Hartigan's Dip for testing bimodality in the response distributions. To test for bias in the latent space of the VQ-VAE network, we train an SVM with encoding vectors from two races, and measure its performance as a binary classifier. 

The contributions of the paper are summarised as follows: 
\begin{itemize}
    \item A demonstration that race bias can be attributed to several characteristics of the response distributions: different means; different variances; bimodality; outliers. 
    \item A demonstration that non-specialised databases, such as RFW, can be used to analyse face anti-spoofing algorithms. 
    \item A VQ-VAE based network for face anti-spoofing. 
\end{itemize}


The rest of the paper is organised as follows. In Section~\ref{sec:sec2}, we review the relevant literature. In Section~\ref{sec:sec3}, we describe the VQ-VAE face anti-spoofing algorithm and the databases we used. In Section~\ref{sec:sec4}, we present the bias analysis on the SiW database, and in Section~\ref{sec:sec5} the bias analysis on the RFW database. We briefly conclude in Section~\ref{sec:sec6}. 

%% file: sec2.tex
\section{Background}
\label{sec:sec2}

We briefly review the area of face anti-spoofing, and then studies of bias in machine learning, and PAD in particular.

\subsection{Face anti-spoofing}

The earlier machine learning approaches to PAD were based on handcrafted features \cite{chingovska2012,boulkenafet2015face}, with Histogram of Oriented Gradient (HOG) \cite{albiol2008face}, and Local Binary Patterns (LBP) \cite{chingovska2012}, among the most popular.



More recent approaches were based on CNNs, \cite{jourabloo2018face,nagpal2019performance}, or combinations of various deep network types \cite{cai2020drl}, leading to the current state of the art being based on various forms of deep learning \cite{zhang2020celeba,yu2020fas,Yu_2020_CVPR,yu2020auto,zhang2020face,cai2020drl,liu2021face,yu2021revisiting}, such as Central Difference Convolutional Networks (CDCN) \cite{Yu_2020_CVPR,yu2020fas}, or transformers \cite{wang2022}. Following some earlier approaches \cite{atoum2017face,liu2018}, the current state of the art algorithms also utilise depth information \cite{wang2020deep,zhang2020face,yu2021revisiting,wu2022}, which can be estimated by a independently trained neural network, while the use of GANs to estimate Near Infrared (NIR) information was proposed \cite{liu2021face}. 


The experiment presented in this paper is based on a one class trained autoencoder. Anomaly detection is a popular approach \cite{xiong2018unknown,jimenezdeep,Nikisins2019,Zhang2020CNNBasedAD,baweja2020anomaly}, offering good generalisation to unseen attacks. In \cite{mohammadi2020}, images from face recognition datasets were added to the two-class training set of an autoencoder, and improved cross-database generalisation was reported. A similar behaviour was reported in \cite{abduh2021training} when images from the in-the-wild were added to the training set. 

\subsection{Databases}

In this paper, the first training set is from {\em Replay-Attack} \cite{chingovska2012}, a database consisting of 50 subjects of three types of ethnicities, 76\% Caucasian, 22\% Asian, and 2\% African. Our second training set is from {\em SiW} \cite{liu2018}, a database consisting of 165 subjects, of four types of ethnicities, 35\% of Asian and 35\% Caucasian and 23\% Indian, and 7\% African American. The bias analysis is performed on SiW with the subject annotated for ethnicity type by us, and the already annotated RFW database \cite{wang2019rfw}. 

Regarding other databases, {\em NUAA} \cite{tan2010face} was one of the first face anti-spoofing large databases. It is rarely used these days as its low quality of imagery poses an unfair challenge in the cross-database validation of algorithms. {\em MSU MFSD} \cite{7031384} consists of 55 subjects, captured by four different devices, while {\em OULU} \cite{OULU_NPU_2017} has again 55 subjects captured by six different mobile devices. {\em WMCA} \cite{george2019biometric} contains 72 subjects and information is captured in RGB, depth, infrared, and thermal. {\em CASIA-SURF}  \cite{zhang2020casia} consists of 1000 subjects captured in RGB, depth, and infrared. 

The first face anti-spoofing database to include explicit ethnic labels was {\em CASIA-SURF CeFA} \cite{Liu_2021_WACV}, which has 1,607 in three ethnicities, captured in three modalities. In this paper, for bias analysis we use the RFW \cite{wang2019rfw}, which includes four types of ethnicities, Caucasian, Asian, Indian, and African. RFW does not specialise in face anti-spoofing, and it is more widely used in the bias analysis literature.


\subsection{Bias in machine learning} 

In \cite{yapo2018ethical}, several high profile cases of machine learning bias are documented; Google search results appeared to be biased towards women in 2015; Hewlett-Packard’s software for web cameras struggled to recognize dark skin tones; and Nikon’s camera software was inaccurately identifying Asian people as blinking. 

Thus, given also the ethical, legal, and regulatory issues associated with the problem of bias within human populations, there is a considerable amount of research on the subject, especially in face recognition (FR). A recent comprehensive survey can be found in \cite{mehrabi2021survey}, where the significant sources of bias \cite{suresh2019framework,olteanu2019social} are categorised and discussed, and the negative effect of bias on downstream learning tasks is pointed out. We also note that while the current deep learning based FR algorithms are under intense scrutiny for potential bias \cite{singh2020robustness}, this is due to their wider deployment in real life applications, rather than any evidence that they are more biased than traditional approaches.  

In one of the earliest studies of bias in FR, predating deep learning, \cite{phillips2011} reported differences in the performance on humans of Caucasian and East Asian decent between Western and East Asia developed algorithms. In \cite{garcia2019harms}, several deep learning based FR algorithms are analysed and a small amount of bias is detected in all of them. Then, the authors show how this bias can be exploited to enhance the power of malicious morphing attacks to FR based security systems. 

In \cite{gluge2020not}, the authors compute cluster validation measures on the clusters of the various demographics inside the whole population, aiming at measuring the algorithm's potential for bias. Their result is negative, and they argue for the need of more sophisticated clustering approaches. We note that in our paper, an investigation in the latent space of the potential for bias, by measuring the discriminative power of SVMs over the various ethnicities, returned a similarly negative result. In \cite{serna2021insidebias}, the aim is the detection of bias by analysing the activation ratios at the various layers of the network. Similarly to our work, their target application is the detection of race bias on a binary classification problem, gender classification in their case. Their result is positive in that they report a correlation between the measured activation ratios and bias in the final outcomes of the classifier. However, it is not clear if their method can be used to measure and assess the statistical significance of the expected bias. 

In Cavazos et al. \cite{cavazos2020accuracy}, similarly to our approach, most of the analysis assumes a one-sided error cost, in their case the false acceptance rate, and the decision thresholds are treated as user defined variables. However, the analytical tools they used, mostly visual inspection of ROC curves, do not allow for a deep study of the distributions of the similarity scores, while, here, we give a more detailed analysis of the distribution of the responses, which is the equivalent of the similarity scores. In Pereira and Marcel \cite{pereira2021}, a fairness metric is proposed, which can be optimised over the decision thresholds, but again, there is no in-depth statistical analysis of the scores, as we do here for the responses, and thus they offer a more limited insight.


\subsubsection{Bias in Presentation Attack Detection}

The literature on bias in presentation attacks is more sparse. Race bias was the key theme in the competition of face anti-spoofing algorithm on the CASIA-SURF CeFA database \cite{liu2020crossethnicity}. Bias was assessed by the performance of the algorithm under a cross-ethnicity validation scenario. Standard performance metrics, such as APCER, BPCER and ACER we reported. In \cite{alshareef2021study}, the standard CNN models Resnet 50 and VGG16, were compared for gender bias against the debiasing-VAE proposed in \cite{amini2019uncovering}, and several performance metrics were reported. In a recent white paper by the ID R\&D company, which develops face anti-spoofing software, the results of a large scale bias assessment experiment conducted by Bixelab, a NIST accredited independent laboratory \cite{idrd2022}. Similarly to our approach, they focus on the bona fide errors, and their aim is the BPCER error metric to be below a prespecified threshold across all demographics. 

Regarding other biometric identification modalities, \cite{fang2021demographic} studied gender bias in iris PAD algorithms. They reported three error metrics, APCER, BPCER, and HTER, finding that female users would be less protected against iris PAD attacks.

%% file: sec3.tex
\section{Experimental setup}
\label{sec:sec3}

In this section, we briefly describe and validate the two classifiers that will be analysed for bias in the subsequent sections. They have the same VQ-VAE based architecture \cite{van2017neural}, and their only difference is in the training; in the first, the network is trained on Replay Attack, and in the second on SiW. We chose the VQ-VAE architecture because of some recently reported impressive results, on various computer vision problems, even though it has not been employed yet in the field of face anti-spoofing.

\subsection{The VQ-VAE network}

The VQ-VAE architecture comprises three parts. A learned {\em codebook}, which is used to discretise the continuous latent vectors to a set of discrete latent variables; each continuous vector is replaced with its nearest vector in the codebook. An {\em encoder} that maps the input to a sequence $z$ of discrete codes. A {\em decoder} that transforms the sequence $z$ of discrete vectors back to an image. The final binary classification decision is taken by comparing the MSE between the input image and the reconstructed one against a threshold determining the operating point of the algorithm. 

Our encoder consists of two convolutional layers with kernel size 4, stride step 2, padding 1, and followed by a ReLU, one convolutional layer with kernel size 3, stride step 1, padding 1 and followed by two residual blocks, which are implemented as ReLU, $3 \times 3$ conv, ReLU, $1 \times 1$ conv for each block. The decoder is a symmetrical structure of the encoder that uses transposed convolutions. The encoder may output a 16x16 grid of vectors, and each one is quantized, using a codebook of size 512, before being fed to the decoder. The weight factor $\beta$ was set to 0.25. 

The ADAM optimiser was used with a learning rate of 1e-3, and the model was trained for 100 epochs with a batch size of 16. The total loss consisted of three components: reconstruction loss, codebook loss, and commitment loss. The code was written entirely in Python, on the Pytorch platform. The experiments were run on an Intel Core i7 CPU, with 64 GB of RAM and an Nvidia GTX 1650.

\begin{figure*}[ht]
\includegraphics[width=\linewidth]{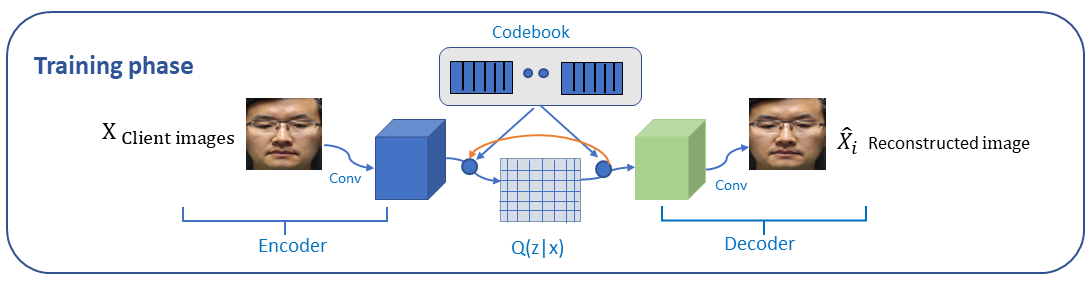}
\caption{The architecture of the proposed VQ-VAE.}
\label{fig:VQ-VAE-architecture}
\end{figure*}

\begin{figure*}[ht]
\includegraphics[width=0.24\textwidth]{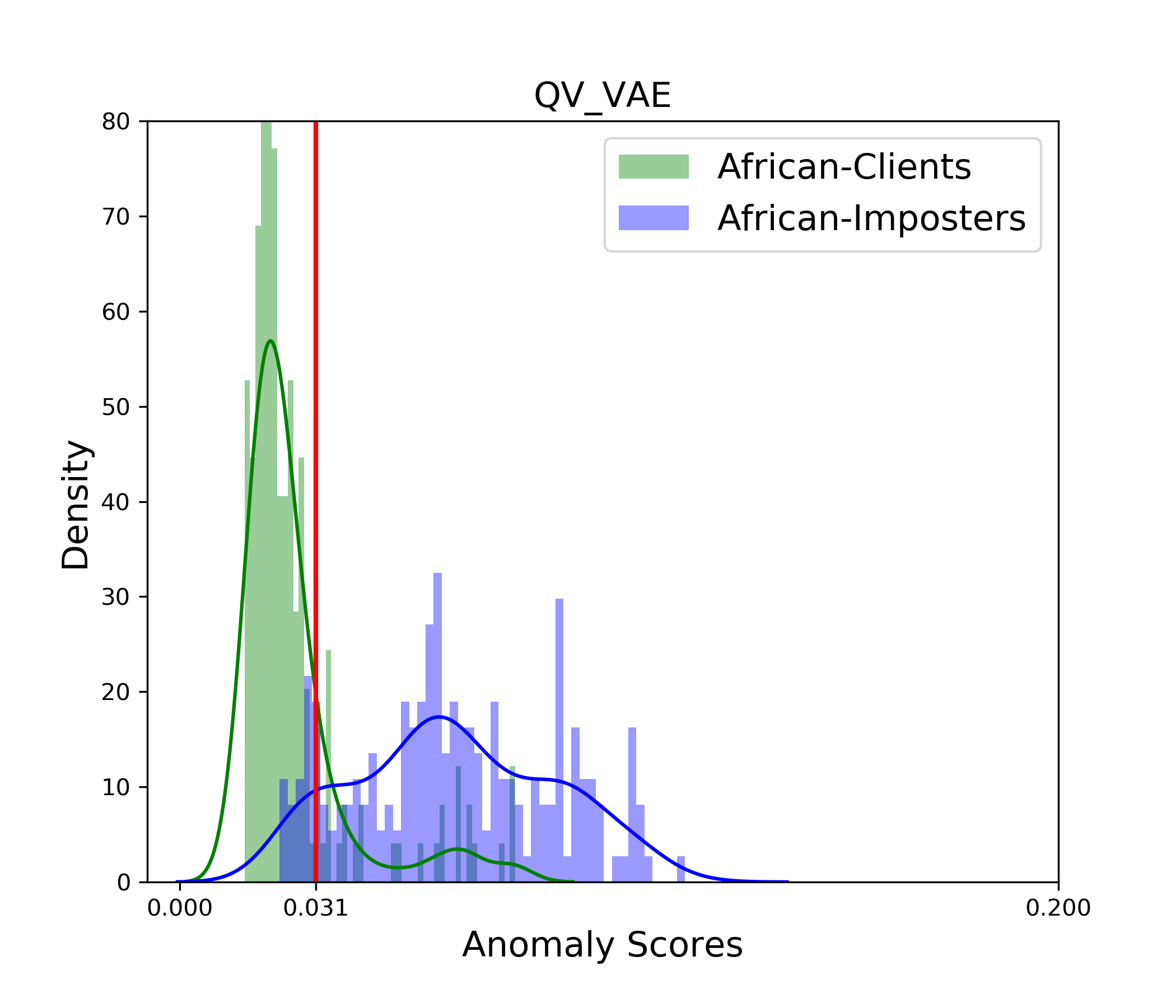} \hfill
\includegraphics[width=0.24\textwidth]{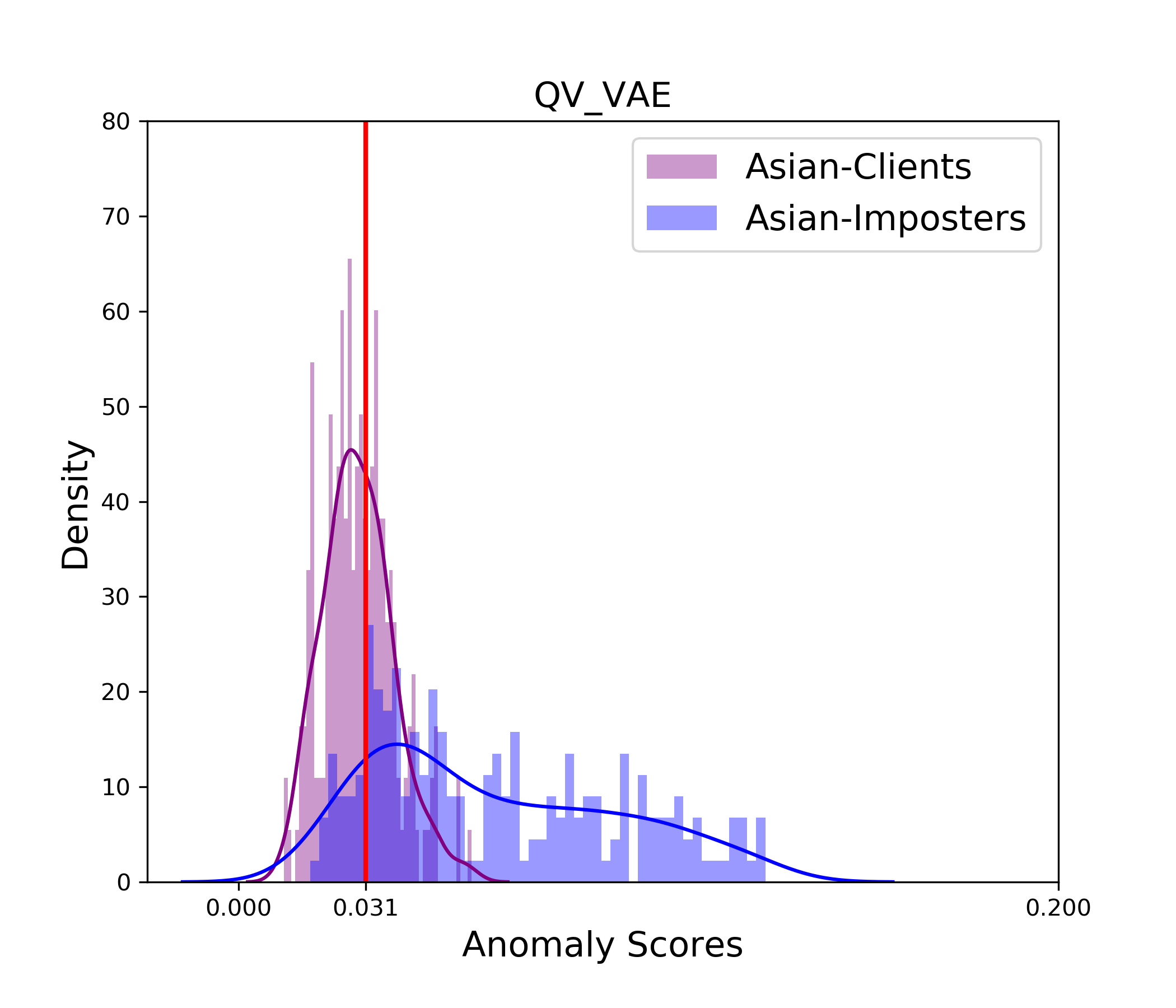} \hfill
\includegraphics[width=0.24\textwidth]{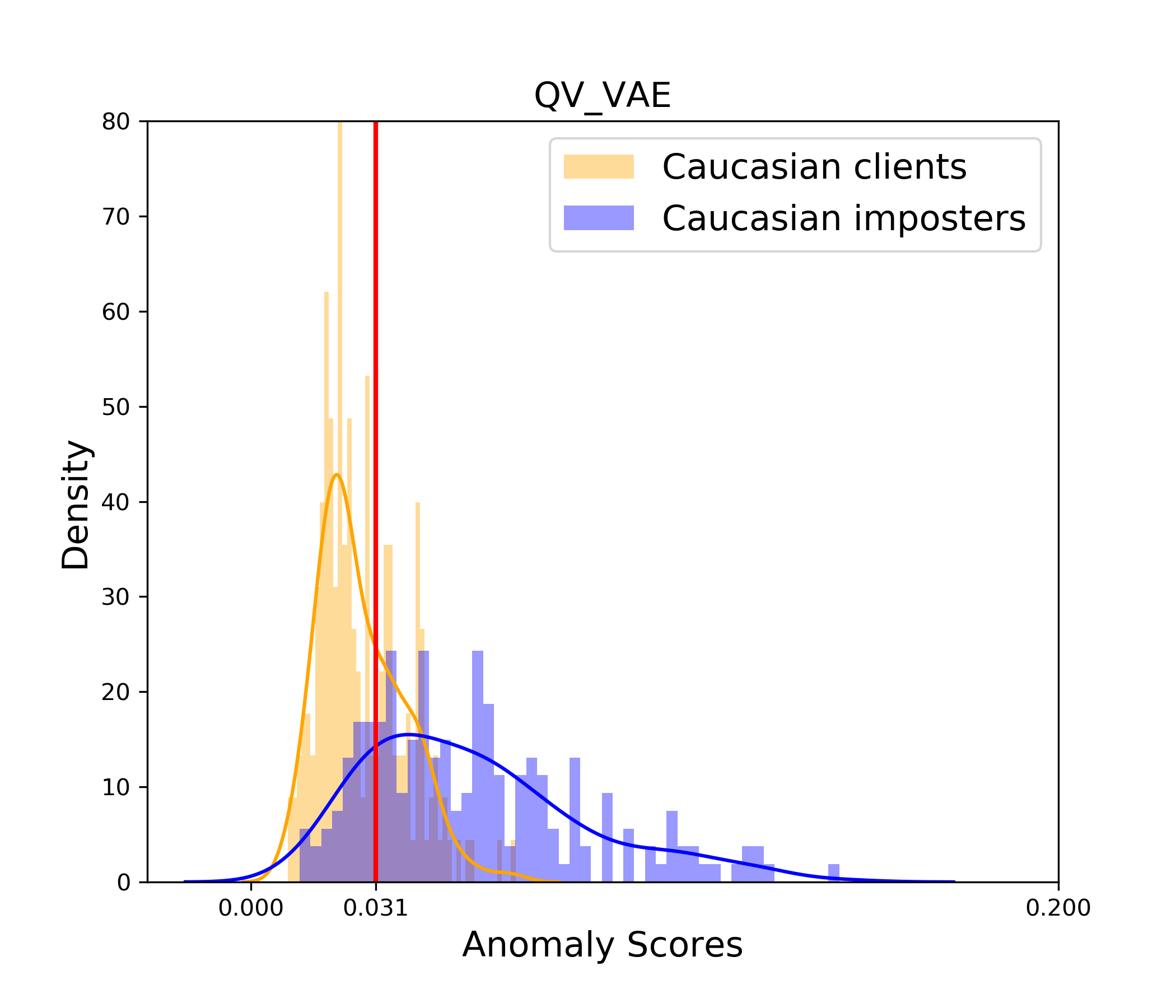} \hfill
\includegraphics[width=0.24\textwidth]{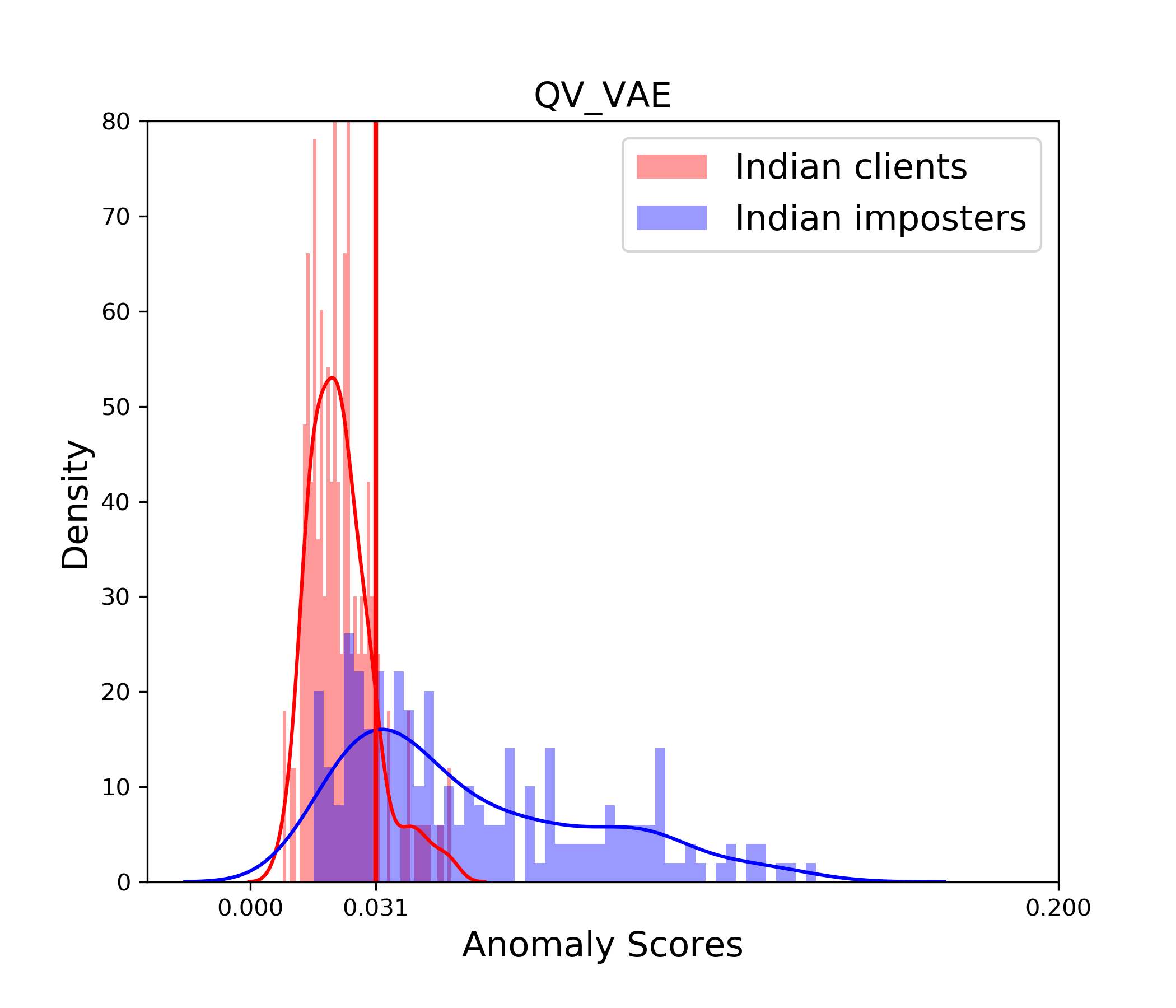}
\includegraphics[width=0.24\textwidth]{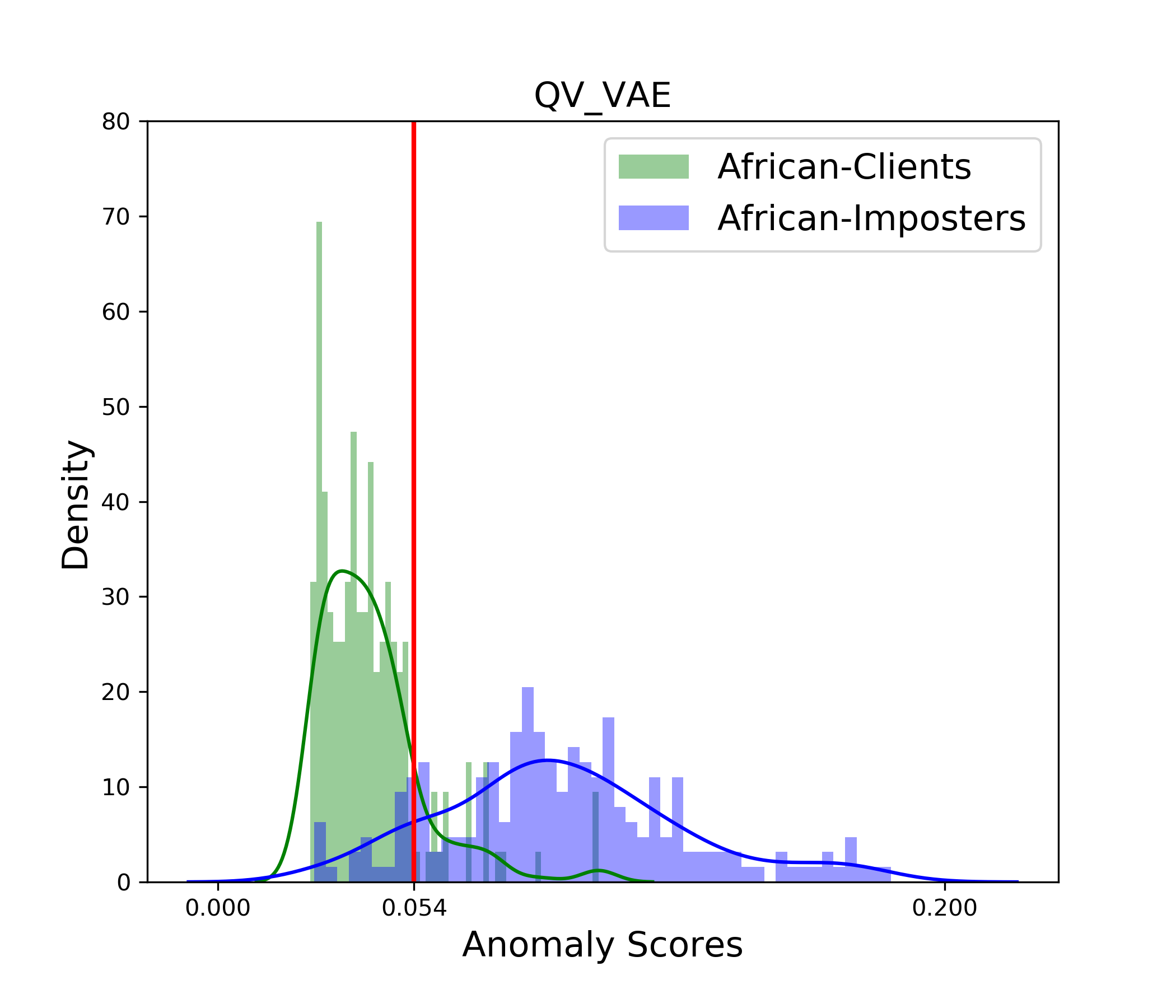} \hfill
\includegraphics[width=0.24\textwidth]{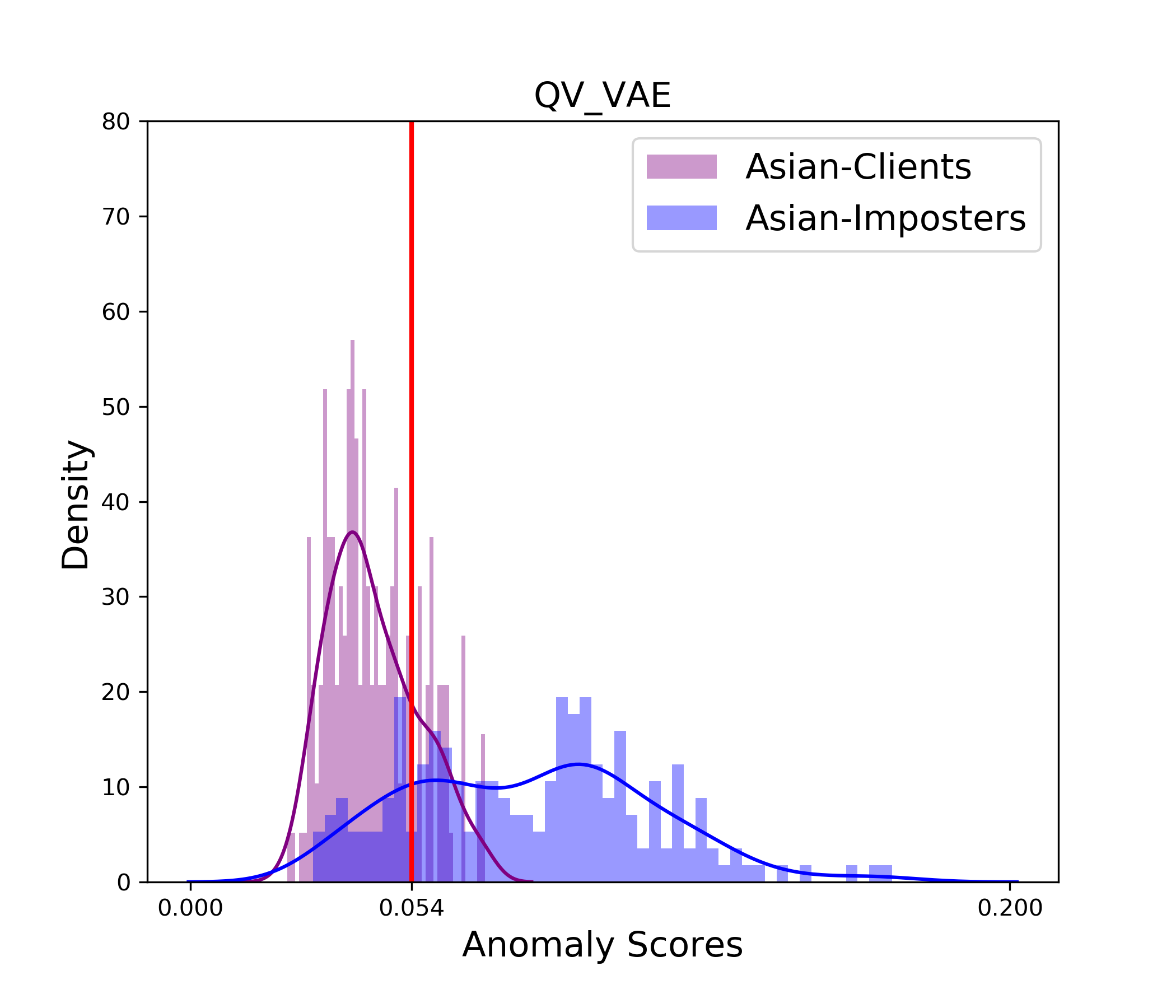} \hfill
\includegraphics[width=0.24\textwidth]{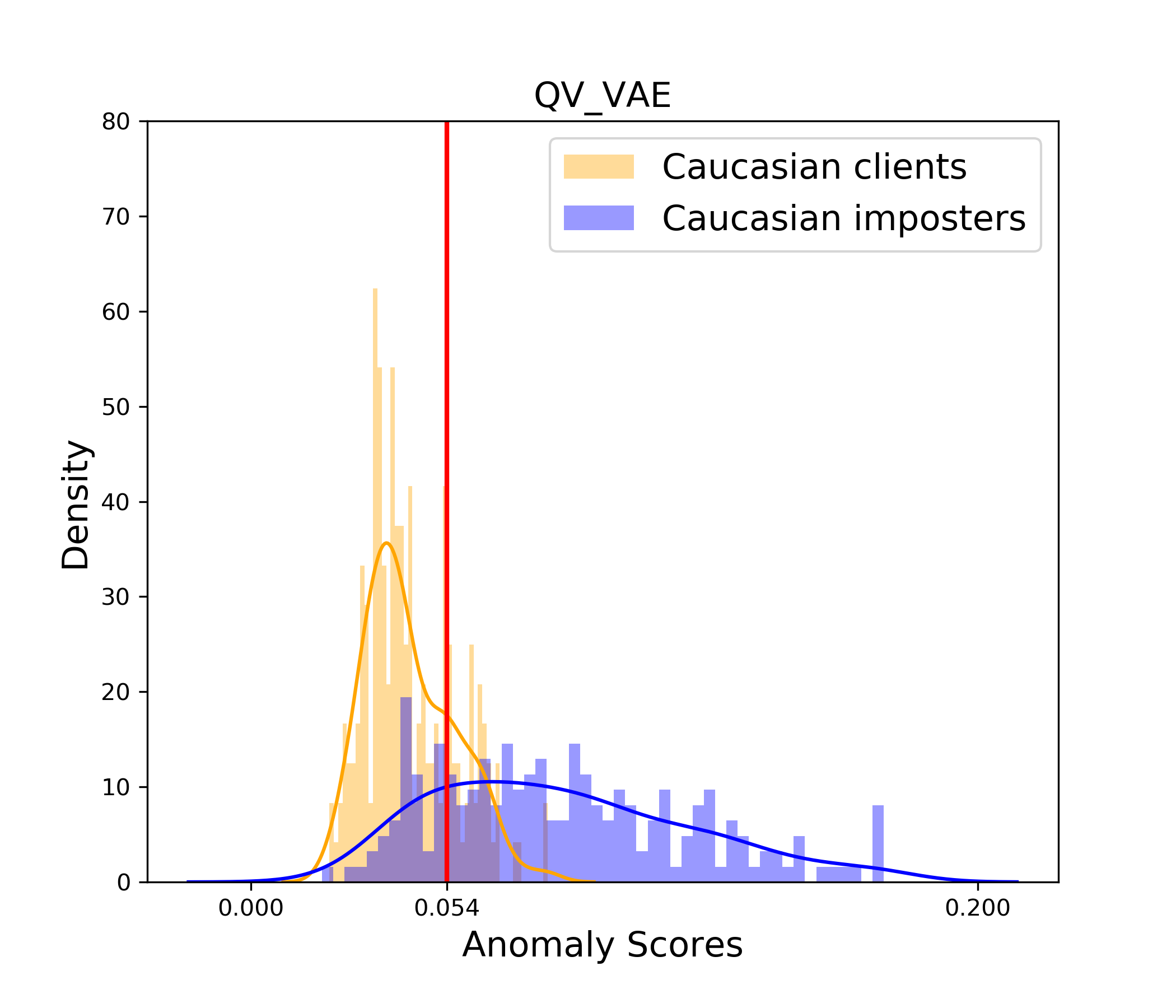} \hfill
\includegraphics[width=0.24\textwidth]{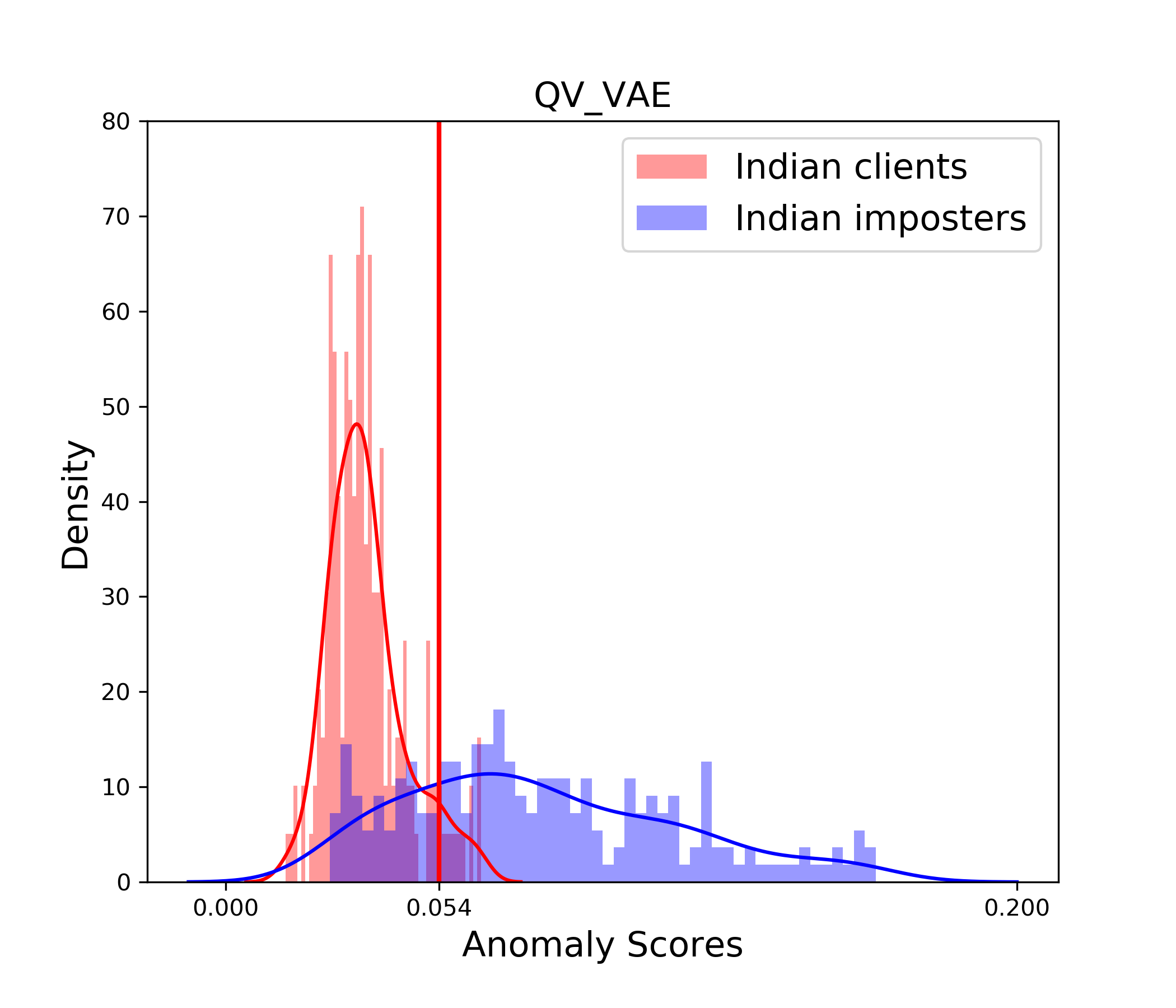}
\caption{{\bf Top:} The responses of the RA trained network on each race of the SiW testset. From left to right: African, Asian, Caucasian, Indian. {\bf Bottom:} The responses of the SiW trained network.}
\label{fig:siw-clientImposter}
\end{figure*}


\subsection{Data preparation} 

For faces detection we used the Multi-Task Cascade Convolutional Neural Network (MTCNN) \cite{zhang2016joint}. The detected faces were rotationally aligned, making the line between the two eye centres horizontal, and cropped at $64 \times 64$ resolution. Table~\ref{tbl:dataSets} summarises the sizes of the datasets. Notice that the training sets consisted of bona fide data, only.


\begin{table}[h]
\caption{Sizes of training and test datasets. The test sets are equally split between clients and imposters. The SiW test set, consists of 400 samples from each race.}
\centering
\begin{tabular}{|l|c|l|c|} \hline 
& \textbf{Training} & \textbf{Test} \\ \hline 
RA & 7465 & 400\\ \hline 
SiW & 124,000 & 1600\\ \hline
\end{tabular}
\label{tbl:dataSets}
\end{table}



\subsection{Network validation}

We validated the classifiers on RA and SiW, first under an intra-database protocol, and then with a cross-database protocol. When testing on SiW, we also report error rates for each race separately. 
\begin{itemize}
\item[] {\bf Intra-database protocol:} on an independent validation set we compute the threshold corresponding to the Equal Error Rate (EER). We use this threshold to compute the Half-Total Error Rate (HTER) on the test set. On SiW, we use a single threshold for all races. 
\item[] {\bf Cross-database protocol:} here, thresholds estimated on one database do not work on the other. So, we just report the EERs computed directly on the test sets. 
\end{itemize}

\noindent Table~\ref{tbl:validation-all} shows the error computed under the testing protocol. The intra-database error is significantly smaller in RA than in SiW (.055 vs .169), indicating an easier classification task, as RA has less diversity in poses and expressions. However, as expected, the error on SiW is smaller when we train on SiW rather than RA (.169 vs .208). 

\begin{table}[h]
\caption{Error rates computed under the testing protocol.}
\centering
\begin{tabular}{|c||l||l|} \hline 
 & RA test & SiW test \\ \hline \hline
RAtr & .055 [thr = .015 (HTER)] & .208 [thr = .031 (EER)]   \\ \hline 
SiWtr & .180  [thr = .065  (EER)] & .169 [thr = .054 (HTER)] \\ \hline 
\end{tabular}
\label{tbl:validation-all}
\end{table}

\noindent Table~\ref{tbl:cross-perRace} shows HTERs for each race of the SiW testset. The thresholds are taken from the corresponding tests of Table~\ref{tbl:validation-all}. We note that the SiW trained network is performing better on the SiW testset, not only in total, but also on each race separately. For each classifier and each race, Figure~\ref{fig:siw-clientImposter} shows the histograms of the responses and the corresponding thresholds. 

\begin{table}[h]
\caption{HTERs computed for each race separately on the SiW testset.}
\centering
\begin{tabular}{|c|c|c|c|c|} \hline 
 & \textbf{Af} & \textbf{As} & \textbf{Ca} & \textbf{In} \\ \hline \hline
HTER RAtr (thr = .031) & .135 & .247 & .260 & .192 \\ \hline 
HTER SiWtr (thr = .054) & .115 & .202 & .210 & .150 \\ \hline 
\end{tabular}
\label{tbl:cross-perRace}
\end{table}

%% file: sec4.tex
\section{Bias analysis on SiW}
\label{sec:sec4}

\begin{figure*}[h]
\includegraphics[width=\textwidth]{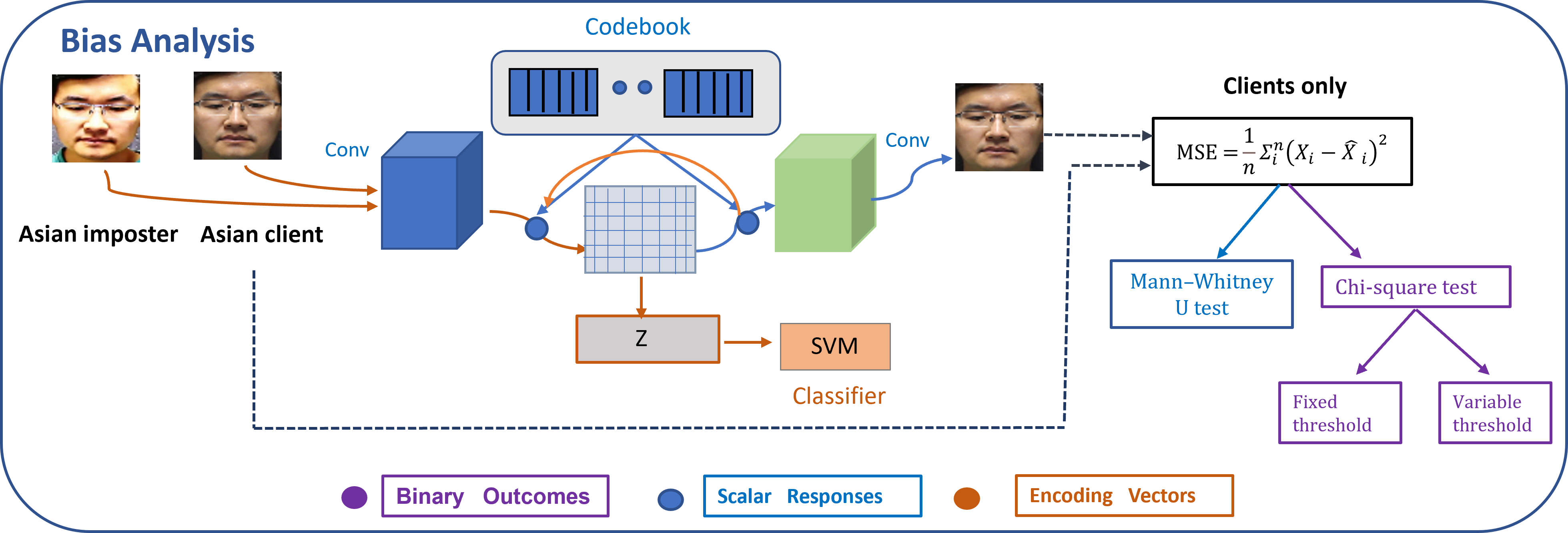}
\caption{The bias analysis process. The binary outcome analysis is shown in purple, the scalar responses analysis in blue, and the latent space analysis in orange.}
\label{fig:Bias_analysis}
\end{figure*}

Here, we use the SiW testset of Section~\ref{sec:sec3}. The binary outcomes of the classifiers are analysed with the chi-squared test, the scalar responses with the Mann–Whitney U test \cite{10.1214/aoms/1177730491}, and finally, the encoding vectors are analysed by using them to train and test an SVM on the task of race classification. The bias analysis process is summarised in Fig.~\ref{fig:Bias_analysis}. 

\subsection{Statistical analysis of the binary outcomes}

First, we analyse the binary outcomes corresponding to the thresholds of Section~\ref{sec:sec3}, that is, 0.031 for the RA-trained classifier and 0.054 for the SiW trained classifier. For each pair of races, we formed the $2\times 2$ contingency tables, and applied the chi-squared test, computing p-values for the hypothesis that the probability of misclassification of a bona fide sample from the race with the most misclassifications is higher. 

The results are summarised in Table~\ref{tbl:siw-binaryTest}. In several cases, the p-values are low, meaning that for any reasonable threshold of statistical significance the bias hypothesis is accepted. In other cases, p-values above 0.05 mean that bias has not been detected. 

\begin{table}[h]
\caption{p-values of the chi-squared tests for the thresholds in Section~\ref{sec:sec3}.}
\centering
\begin{tabular}{|c||c|c|c|c|c|c|} \hline 
          &  Af-As & Af-Ca & Af-In & As-Ca & As-In & Ca-In \\ \hline\hline
RAtr  &  .0000 & .0001 & .1465 & .2532& .0000 & .0000\\ \hline 
SiWtr & .1158 & .0104 & .0147 & .0000 & .0000 & 1.0 \\ \hline 
\end{tabular}
\label{tbl:siw-binaryTest}
\end{table}

\begin{figure*}[h]
\includegraphics[width=0.33\textwidth]{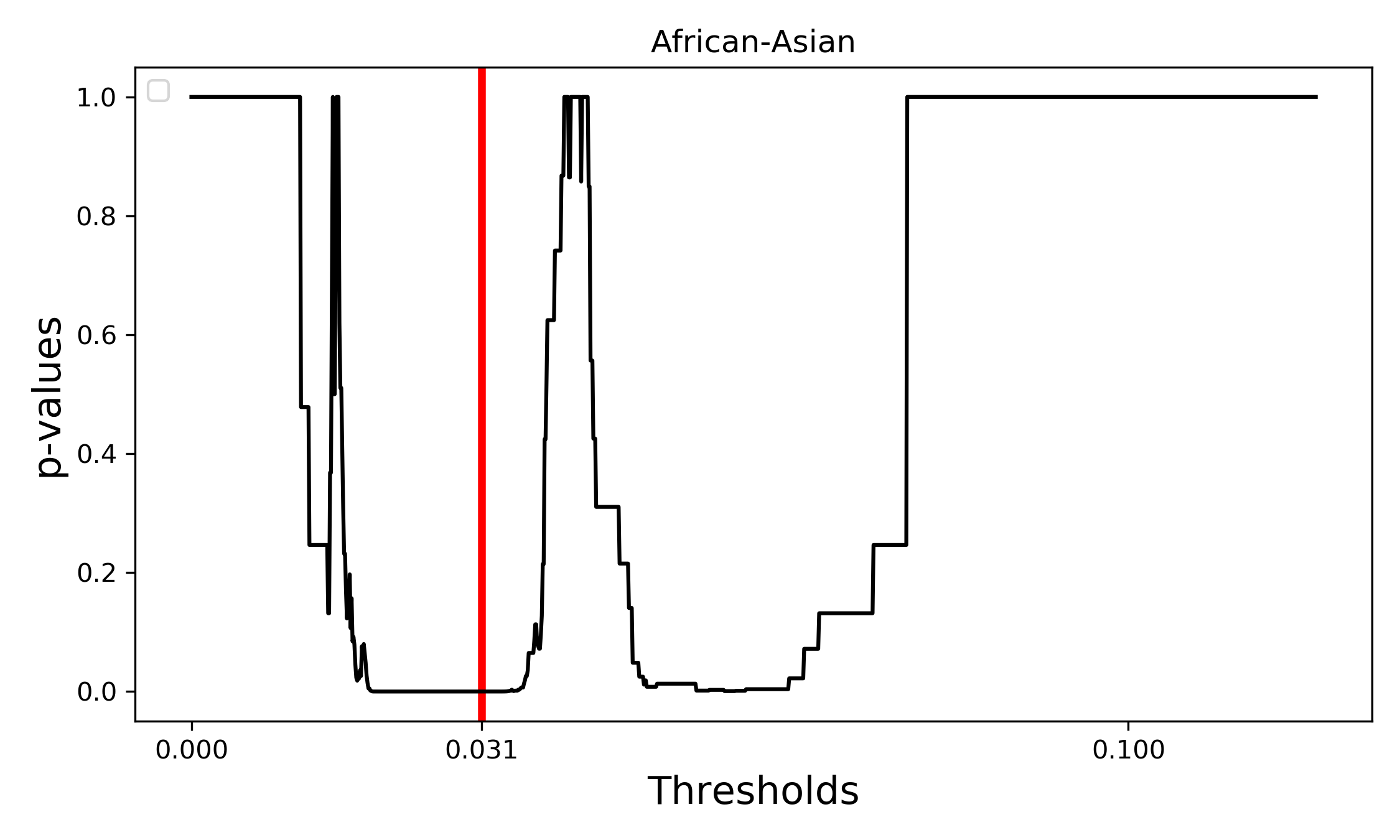} \hfill
\includegraphics[width=0.33\textwidth]{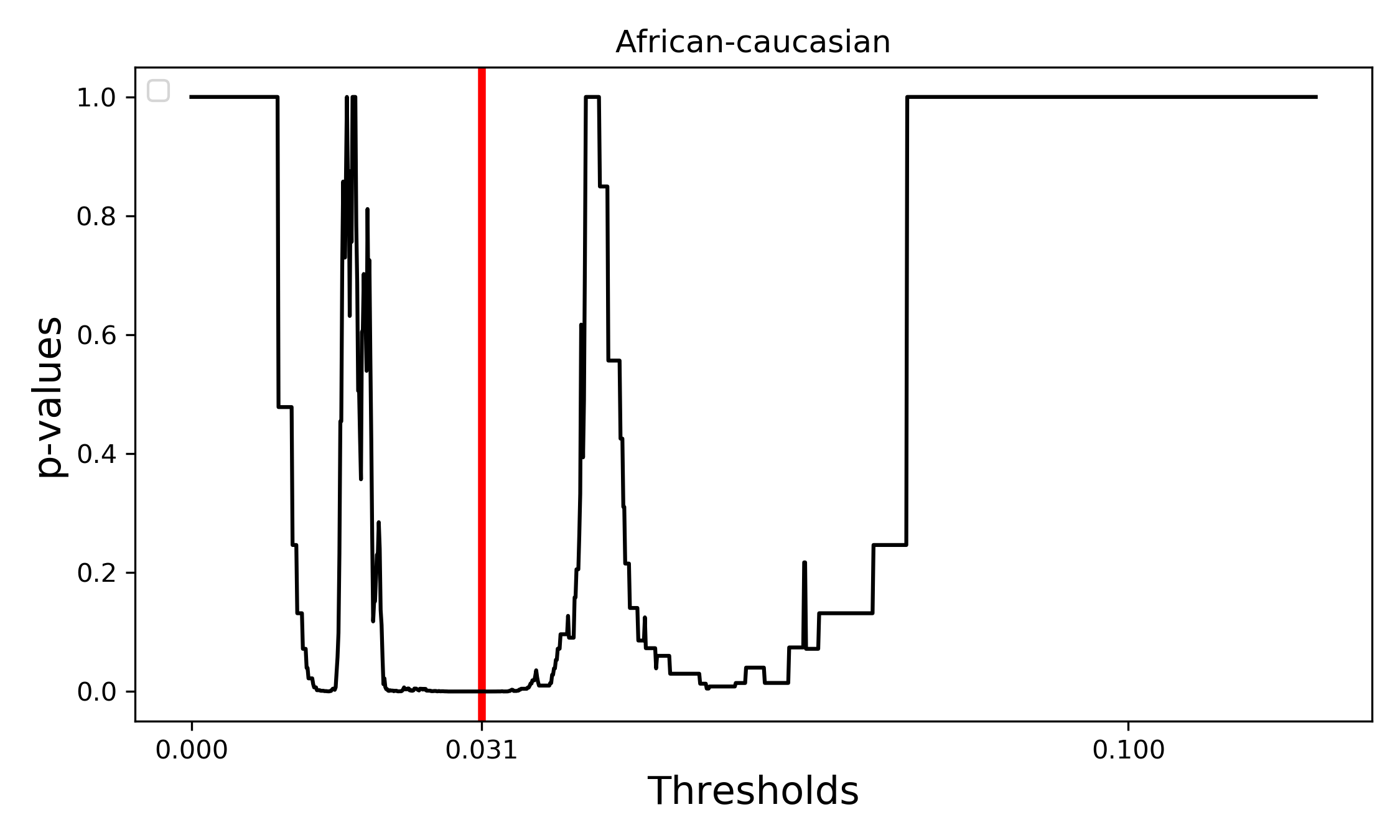} \hfill 
\includegraphics[width=0.33\textwidth]{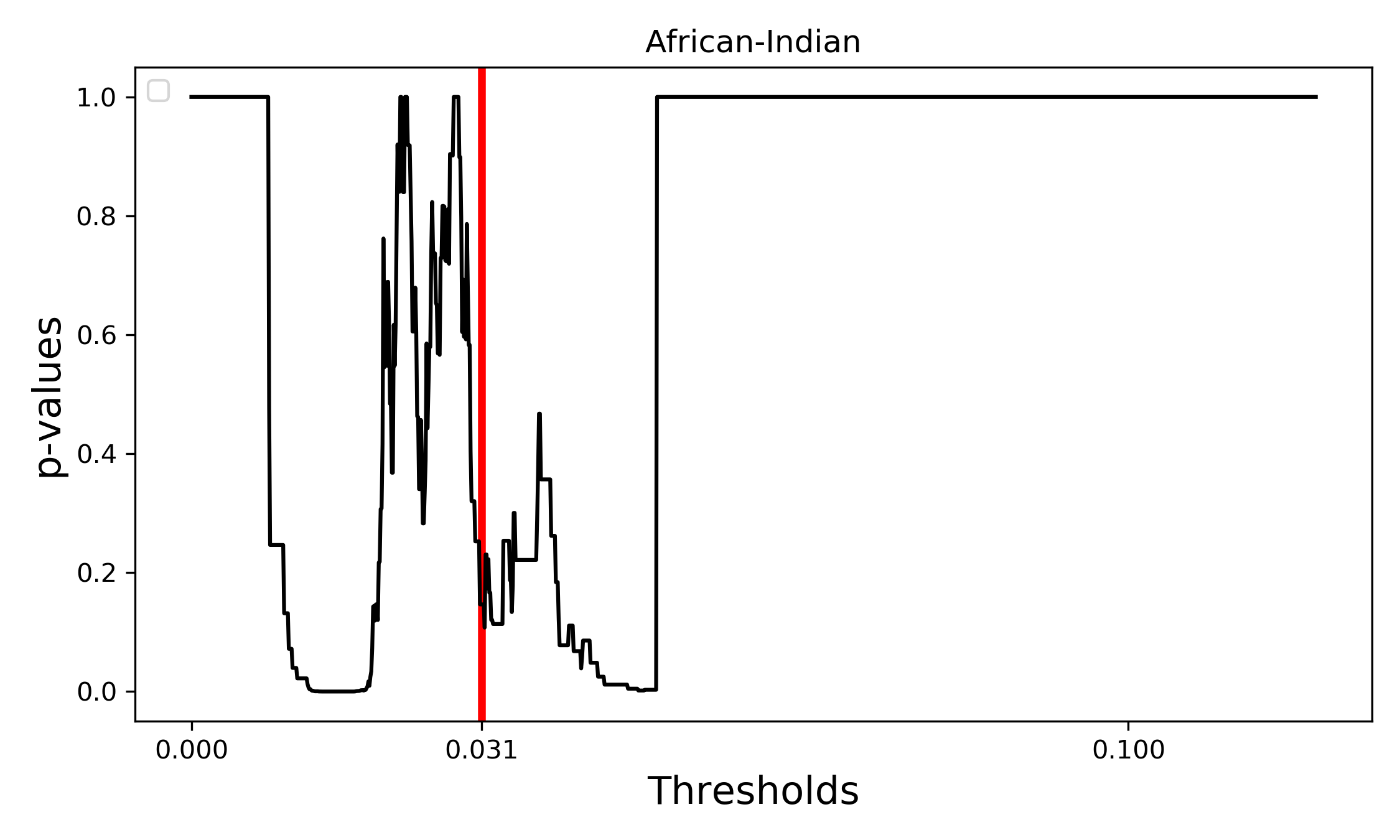}
\includegraphics[width=0.33\textwidth]{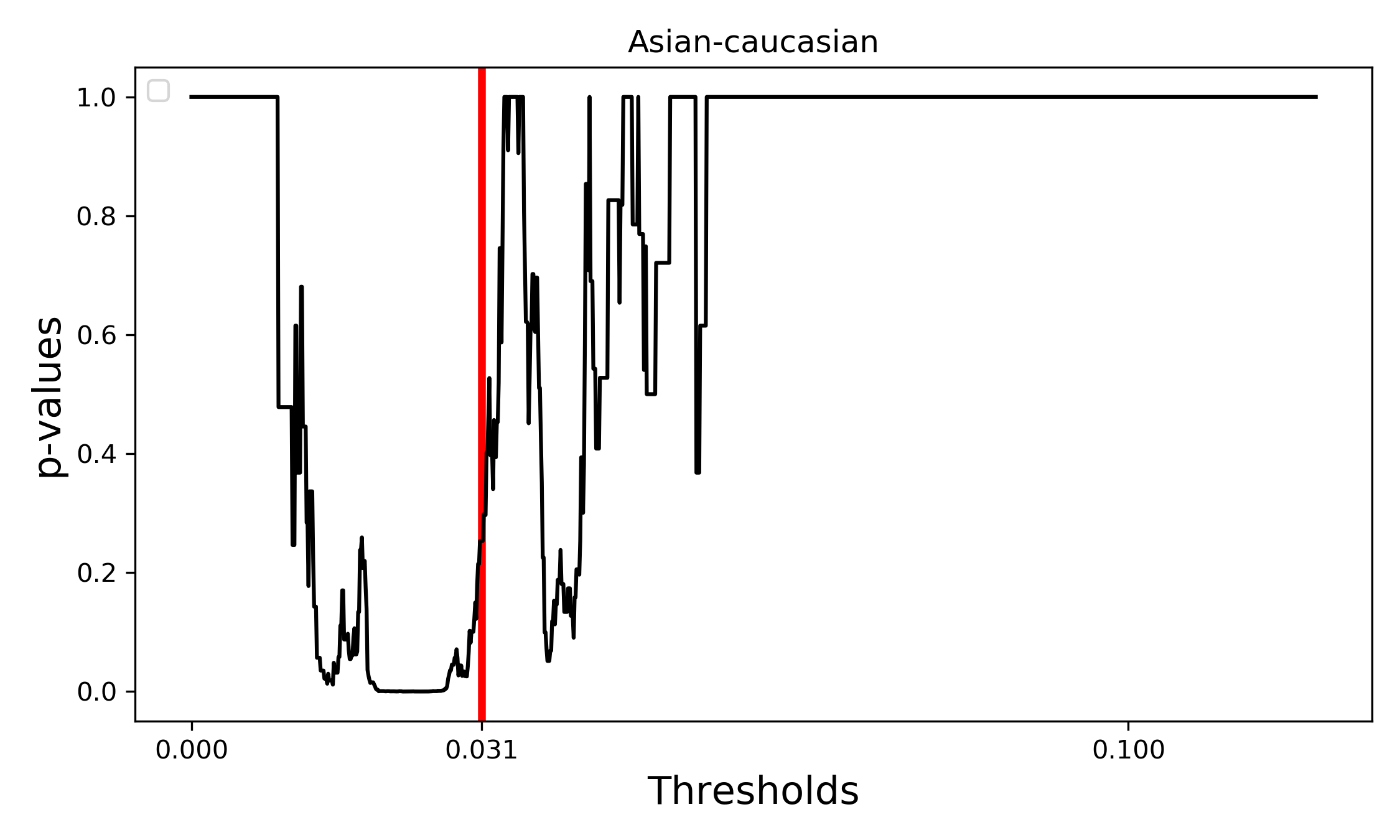} \hfill 
\includegraphics[width=0.33\textwidth]{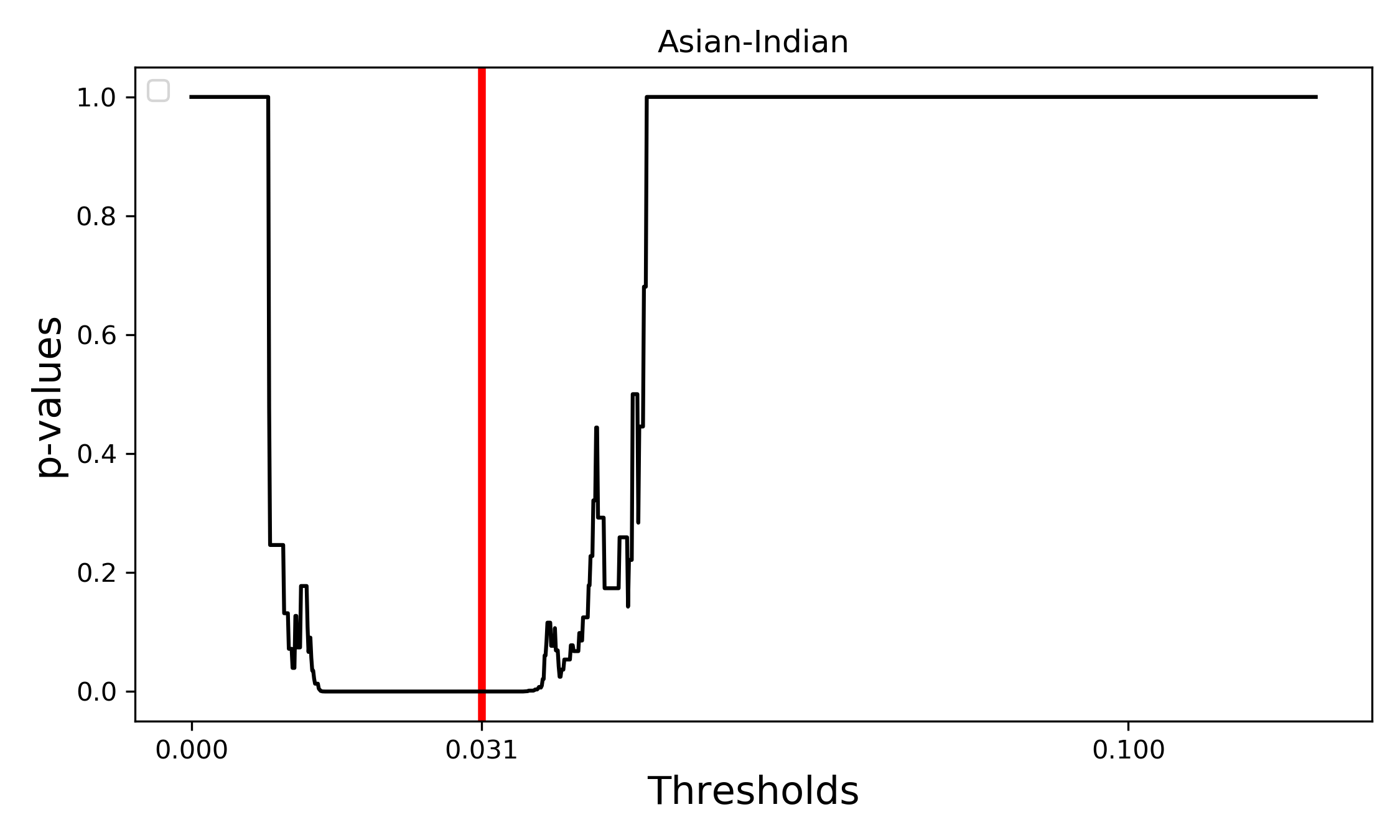} \hfill 
\includegraphics[width=0.33\textwidth]{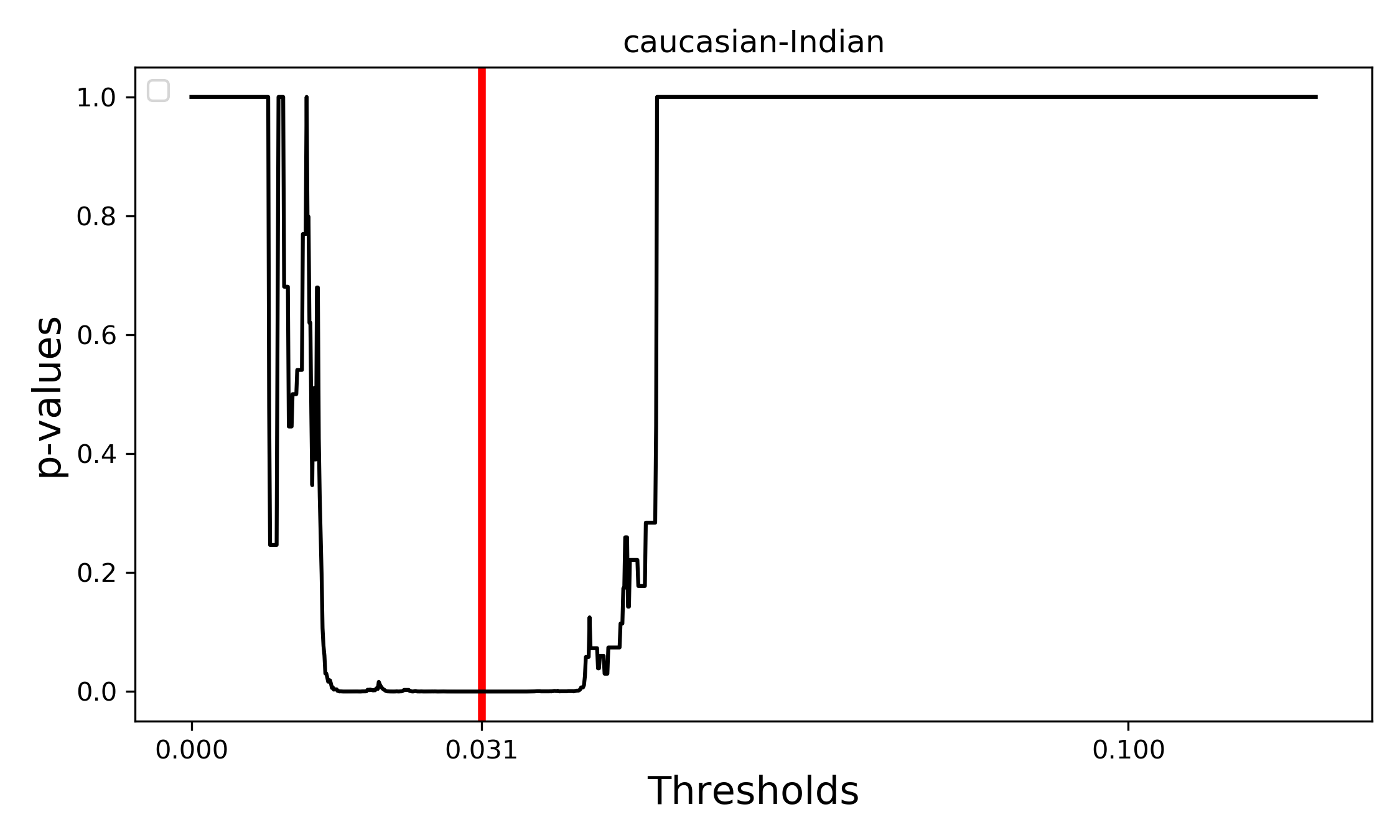}
\caption{For each pair of races, graphs of the p-value as a function of the threshold. The classifier was trained on RA.}
\label{fig:SiWp-values-RA}
\end{figure*}

\begin{figure*}[ht]
\includegraphics[width=0.33\textwidth]{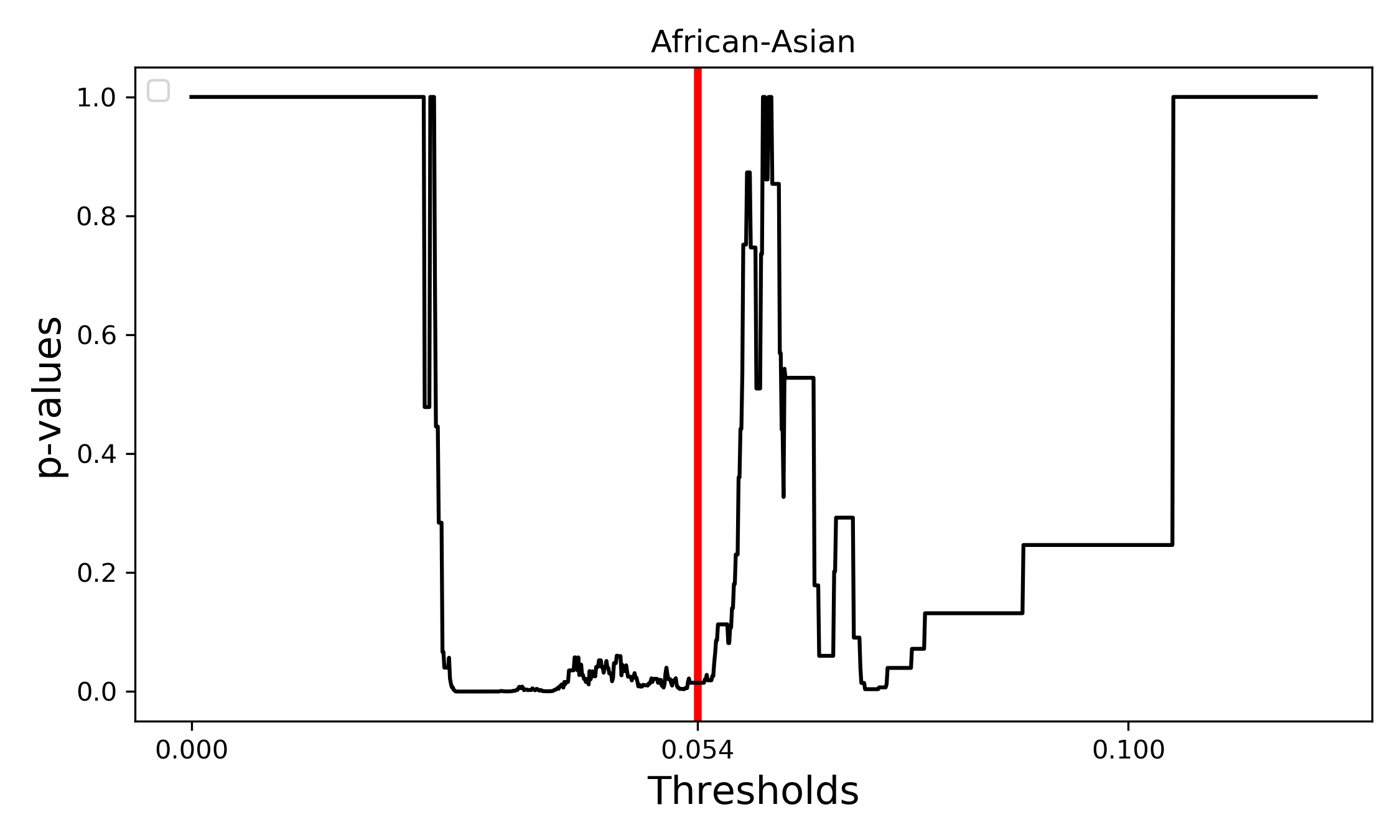} \hfill
\includegraphics[width=0.33\textwidth]{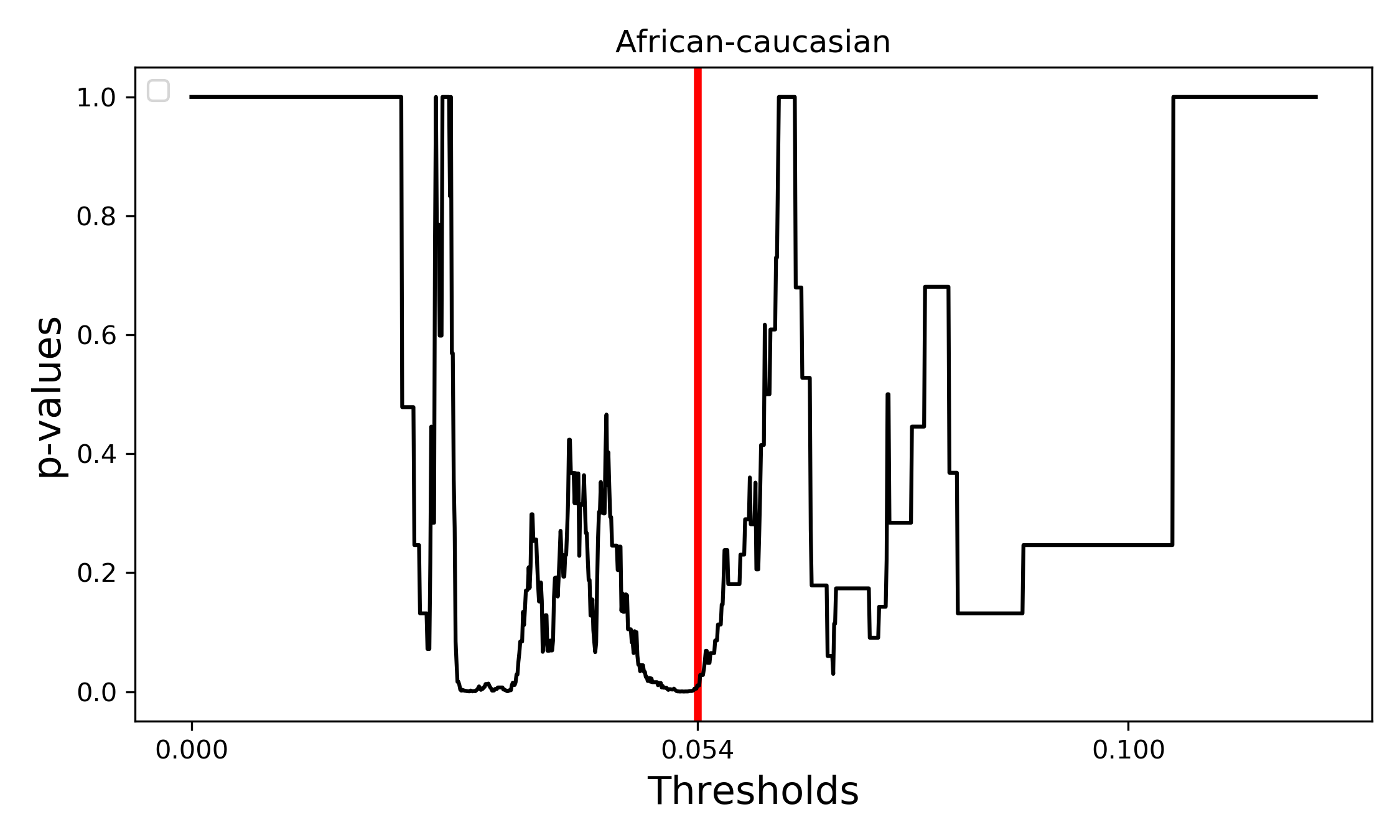} \hfill 
\includegraphics[width=0.33\textwidth]{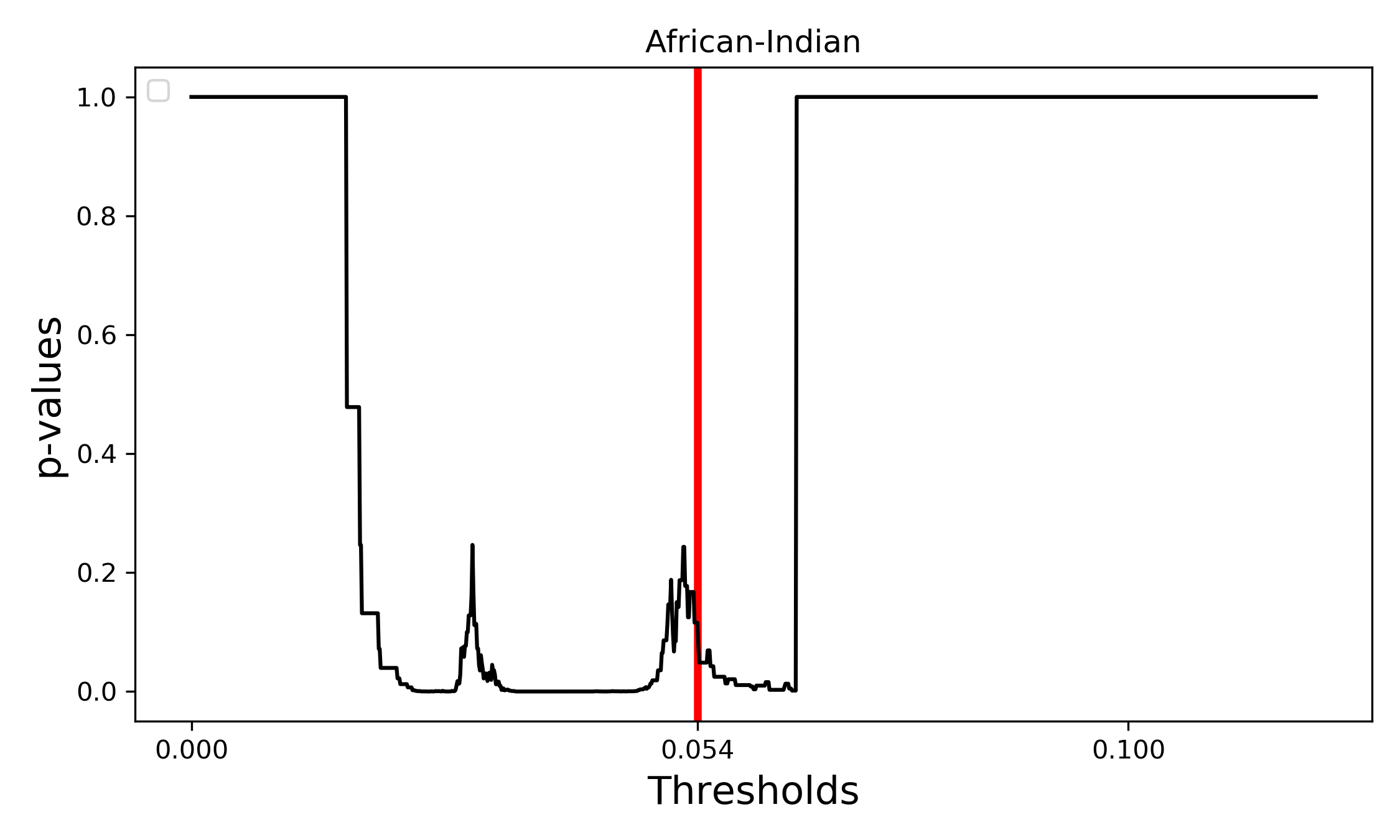}
\includegraphics[width=0.33\textwidth]{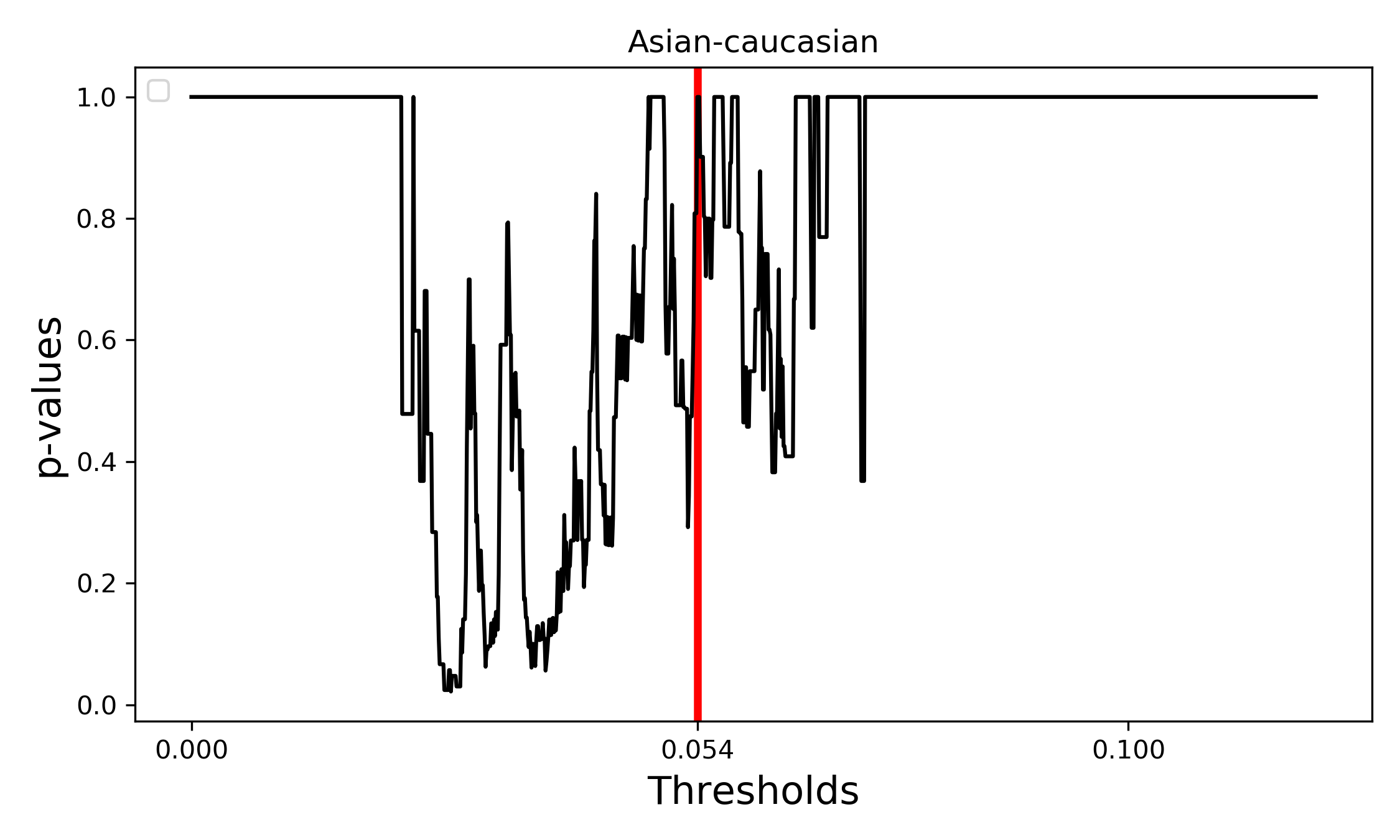} \hfill 
\includegraphics[width=0.33\textwidth]{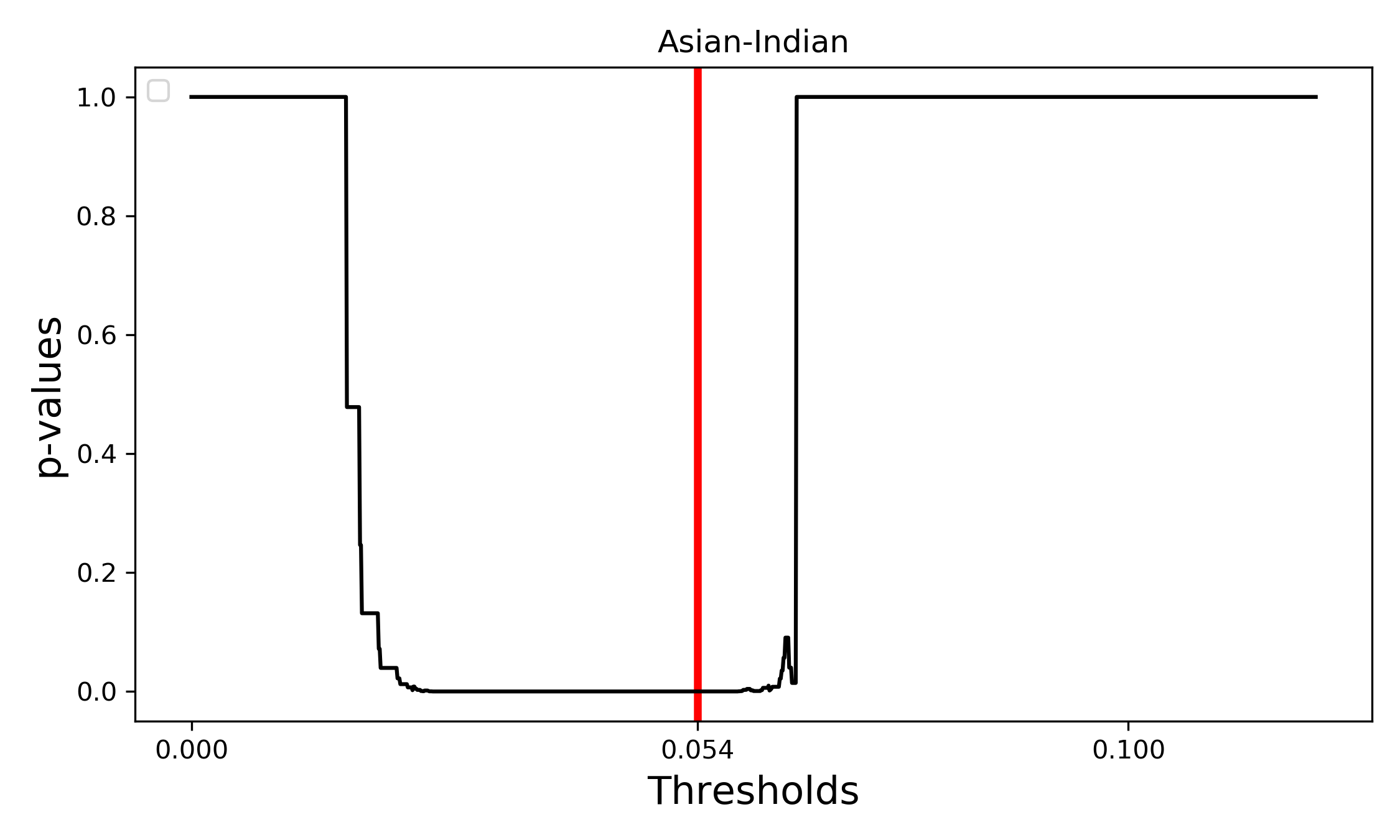} \hfill 
\includegraphics[width=0.33\textwidth]{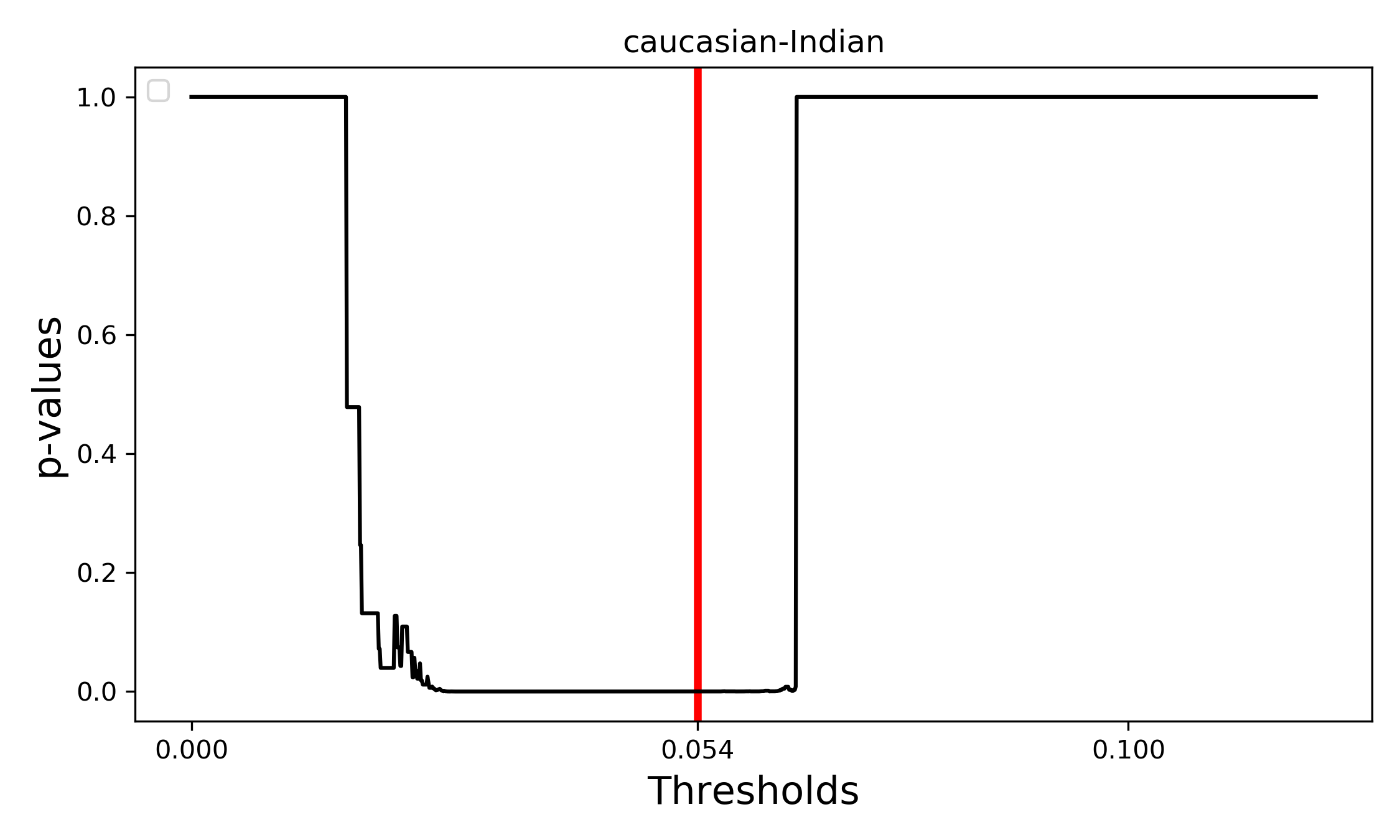}
\caption{For each pair of races, graphs of the p-value as a function of the threshold. The classifier was trained on RA.}
\label{fig:SiWp-values-SiW}
\end{figure*}

Next, we treat the thresholds, that is, the operating points on the ROC curve, as a variable. In Figs.~\ref{fig:SiWp-values-RA},~\ref{fig:SiWp-values-SiW}, we plot the p-value as a function of the response, for the two classifiers and the six pairs of races. We note that, over the range of thresholds, there are several disconnected intervals corresponding to high bias, which means that threshold optimisation for low bias should not assume unique solutions, as it is often implicit in the literature. 



\subsection{Statistical analysis of the scalar responses}

\begin{figure*}[ht]
\includegraphics[width=0.33\textwidth]{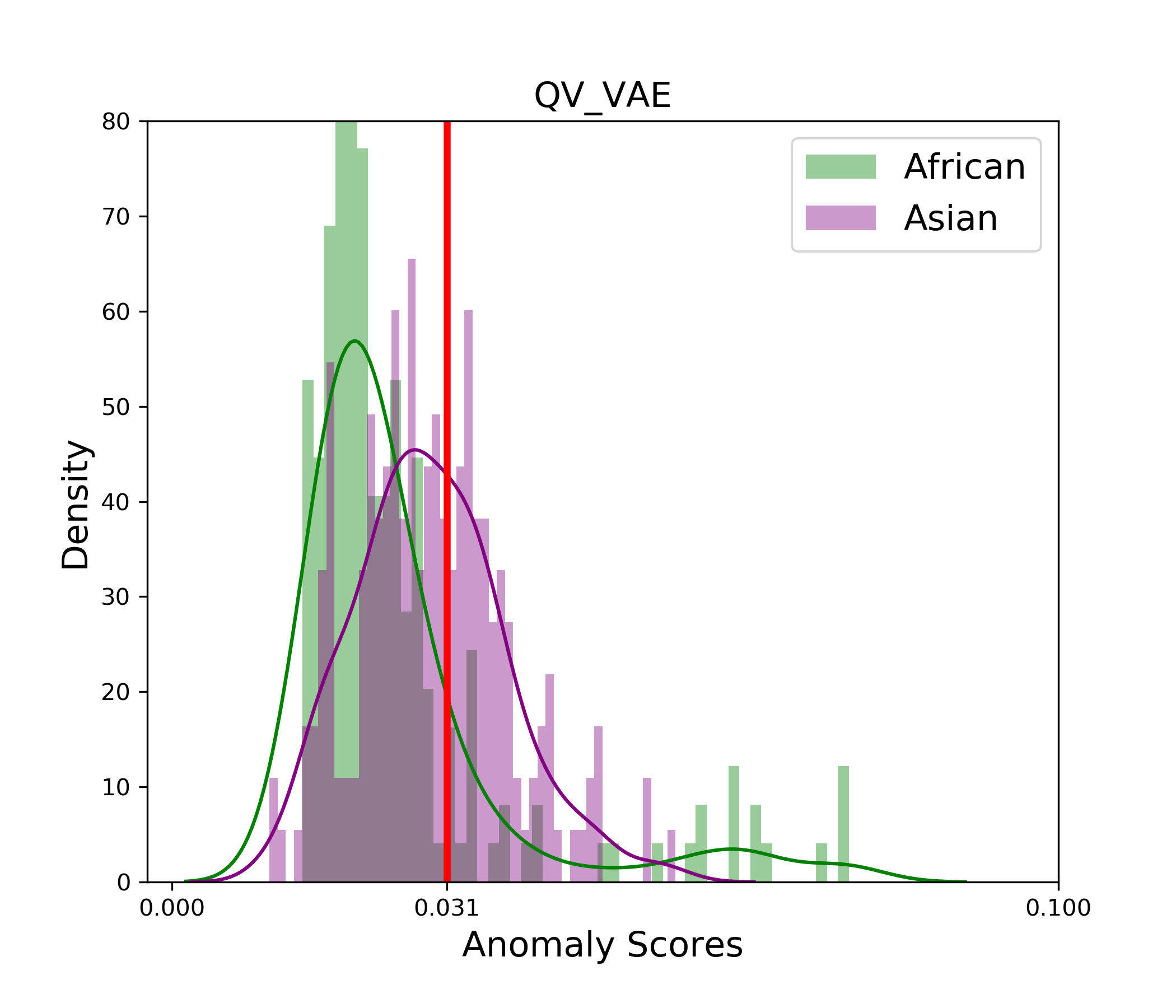} \hfill
\includegraphics[width=0.33\textwidth]{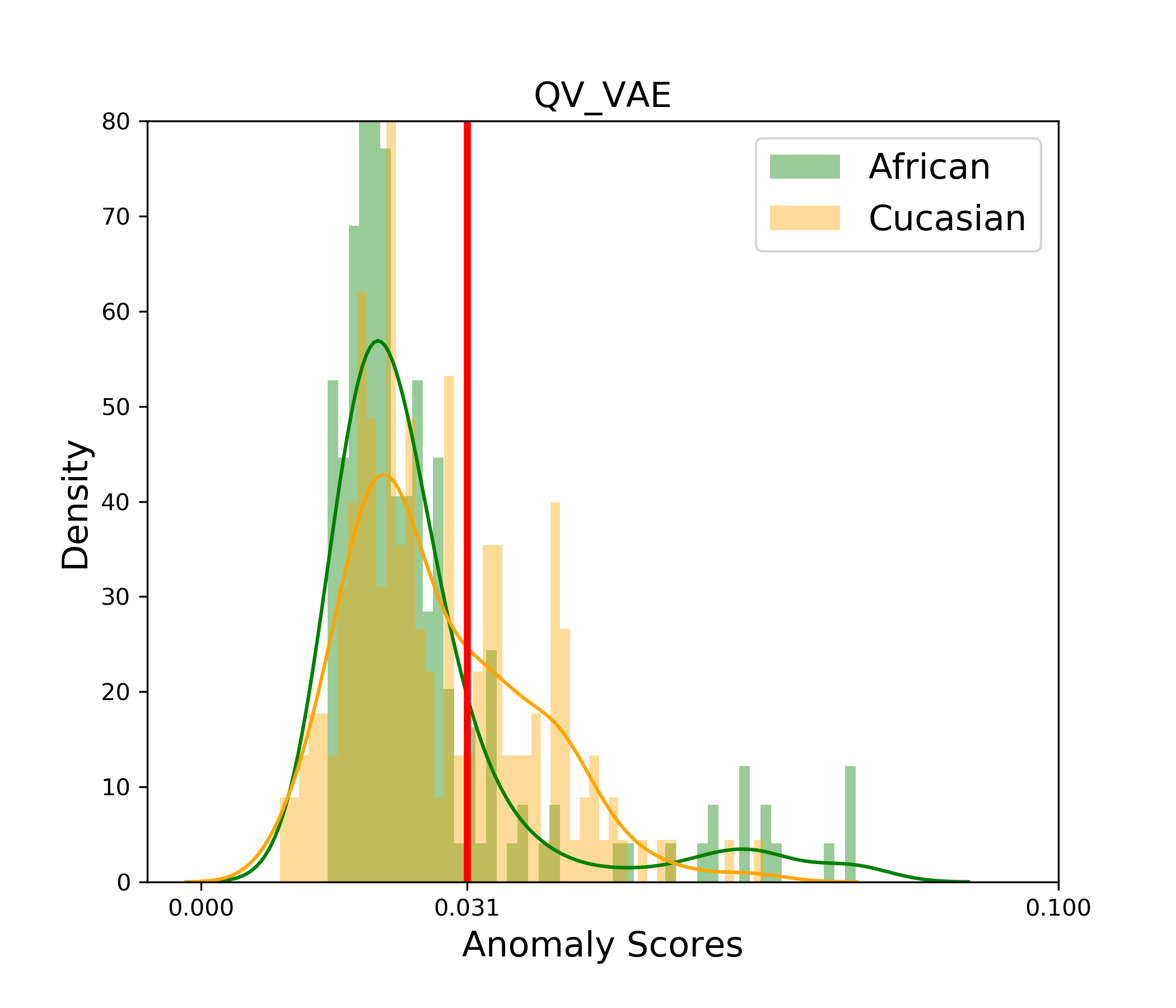} \hfill
\includegraphics[width=0.33\textwidth]{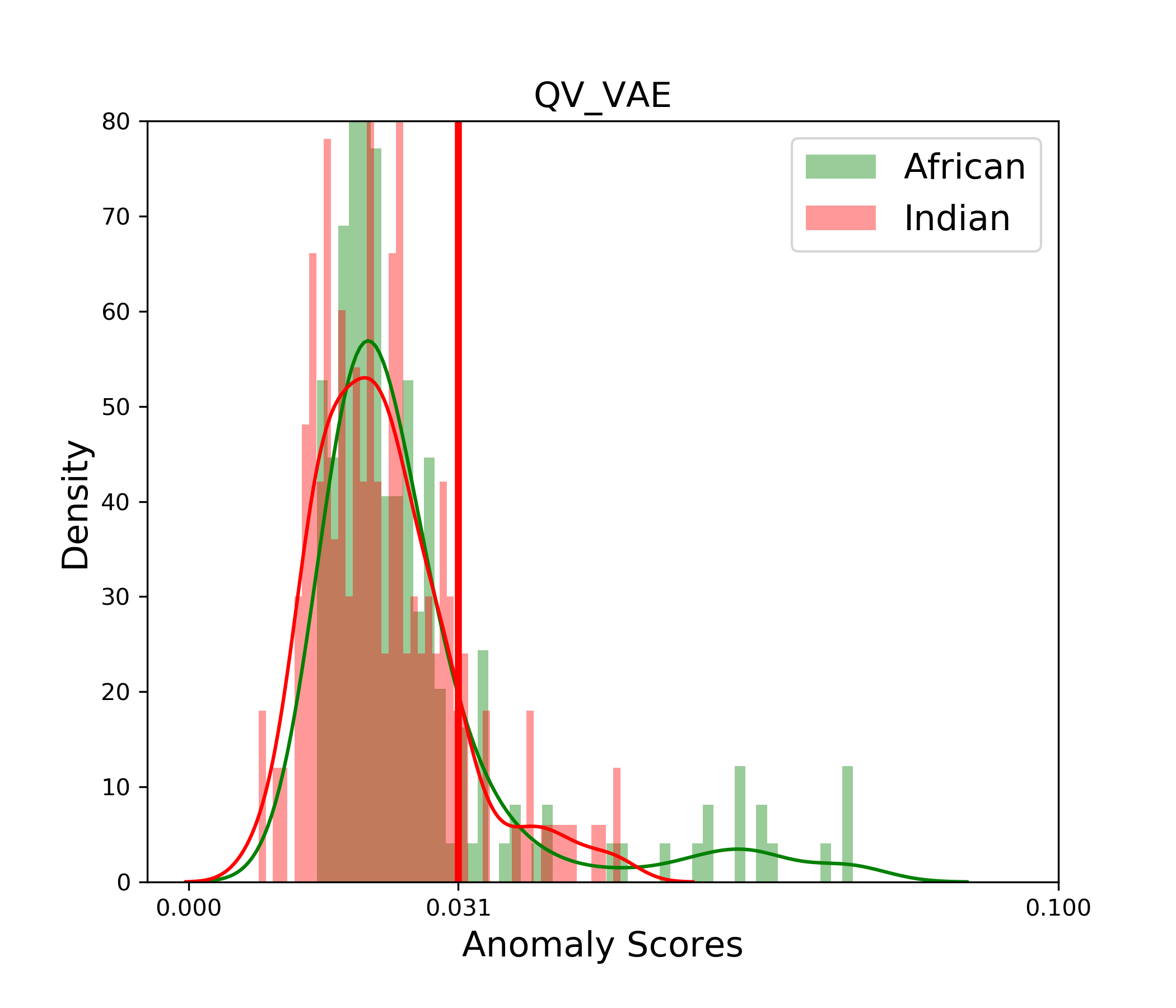} 
\includegraphics[width=0.33\textwidth]{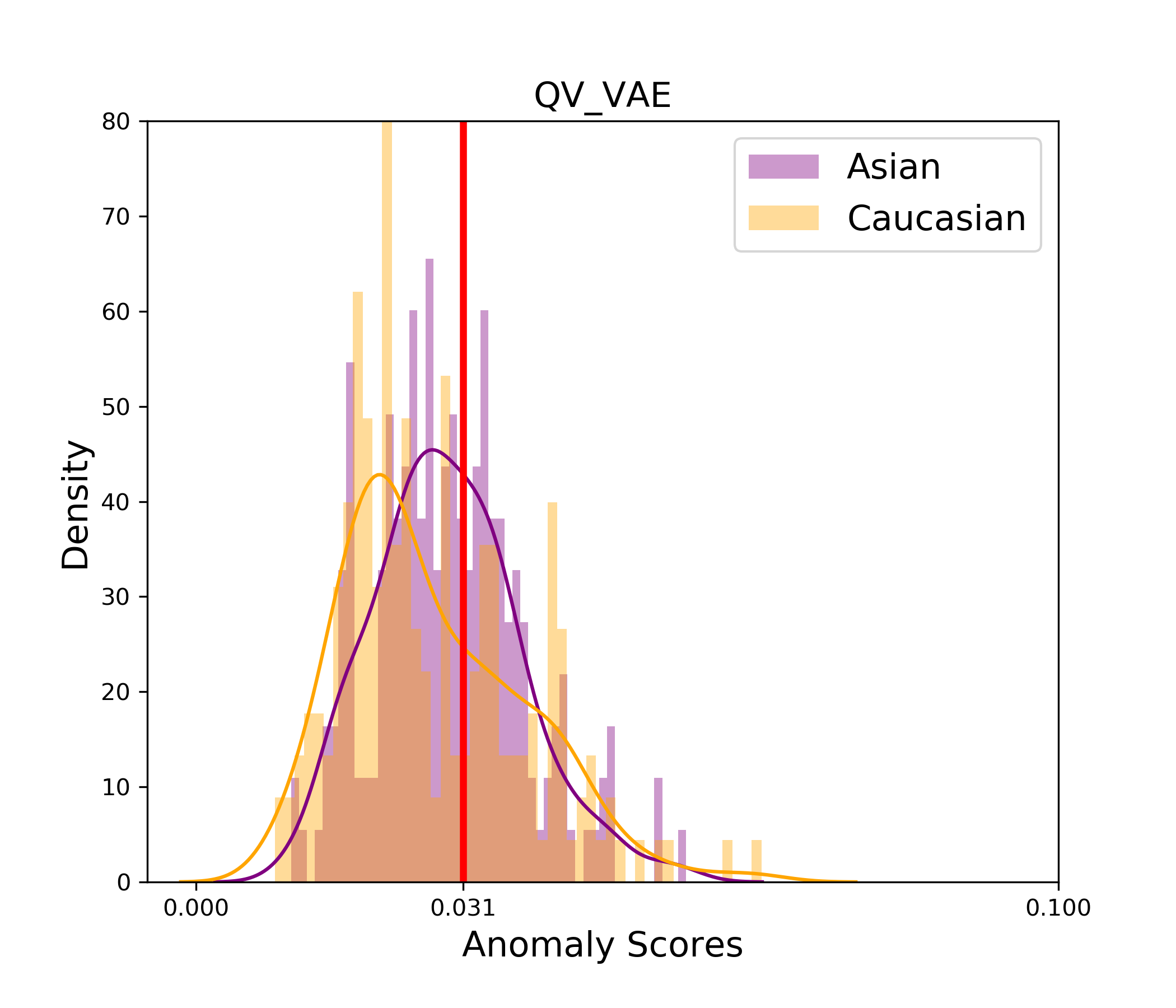} \hfill
\includegraphics[width=0.33\textwidth]{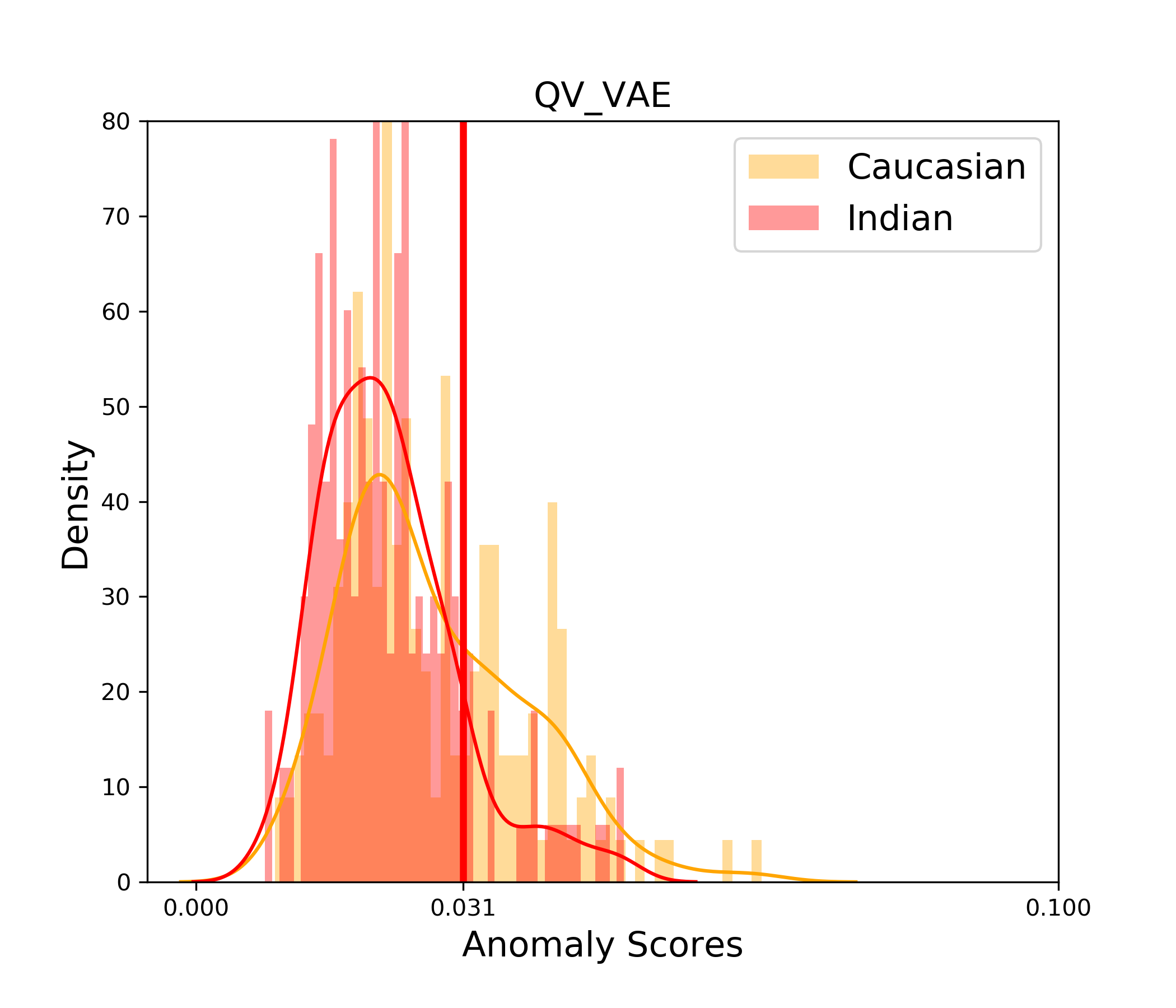} \hfill
\includegraphics[width=0.33\textwidth]{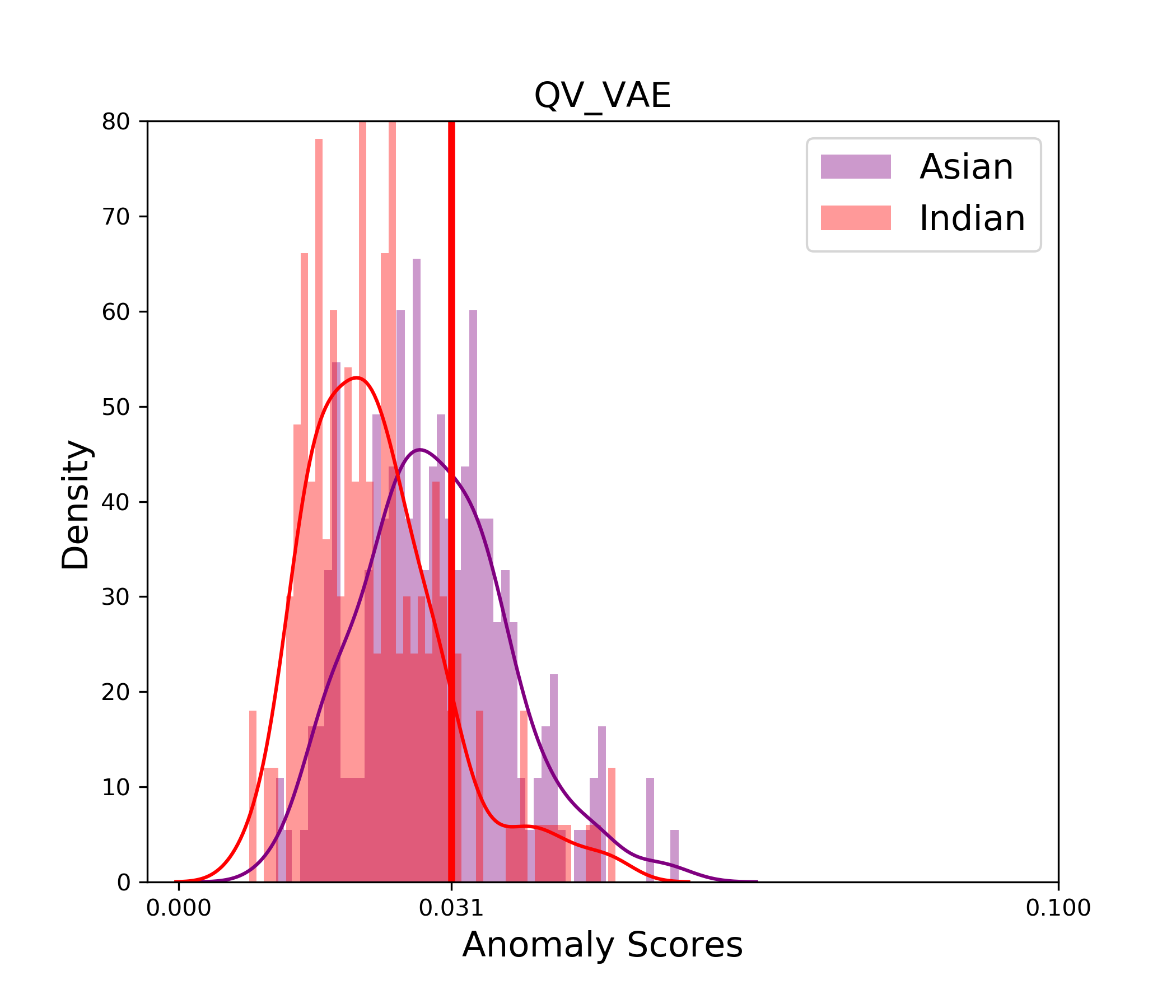} 
\caption{For each pair of races in SiW, the histogram of the responses on bona-fide images. The classifier was trained on RA.}
\label{fig:SiW-Histograms-RA}
\end{figure*}

\begin{figure*}[ht]
\includegraphics[width=0.33\textwidth]{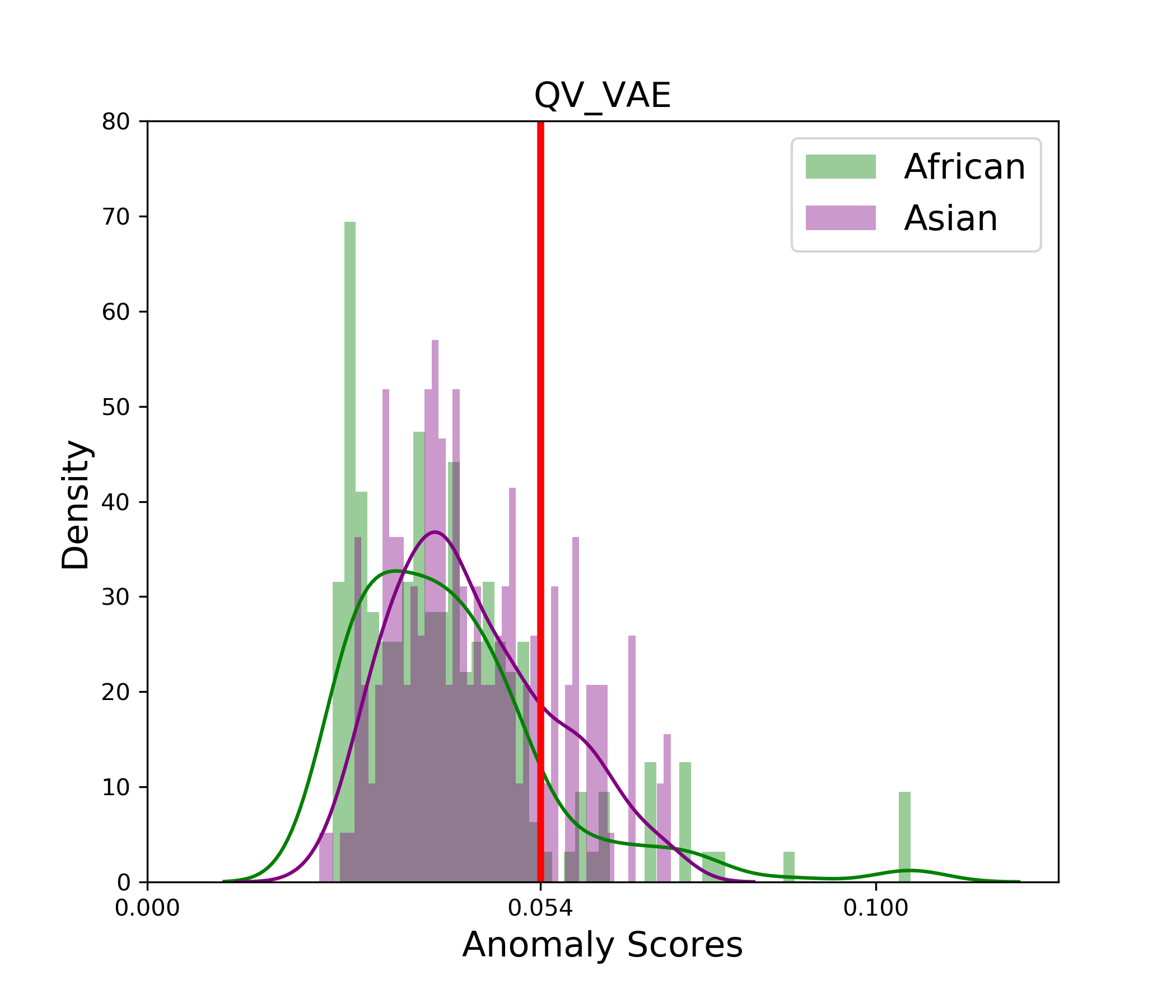} \hfill
\includegraphics[width=0.33\textwidth]{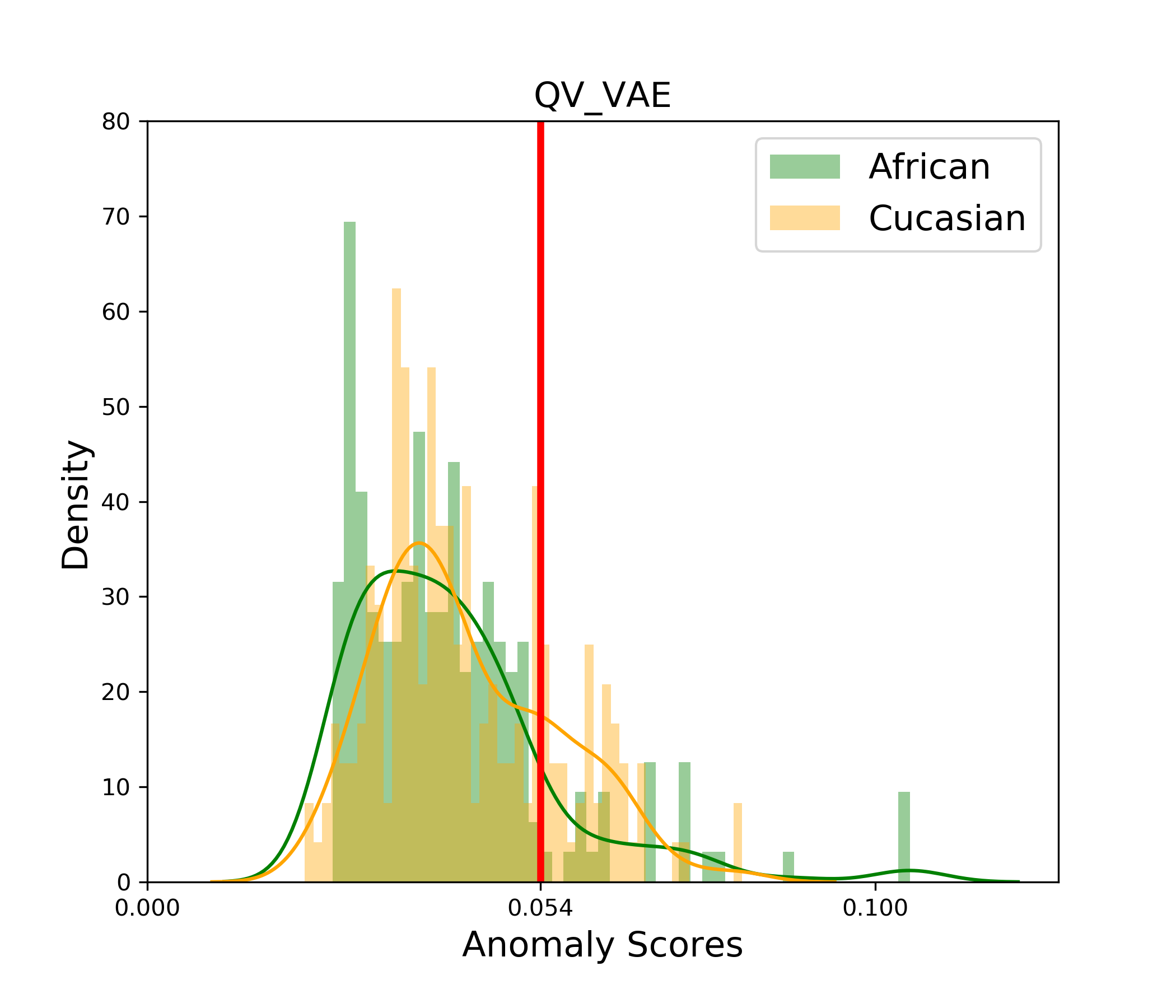} \hfill
\includegraphics[width=0.33\textwidth]{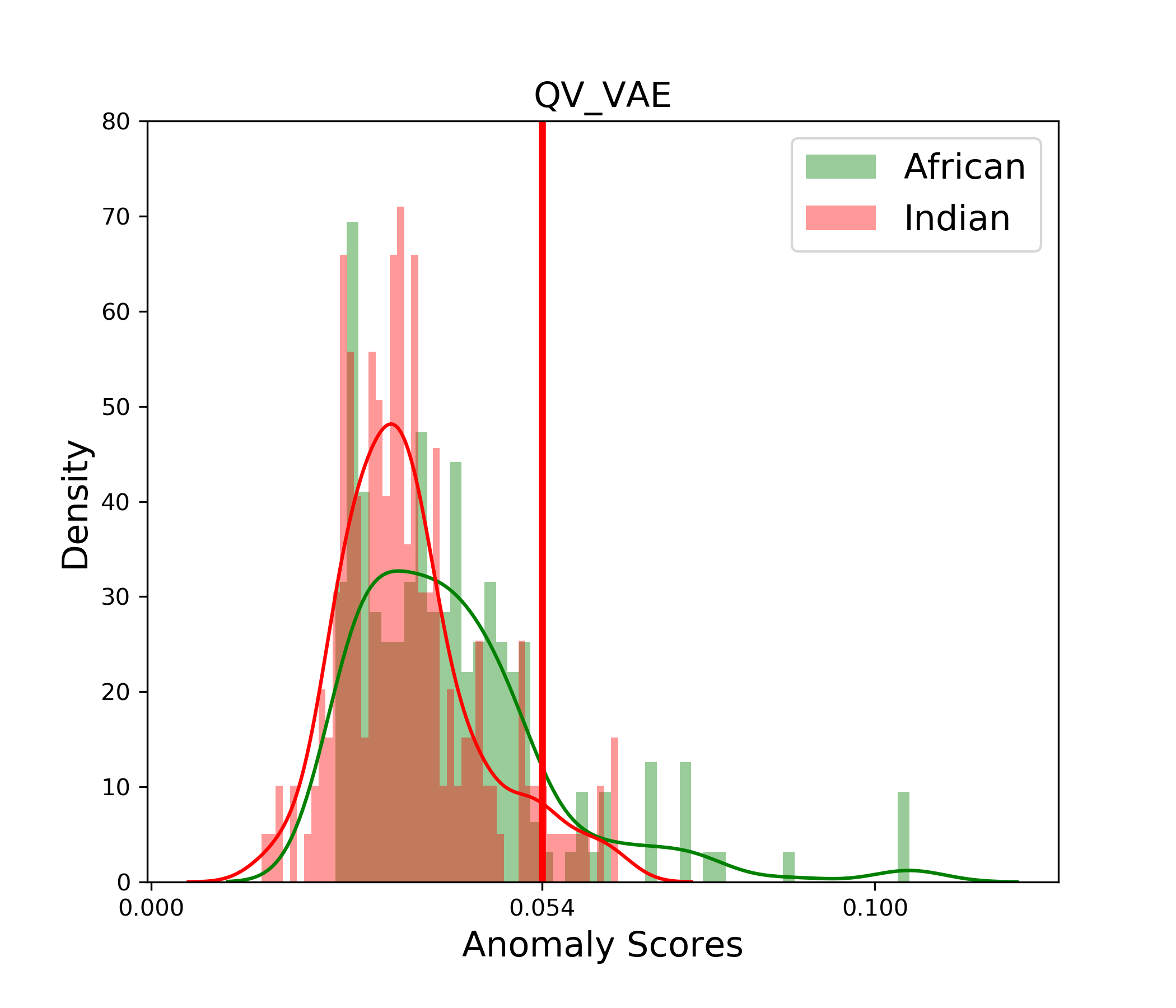} 
\includegraphics[width=0.33\textwidth]{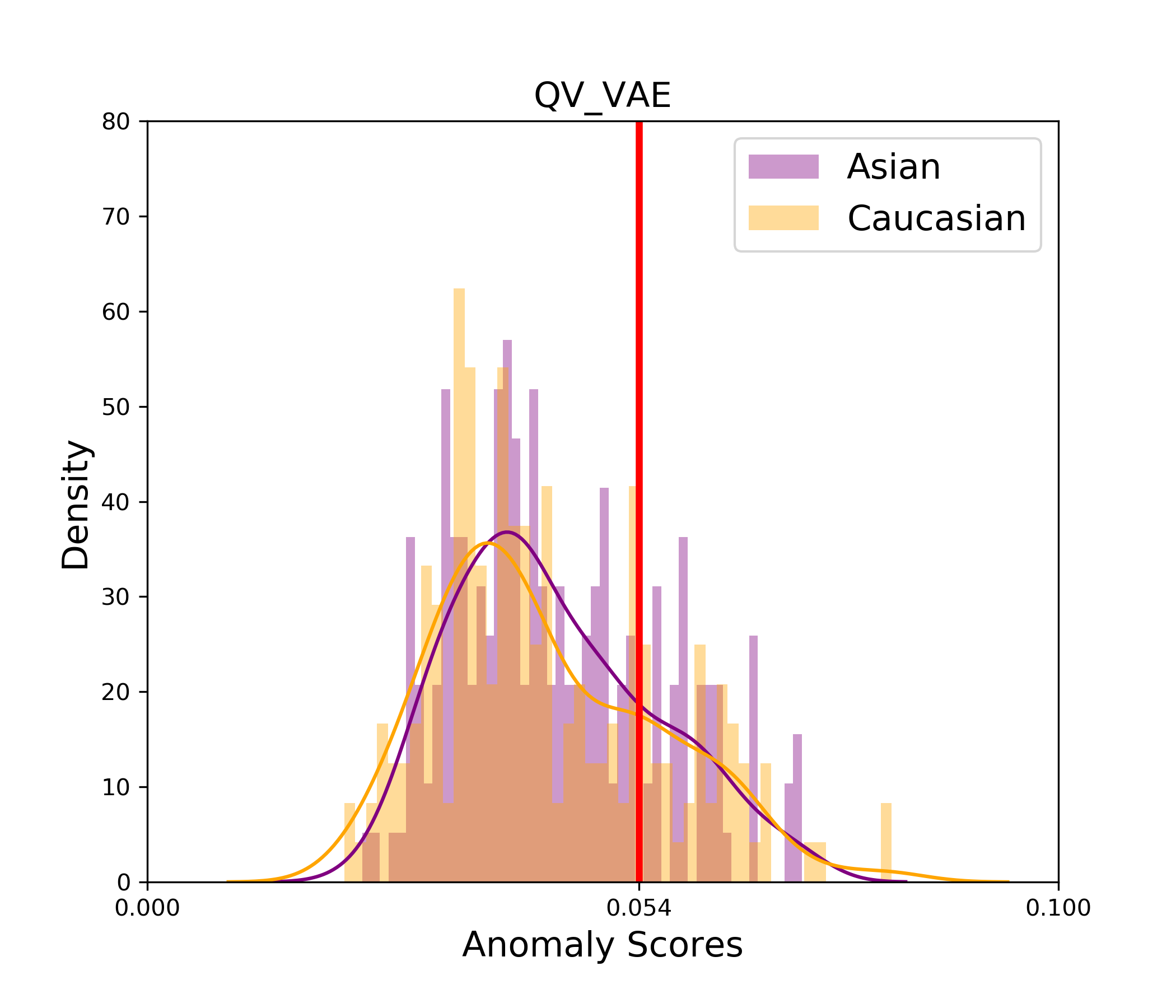} \hfill
\includegraphics[width=0.33\textwidth]{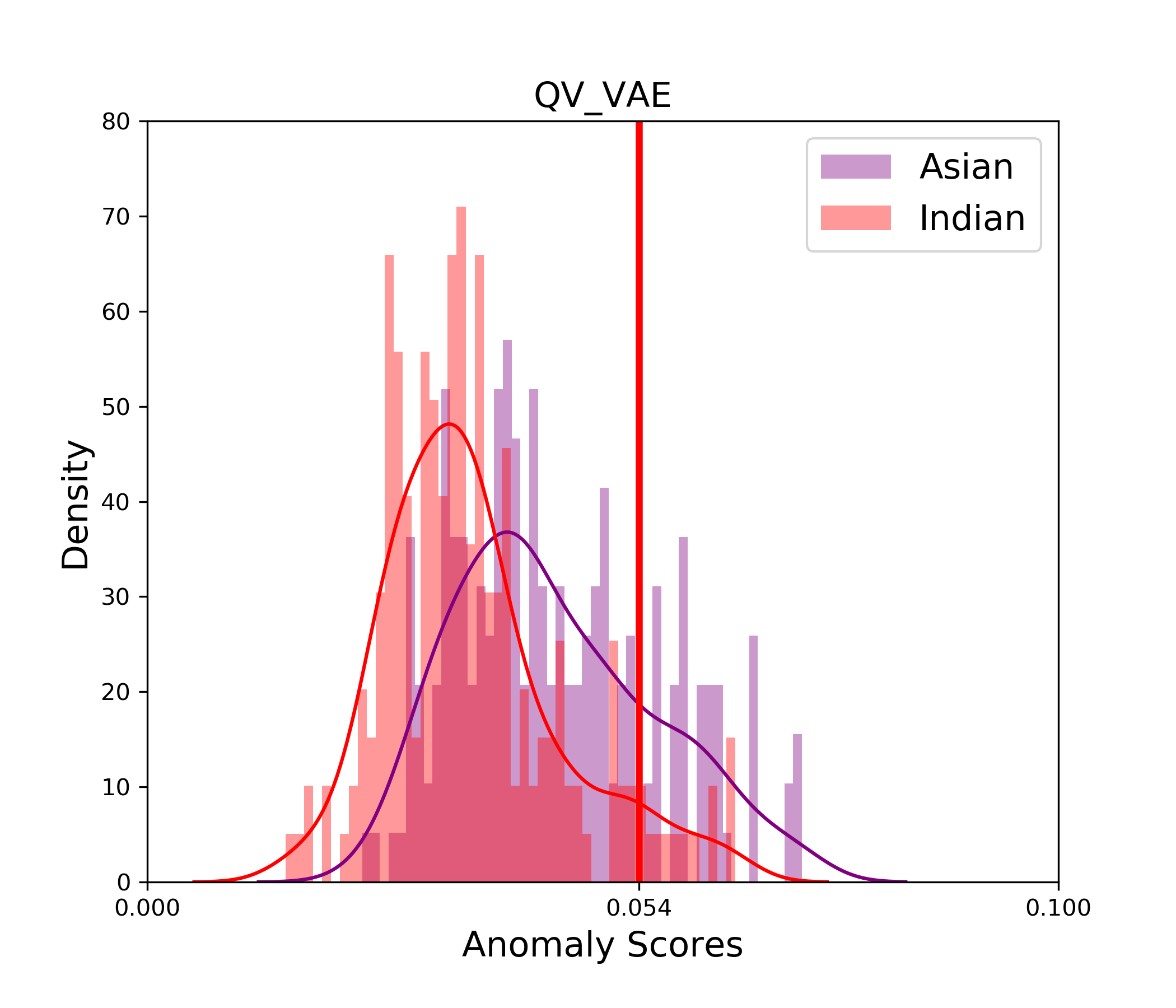} \hfill
\includegraphics[width=0.33\textwidth]{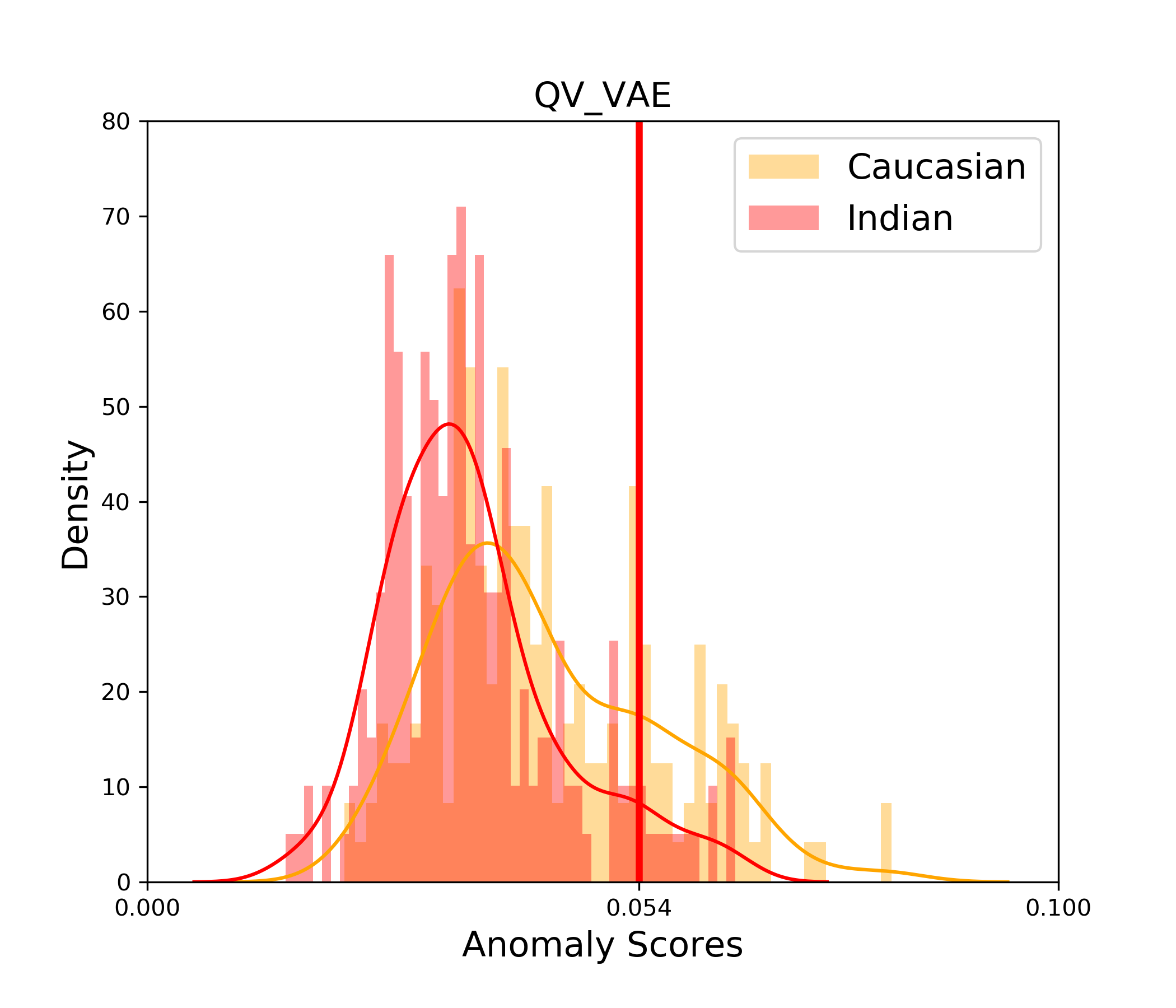} 
\caption{For each pair of races in SiW, the histogram of the responses of bona-fide images. The classifier was trained on SiW.}
\label{fig:SiW-Histograms-SiW}
\end{figure*}

For an insight in the behaviour of the graphs of the p-values we analyse the scalar responses of the algorithm. Figs.~\ref{fig:SiW-Histograms-RA},~\ref{fig:SiW-Histograms-SiW} show a comparison of the histograms of the responses for each pair of races. We note the complex behaviour of the density functions, which induce a complex bias behaviour. In particular, bias at certain thresholds can be caused by differences in the means of the responses; differences in their variances; response bimodality; or outliers. 

Regarding the means of the responses, since the Shapiro-Wilk test rejected the hypothesis of normal distributions, we opted for the Mann–Whitney U test.
Table~\ref{tbl:tTest-SiW} shows the computed p-values on each pair of races, for the hypothesis that the values of randomly selected responses from the two populations are different. 
\begin{table}[ht]
\caption{p-values of the Mann–Whitney U test on each pair of races.}
\centering
\begin{tabular}{|c||c|c|c|c|c|c|} \hline 
            &  Af-As & Af-Ca & Af-In & As-Ca & As-In & Ca-In \\ \hline\hline
RAtr  &  .0000&  .0015 & .0085 & .0052   & .0000 & .0000 \\ \hline 
SiWtr &  .0001 & .0078  &.0000& .1560 & .0000  & .0000 \\ \hline 
\end{tabular}
\label{tbl:tTest-SiW}
\end{table}

Regarding bimodality, we used Hartigan's Dip Test \cite{hartigan1985dip}, with 50 bins, to test each race for bimodality. We note that for the 200 samples we have from each race, a statistical significance of 95\% corresponds to a critical value of 0.037. Table~\ref{tbl:meansVAR-responses-SiW} shows the computed dip values, together with the means and standard deviations for each population. 

We note that the biggest difference in mean response is between Asian and Indian from the SiW trained classifier, while the corresponding standard deviations are similar. Thus, the bias we observe in the corresponding diagram in Fig~\ref{fig:SiWp-values-SiW} is due to the classifier's higher mean response on Asians compared to Indians. 

On the other hand, the smallest difference in mean response is between Asians and Caucasians, again under the SiW trained classifier. Thus, the bias we observe in the corresponding diagram in Fig~\ref{fig:SiWp-values-SiW}, which for very small threshold values is statistically significant, is due to different variances. We note here, that while at such low threshold values a standalone classifier wouldn't be very useful, it is still an interesting case when constructing ensembles of weak classifiers. 

Finally, we notice that all Hartigan's Dip Test values are below the significance threshold, and thus, all populations should be considered unimodal. In particular, that means that some very high responses on African people, especially from the RA trained classifier, should be treated as outliers. We note that, nevertheless, against all the other three races, these outliers create a second, or third region of high bias, and in those regions samples from the African population are treated less favourably.

\clearpage

\begin{table}[ht]
\caption{SiW testset: means, standard deviations, and Hartigan's dip values for the responses of the RA and SiW trained classifiers.}
\centering
\begin{tabular}{|c|c|c|c|c|} \hline 
 & \textbf{Af} & \textbf{As} & \textbf{Ca} & \textbf{In} \\ \hline \hline
RAtr $\mu$ & .0257 & .0291 & .0274 & .0223 \\ \hline 
SiWtr $\mu$ & .0418 & .0446 & .0438 & .0355 \\ \hline \hline
RAtr $s.d.$ & .0126 & .0084 & .0104 & .0079 \\ \hline 
SiWtr $s.d.$ & .0142 & .0109 & .0122 & .0096\\ \hline \hline
RAtr dip & .0221 & .0360& .0335 & .0349 \\ \hline 
SiWtr dip & .0299 & .0233 & .0366 & .0324 \\ \hline 
\end{tabular}
\label{tbl:meansVAR-responses-SiW}
\end{table}


\subsection{Discrete latent space}

The proposed VQ-VAE encodes each image with a 32768 number long sequence of integers in [0..1023], corresponding to indices of the vectors in the codebook. Using these latent space representations of the images as inputs, for each pair of races, we train an SVM with an RBF kernel of Gaussian type. Table~\ref{tbl:QV-SVM-SiW} shows AUC values for these SVM classifiers.

\begin{table}[h]
\caption{AUC values for an SVM trained on the VQ-VAE's latent space encodings of the images. SiW testset.}
\centering
\begin{tabular}{|c||c|c|c|c|c|c|} \hline 
      &  Af-As & Af-Ca & Af-In & As-Ca & As-In & Ca-In \\ \hline\hline
RAtr  &  .88 & .87 & .74 &  .71 & .78 &  .83 \\ \hline 
SiWtr &  .74 & .85 & .76 & .72  &  .76 & .83\\ \hline 
\end{tabular}
\label{tbl:QV-SVM-SiW}
\end{table}

We note that in some cases the SVMs perform quite well, meaning that the VQ-VAE encodings of the face images contain enough information to discriminate between races, despite the fact that VQ-VAEs themselves were not trained with race labels. That means that there is a potential for biased behaviour, however, on the other hand, the statistical significance of the result and its practical implications are not clear, and a deeper investigation is required.

%% file: sec5.tex
\section{Bias analysis on RFW}
\label{sec:sec5}

In this section, we apply the same bias analysis on a testset from the RFW database, consisting of 200 images from each race. We note that this time the race labels come as part of the database, rather than being annotated by us. Another difference from SiW is that the RFW database is not a specialised face anti-spoofing database. Thus, as we do not have imposter images, we do not have empirically established thresholds, as for example the ones corresponding to EER values. For the part of the analysis requiring specific thresholds, we use thresholds corresponding to some predetermined TPR values.


\subsubsection{Statistical analysis of the binary outcomes}

Table~\ref{tbl:thresholdsRfW} shows the thresholds corresponding to TPR values of 1\%, 2\%, 5\%, 10\%, 20\%. 
\begin{table}[h]
\caption{Threshold values corresponding to some predetermined TPR values.}
\centering
\begin{tabular}{|c||c|c|c|c|c|} \hline 
          & 1\%   & 2\%   & 5\%   & 10\%  & 20\% \\ \hline\hline 
RAtr  & .0460 & .0440 & .0389 & .0336 & .0279  \\ \hline 
SiWtr & .1206 & .1079 & .0971 & .0838 & .0712  \\ \hline 
\end{tabular}
\label{tbl:thresholdsRfW}
\end{table}

\begin{figure*}[h]
\includegraphics[width=0.33\textwidth]{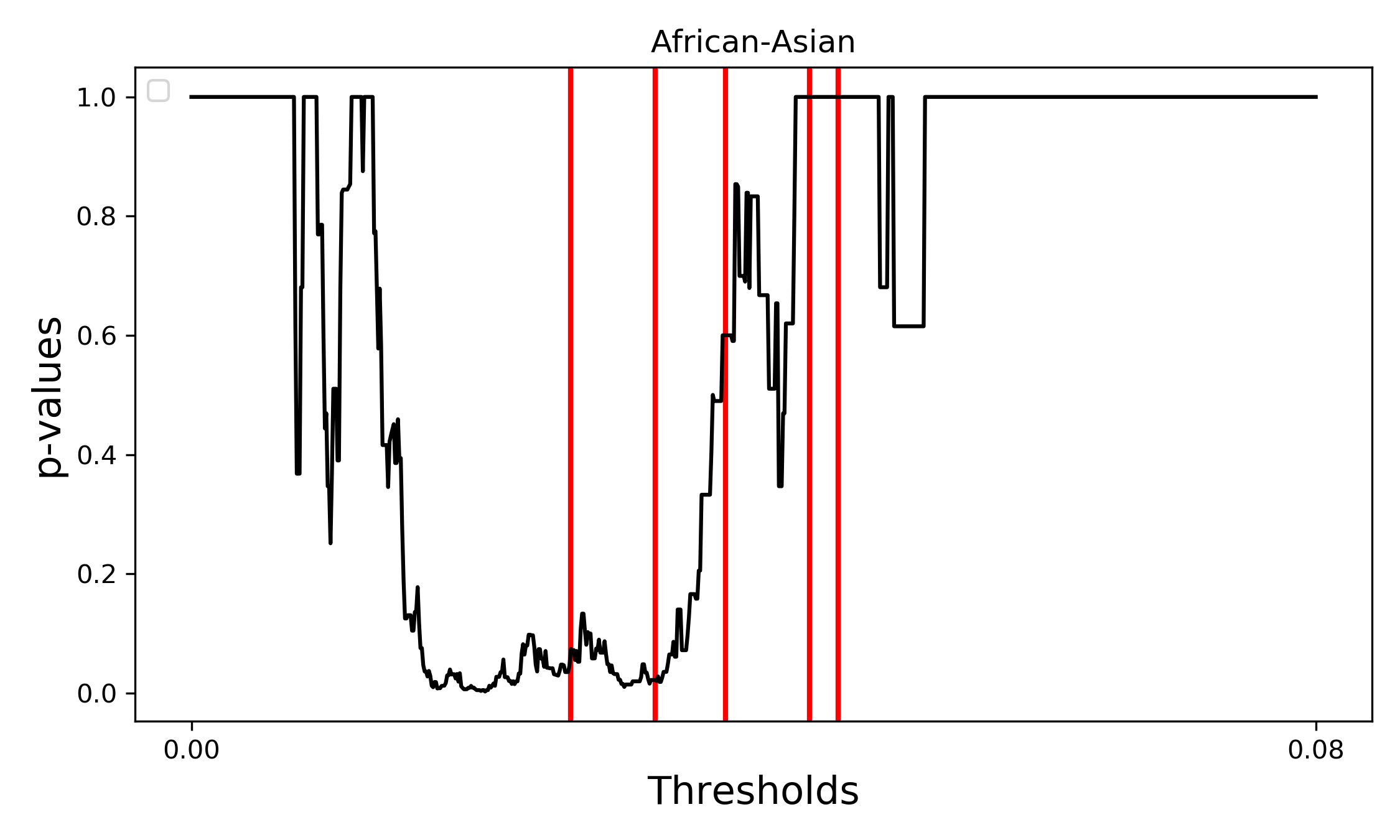} \hfill
\includegraphics[width=0.33\textwidth]{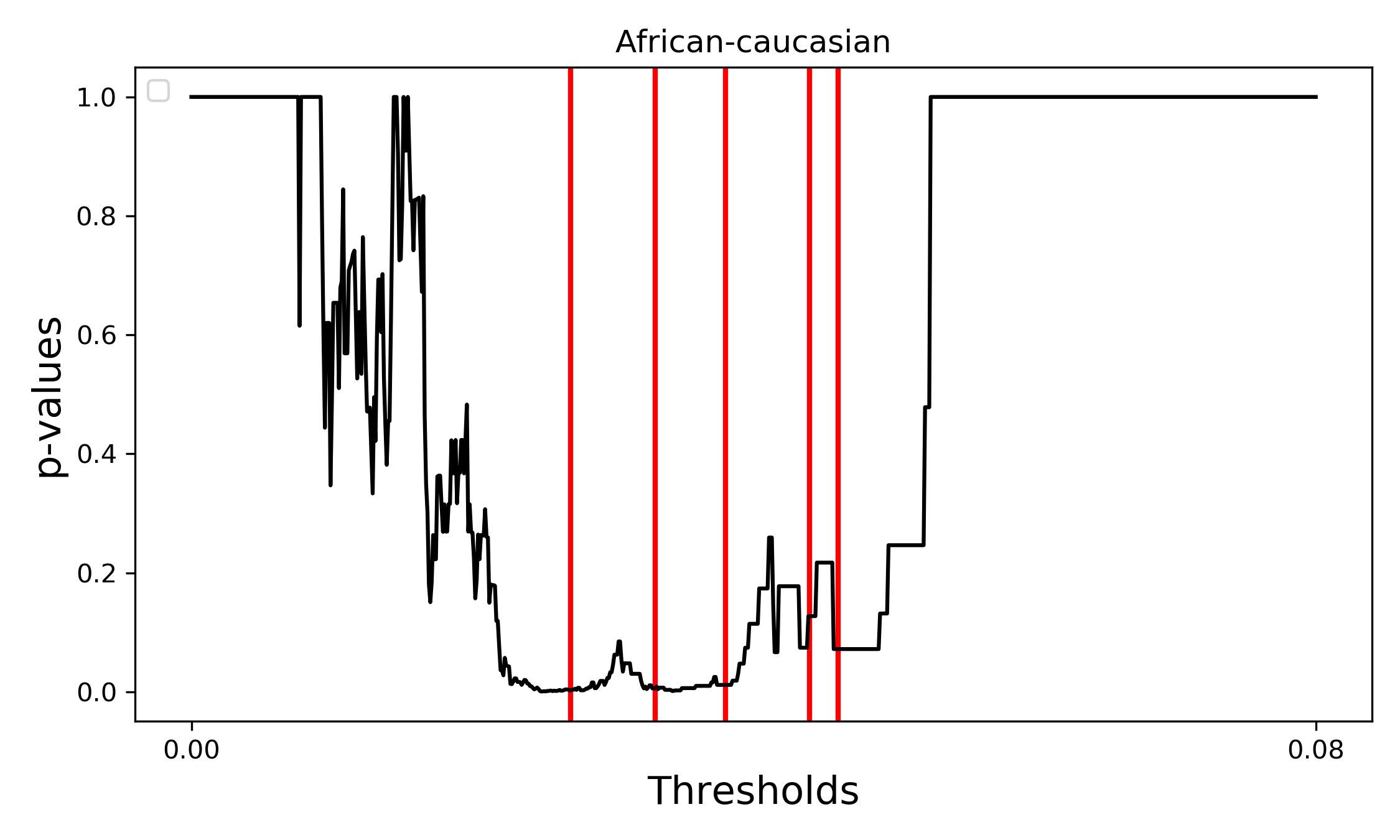}\hfill 
\includegraphics[width=0.33\textwidth]{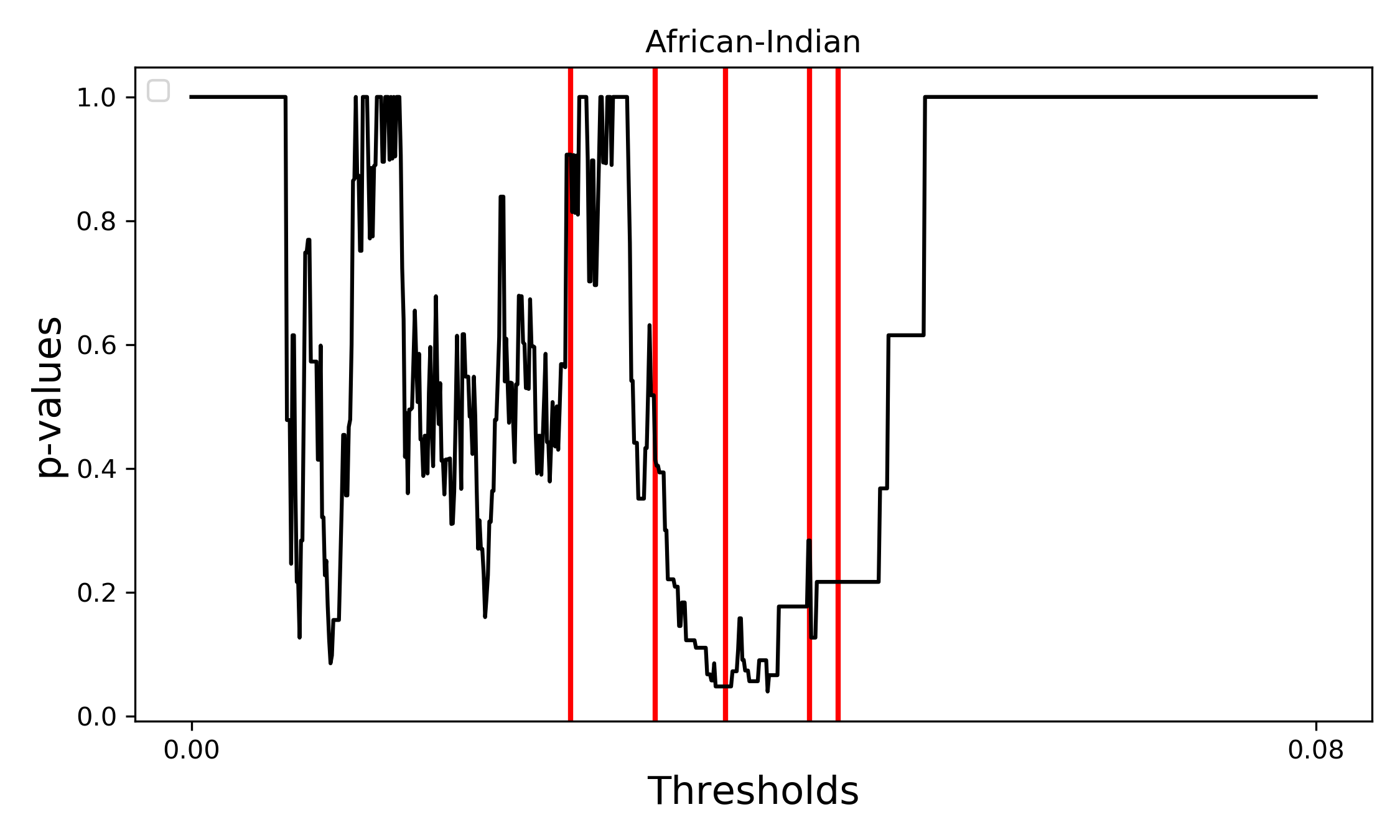}\hfill
\includegraphics[width=0.33\textwidth]{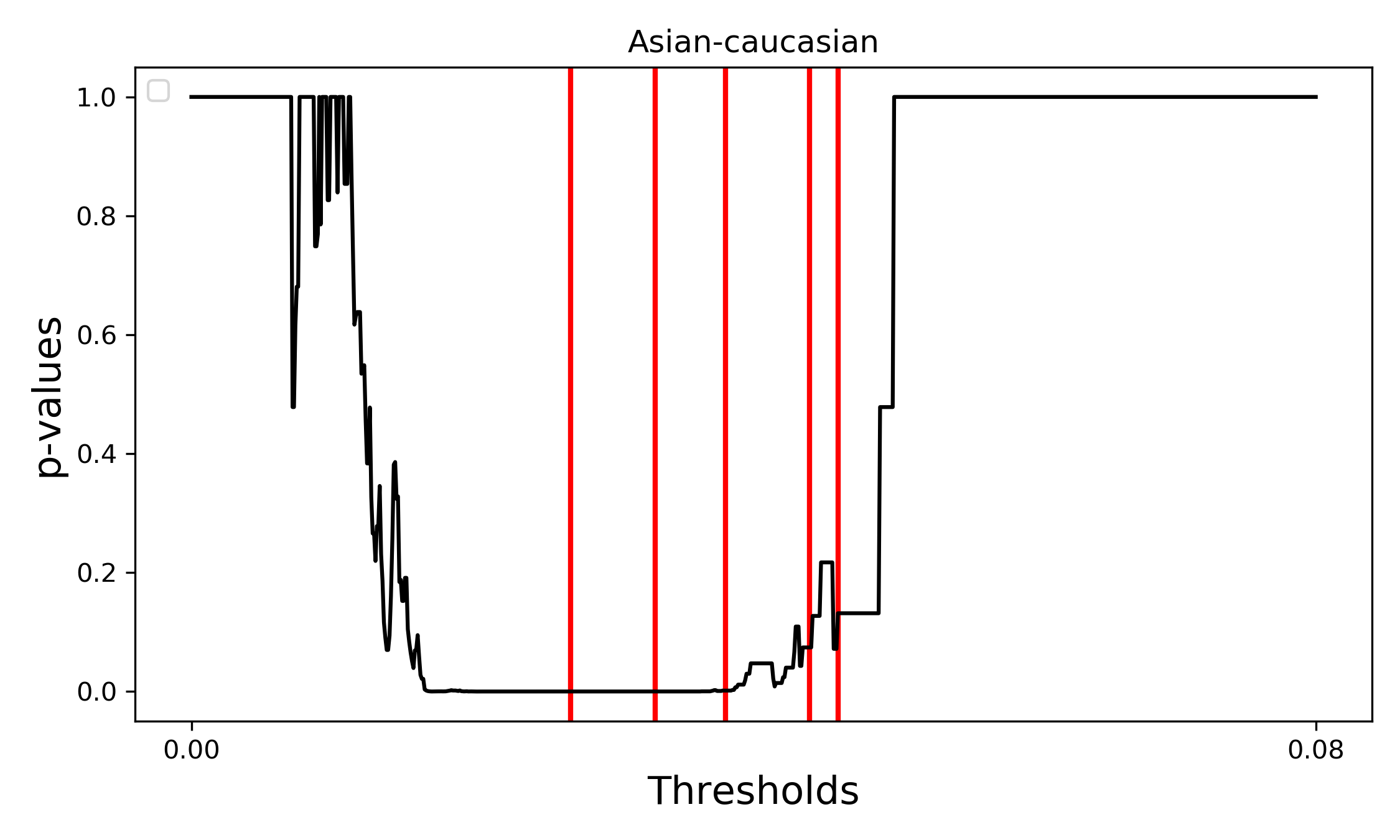}\hfill 
\includegraphics[width=0.33\textwidth]{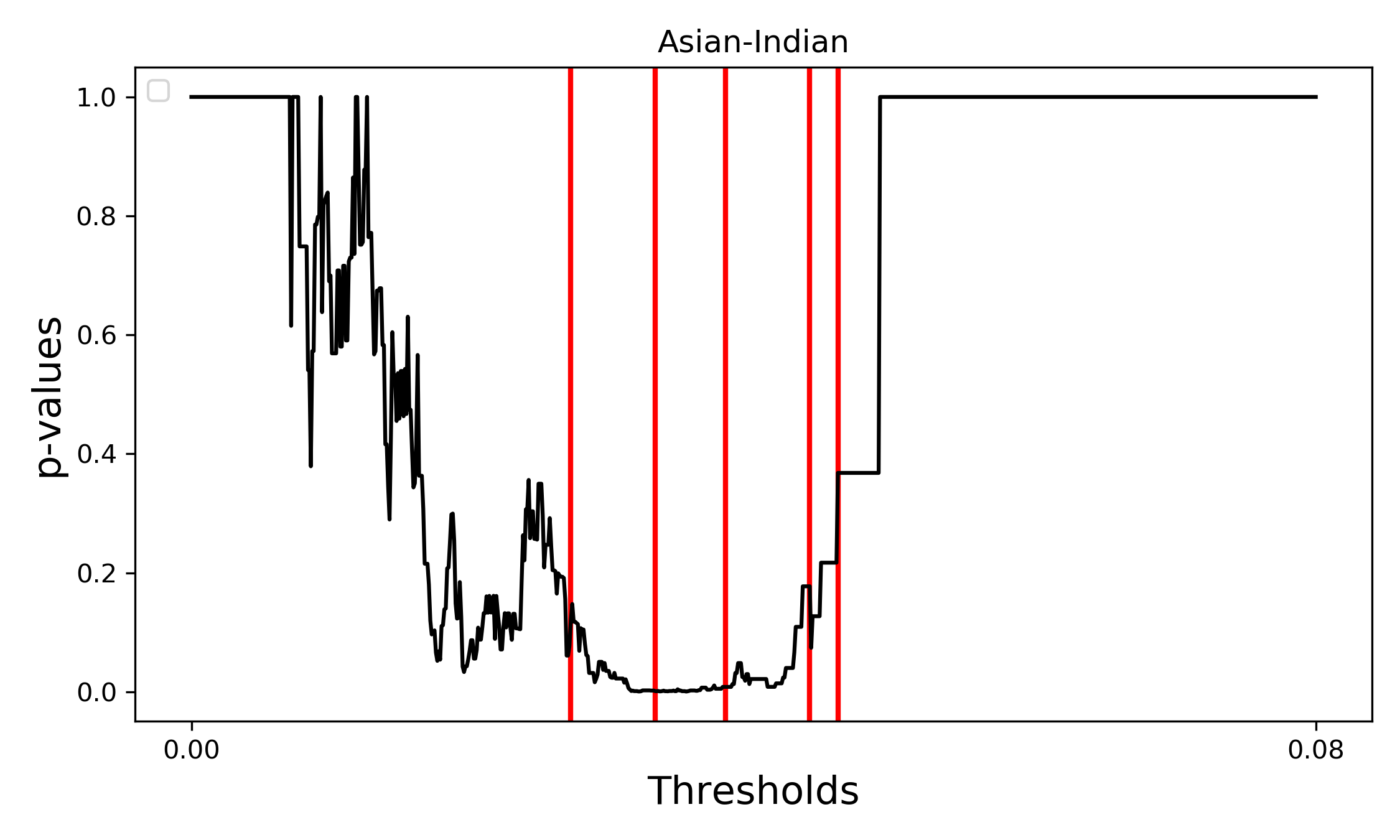}\hfill
\includegraphics[width=0.33\textwidth]{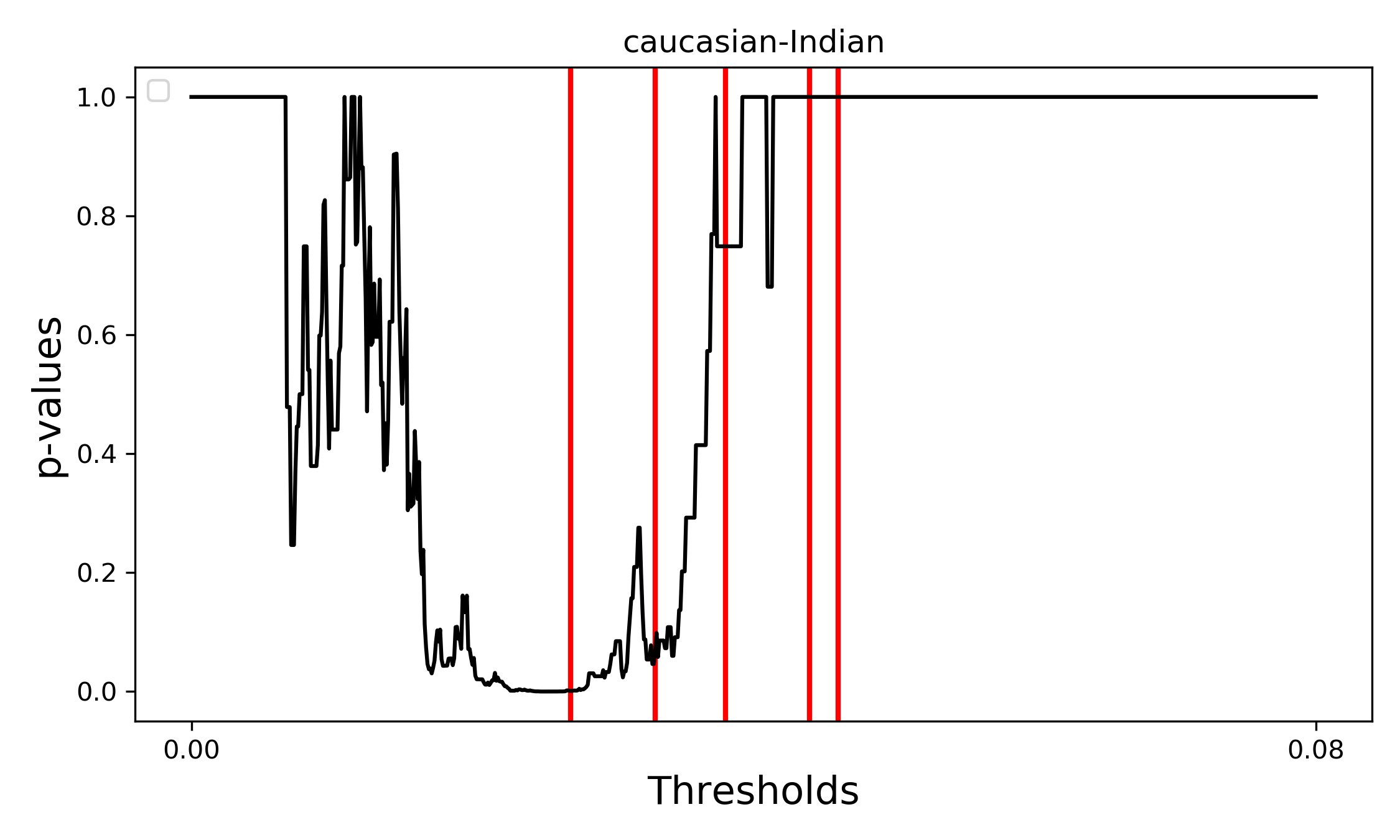} \hfill 
\caption{For each pair of races, graphs of the p-value as a function of the threshold. The classifier was trained on RA.}
\label{fig:rfwBinaryRA}
\end{figure*}

\begin{figure*}[h]
\includegraphics[width=0.33\textwidth]{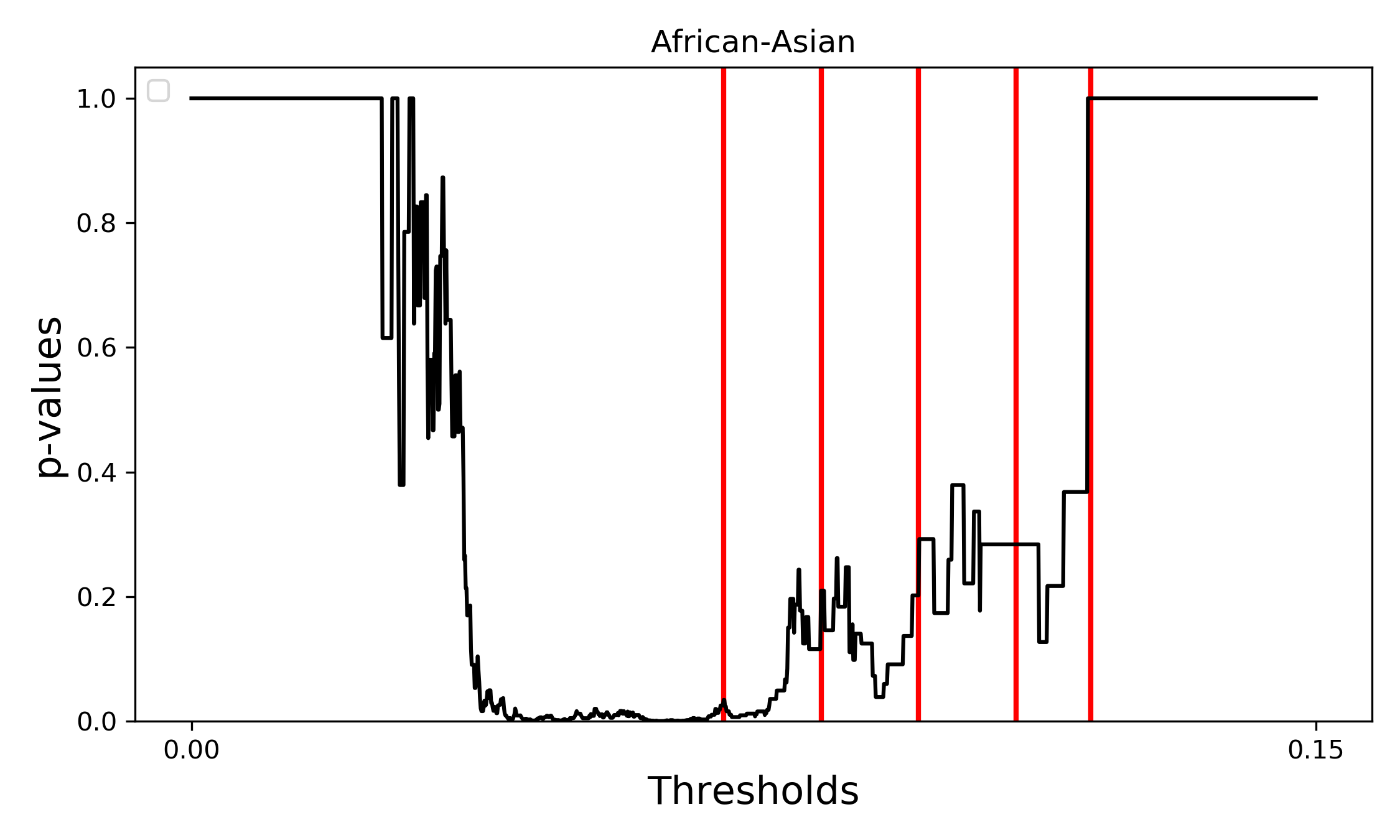} \hfill
\includegraphics[width=0.33\textwidth]{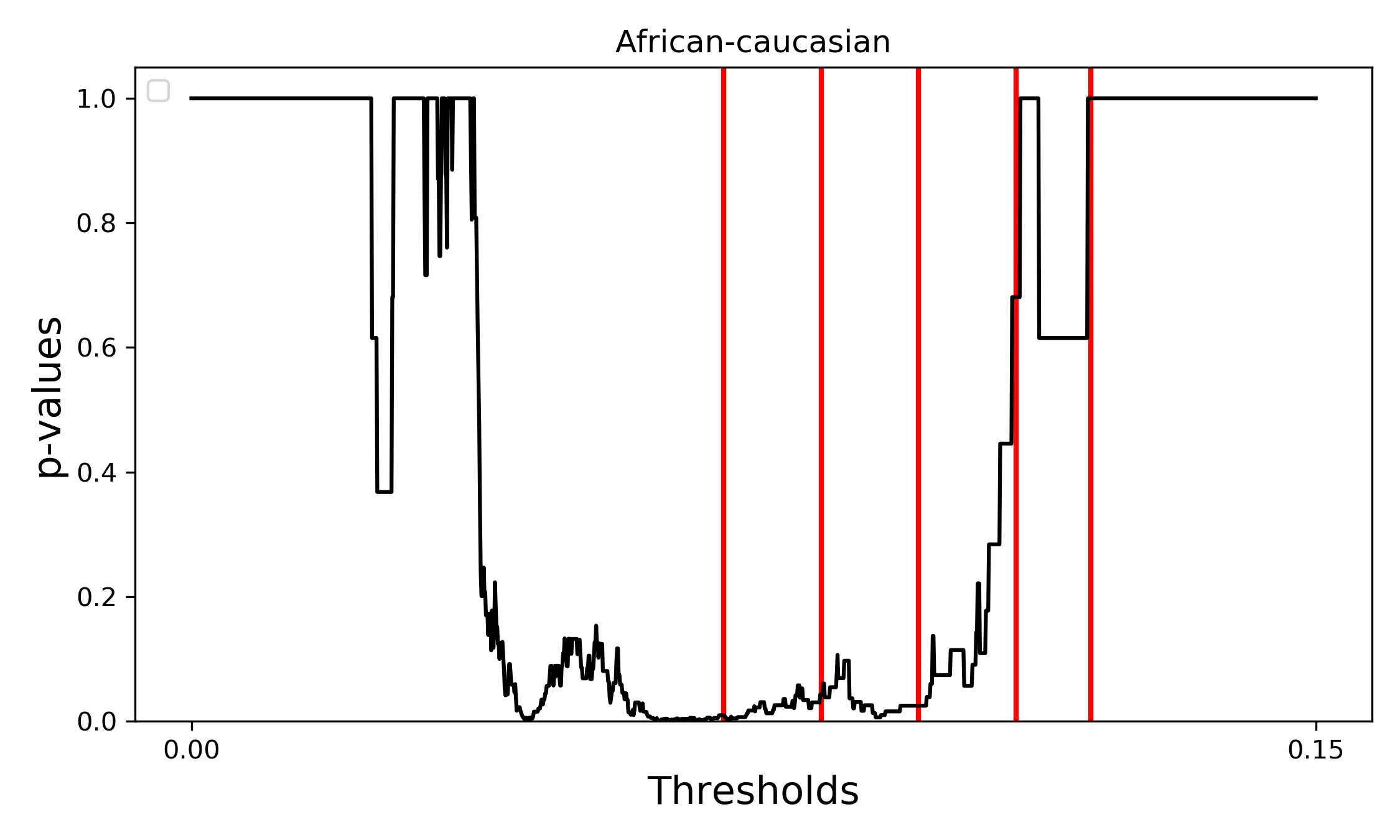}\hfill 
\includegraphics[width=0.33\textwidth]{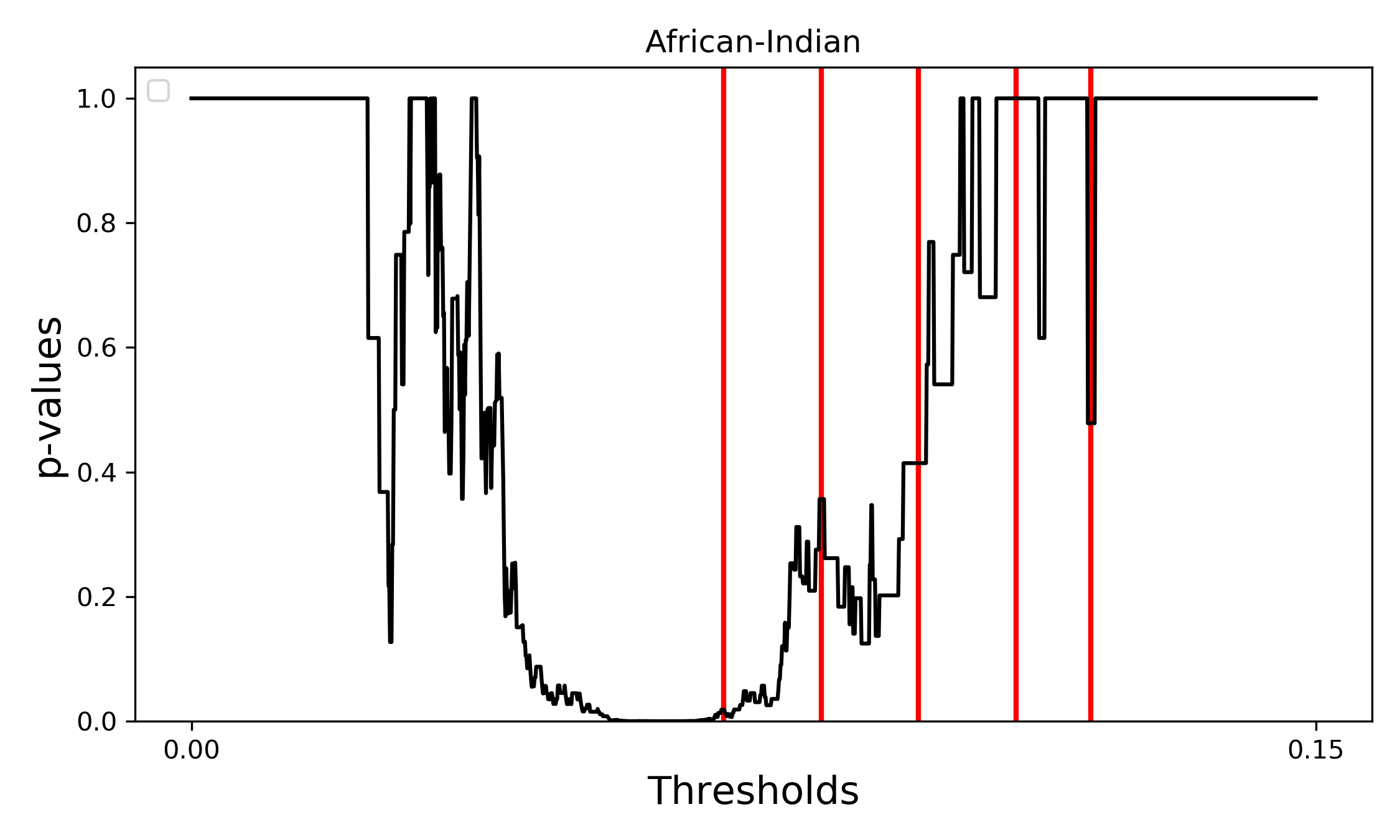}\hfill
\includegraphics[width=0.33\textwidth]{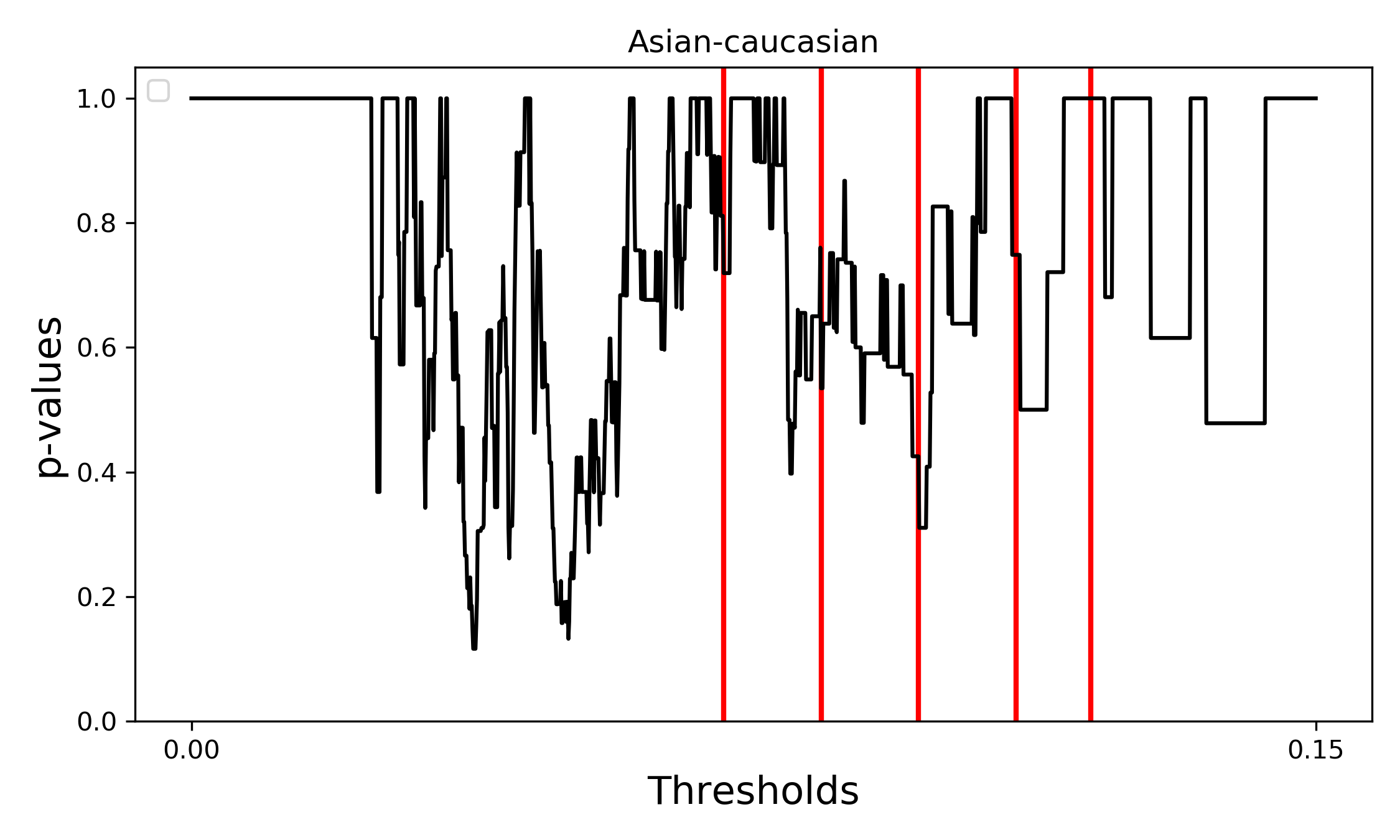}\hfill 
\includegraphics[width=0.33\textwidth]{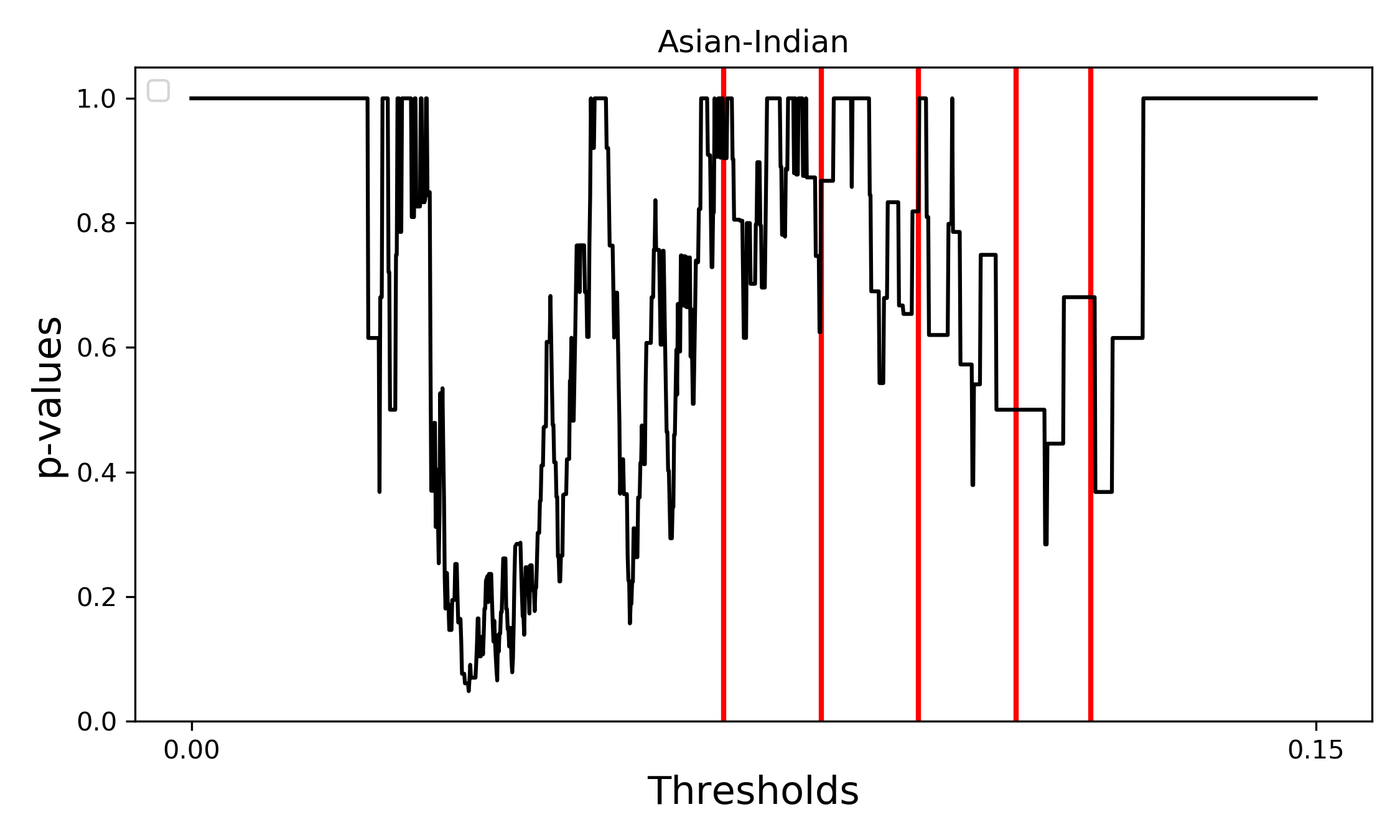}\hfill
\includegraphics[width=0.33\textwidth]{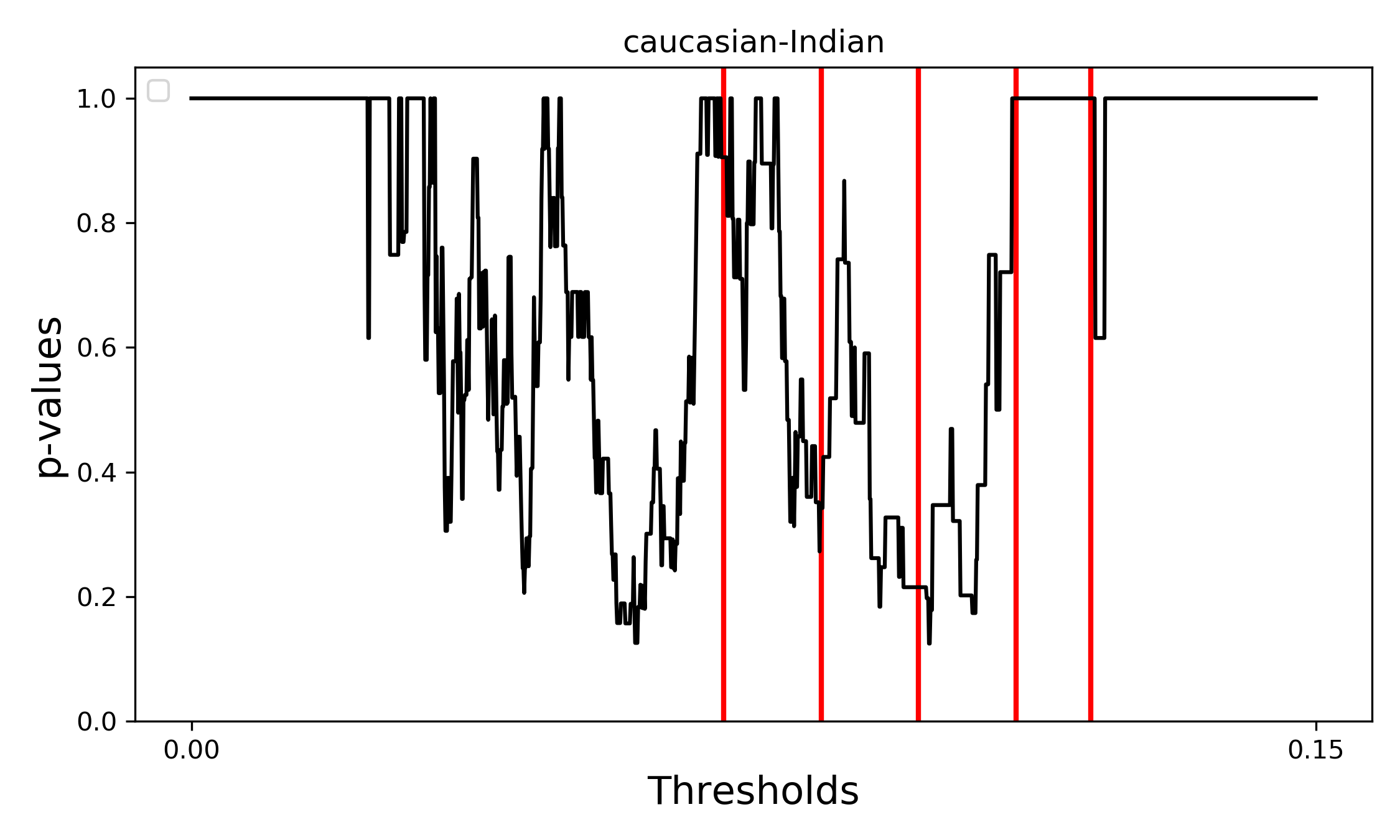} \hfill 
\caption{For each pair of races, graphs of the p-value as a function of the threshold. The classifier was trained on SiW.}
\label{fig:rfwBinarySIW}
\end{figure*}

Table~\ref{tbl:tTest-RfW} shows the results of the chi-squared tests on the binary outcomes. We note that bias can be detected for some of the thresholds, for some pairs of races. In Figs.~\ref{fig:rfwBinaryRA},~\ref{fig:rfwBinarySIW}, for each classifier, and for each pair of races, we show the histograms of responses on bona fide images, and the p-values of the chi-squared test as a function of the threshold. We observe behaviours similar to those from the tests on the SiW database. 

\begin{table}[h]
\caption{p-values of the chi-squared tests for the thresholds shown in Table~\ref{tbl:thresholdsRfW}.}
\centering
\begin{tabular}{|c||c|c|c|c|c|c|} \hline 
            &  Af-As & Af-Ca & Af-In & As-Ca & As-In & Ca-In \\ \hline\hline 
RAtr 1\% & 1.0 & .0718 & .2171 & .1316 & .3680 & 1.0 \\ \hline
RAtr 2\% & 1.0 & .1271 & .2839 & .0741 & .1774 & 1.0 \\ \hline 
RAtr 5\% & .8494 & .0299 & .1082 & .0116 & .0483 & .7487 \\ \hline 
RAtr 10\% & .0356 & .0069 & .3938 & .0000 & .0021 & .0857 \\ \hline 
RAtr 20\% & .1333 & .0025 & 1.0 & .0000 & .1047 & .0037 \\ \hline \hline 
SiWtr 1\% & 1.0 & 1.0 & 1.0 & 1.0 & .3680 & .6153 \\ \hline 
SiWtr 2\% & .2839 & .4456& 1.0 &1.0& .5001 & .7209 \\ \hline 
SiWtr 5\% & .2924 & .0249 & .4143& .31053 &1.0& .2152 \\ \hline 
SiWtr 10\% & .1158 & .0302 & .3567 & .6501 & .6246 & .2725 \\ \hline 
SiWtr 20\% & .0236& .0062 & .01241 & .7194 & 0.9039 & .9054 \\ \hline 
\end{tabular}
\label{tbl:tTest-RfW}
\end{table}


\subsection{Statistical analysis of the scalar responses}

Table~\ref{tbl:tTest-RFW} shows the p-values of the Mann–Whitney U test for each pair of races, and Table~\ref{tbl:meansVar-responses-RFW} shows the means, standard deviations and dip values for each population. 

\clearpage

\begin{figure*}[h]
\includegraphics[width=0.33\textwidth]{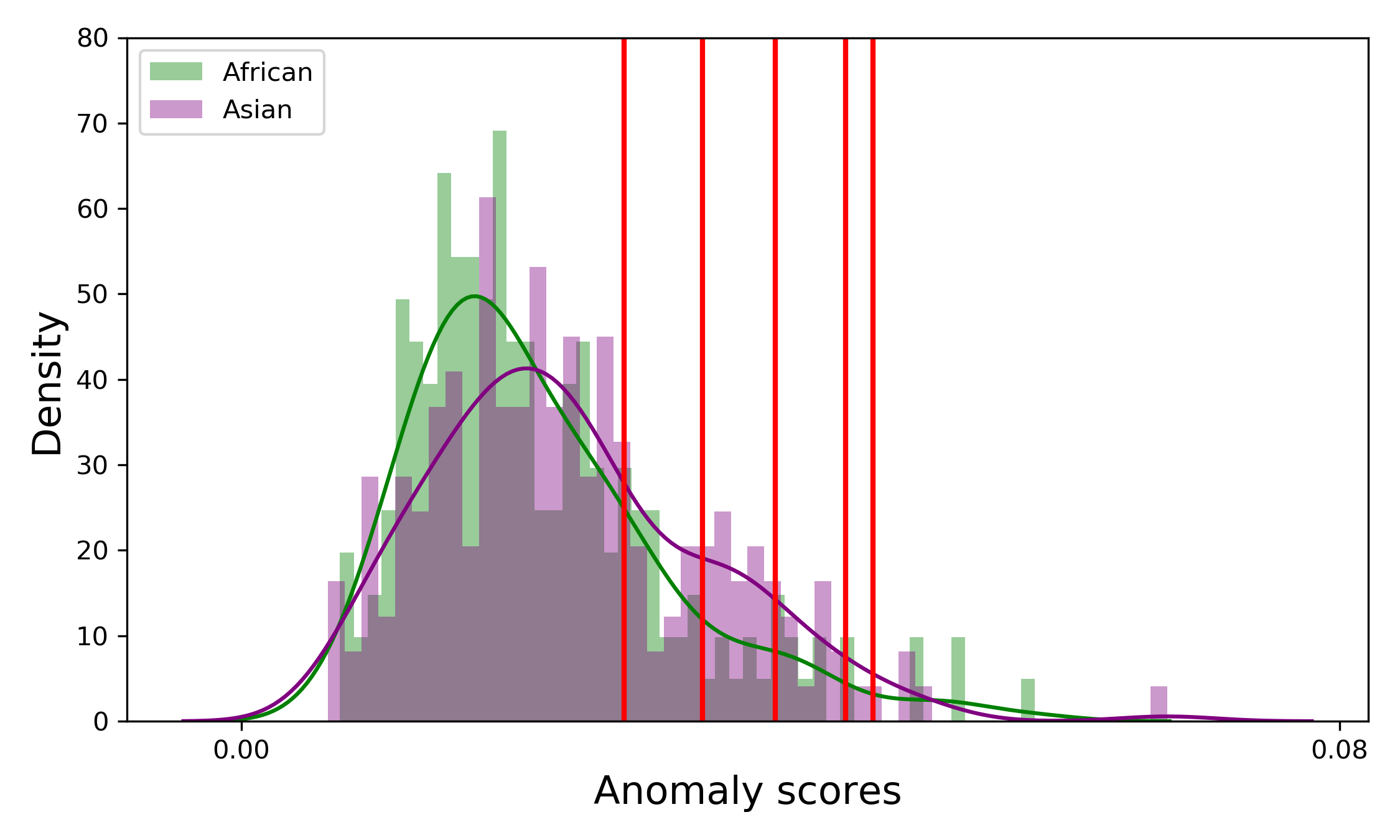} \hfill
\includegraphics[width=0.33\textwidth]{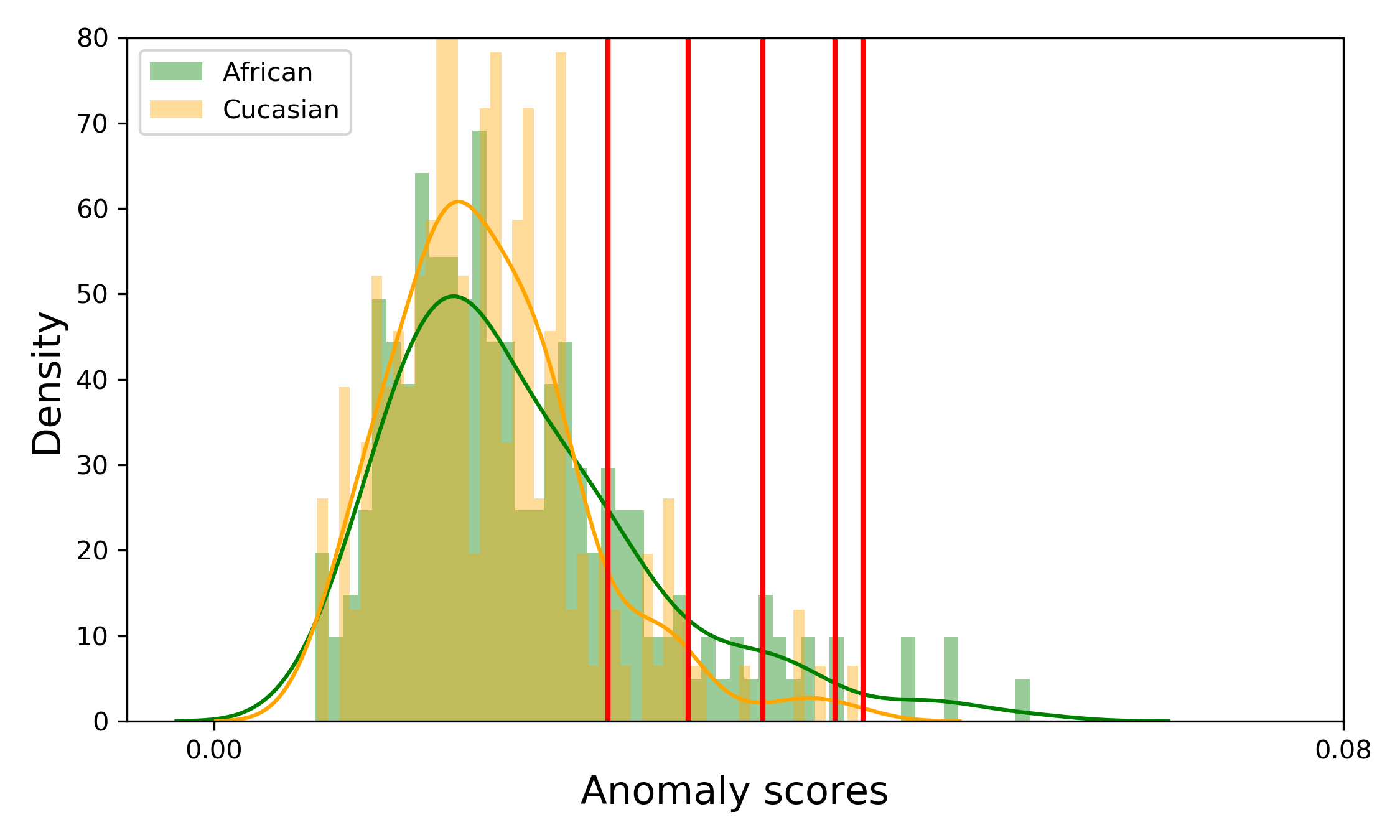}\hfill
\includegraphics[width=0.33\textwidth]{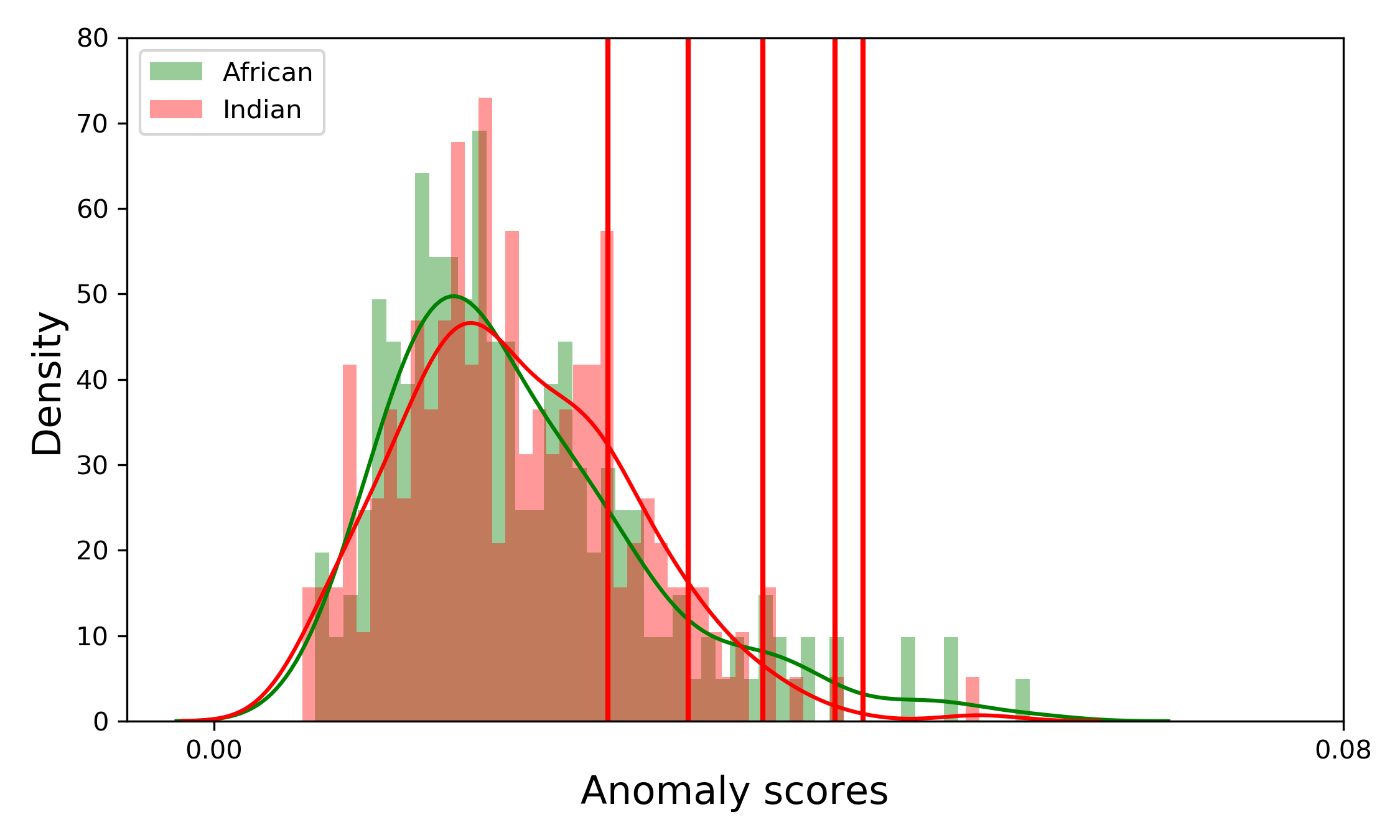}
\includegraphics[width=0.33\textwidth]{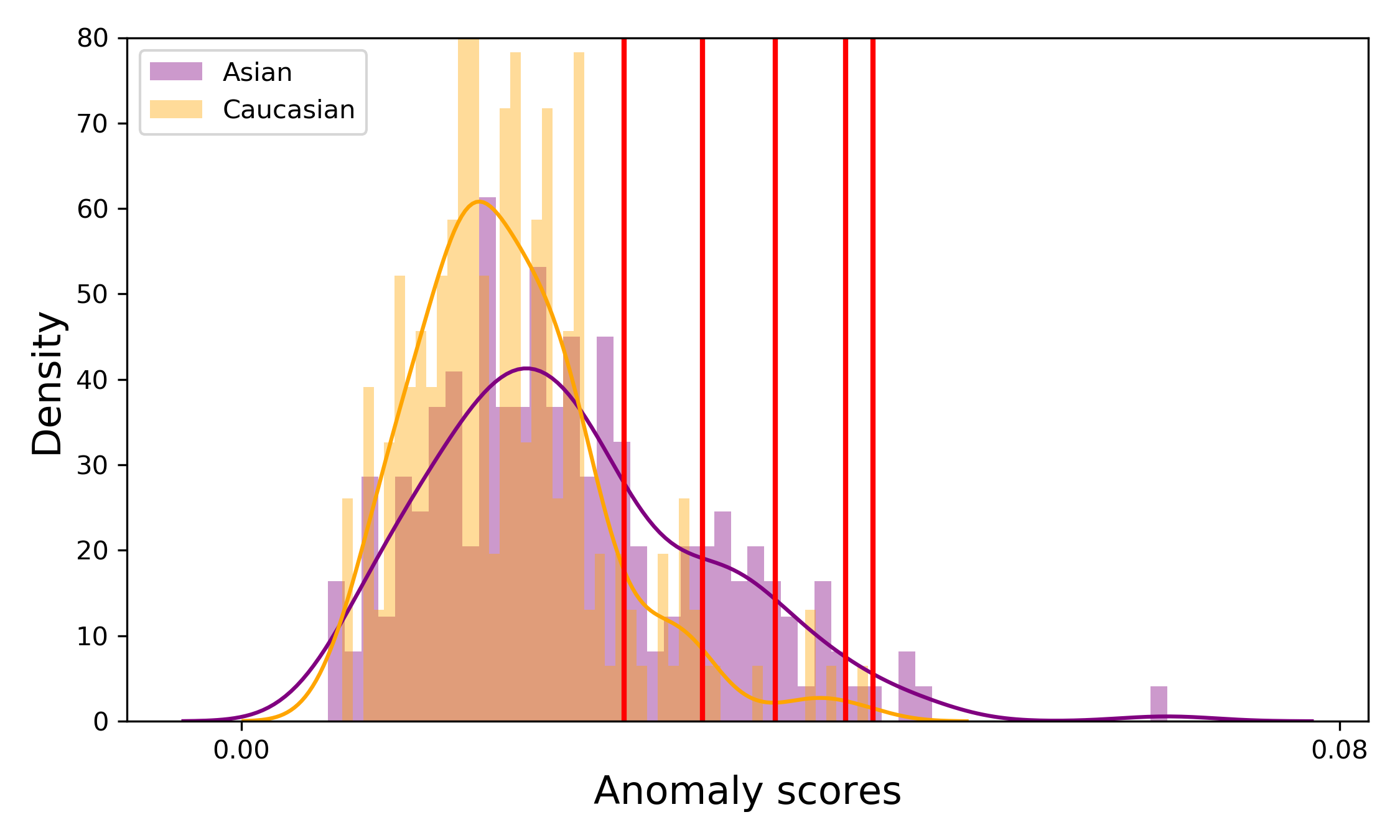} \hfill
\includegraphics[width=0.33\textwidth]{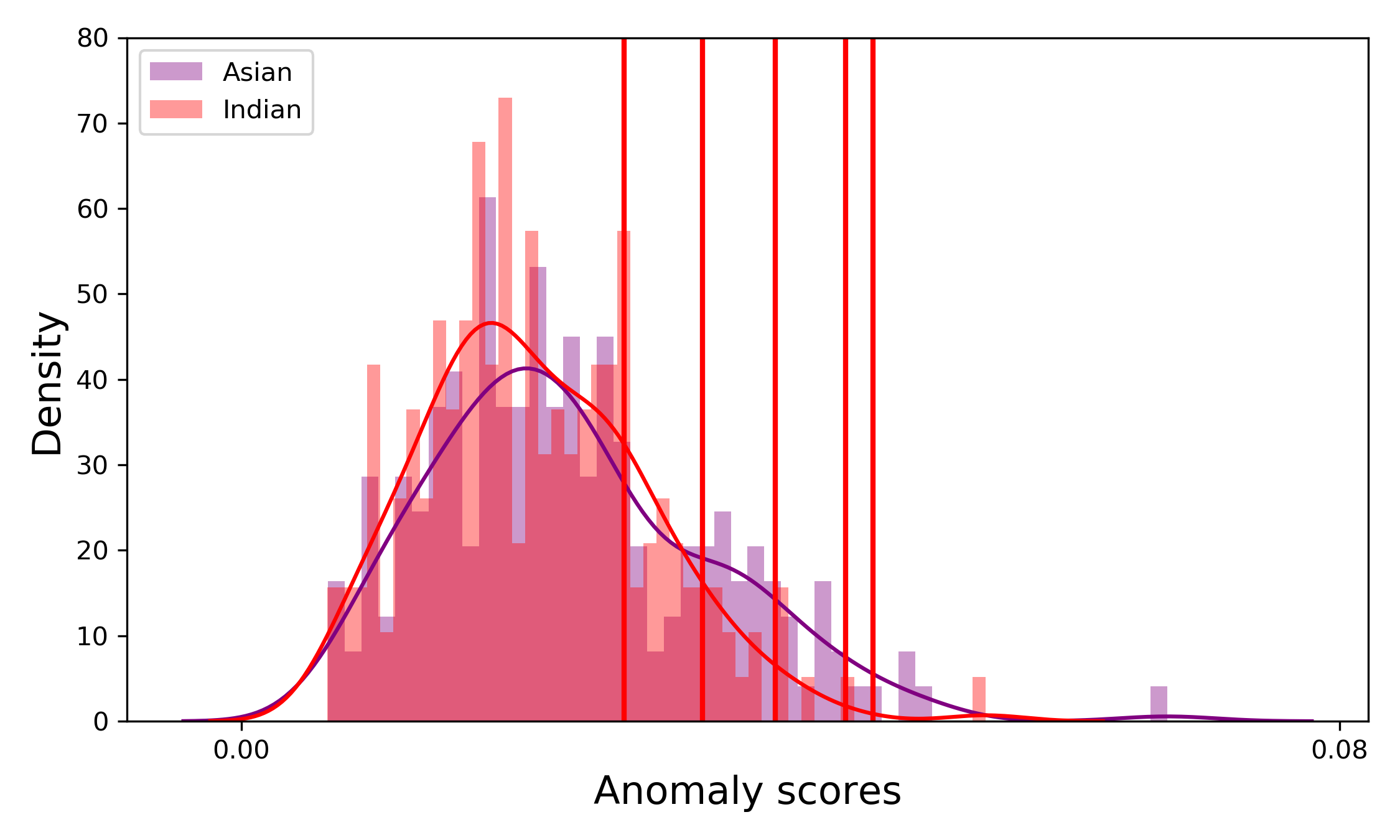}\hfill
\includegraphics[width=0.33\textwidth]{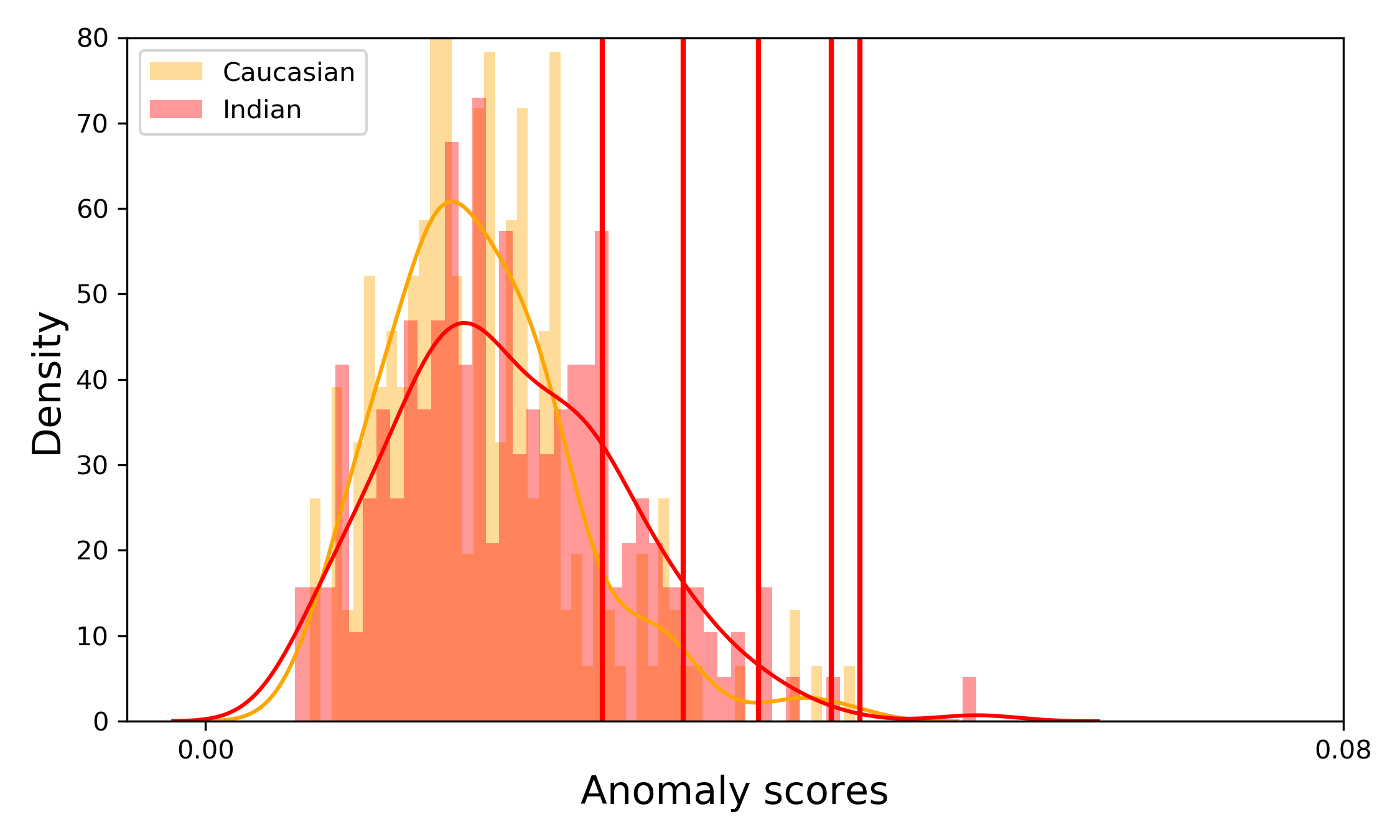}
\caption{For each pair of races in RFW, the histogram of the responses of bona-fide images. The classifier was trained on RA.}
\label{fig:rfwHistogramsRA}
\end{figure*}

\begin{figure*}[h]
\includegraphics[width=0.33\textwidth]{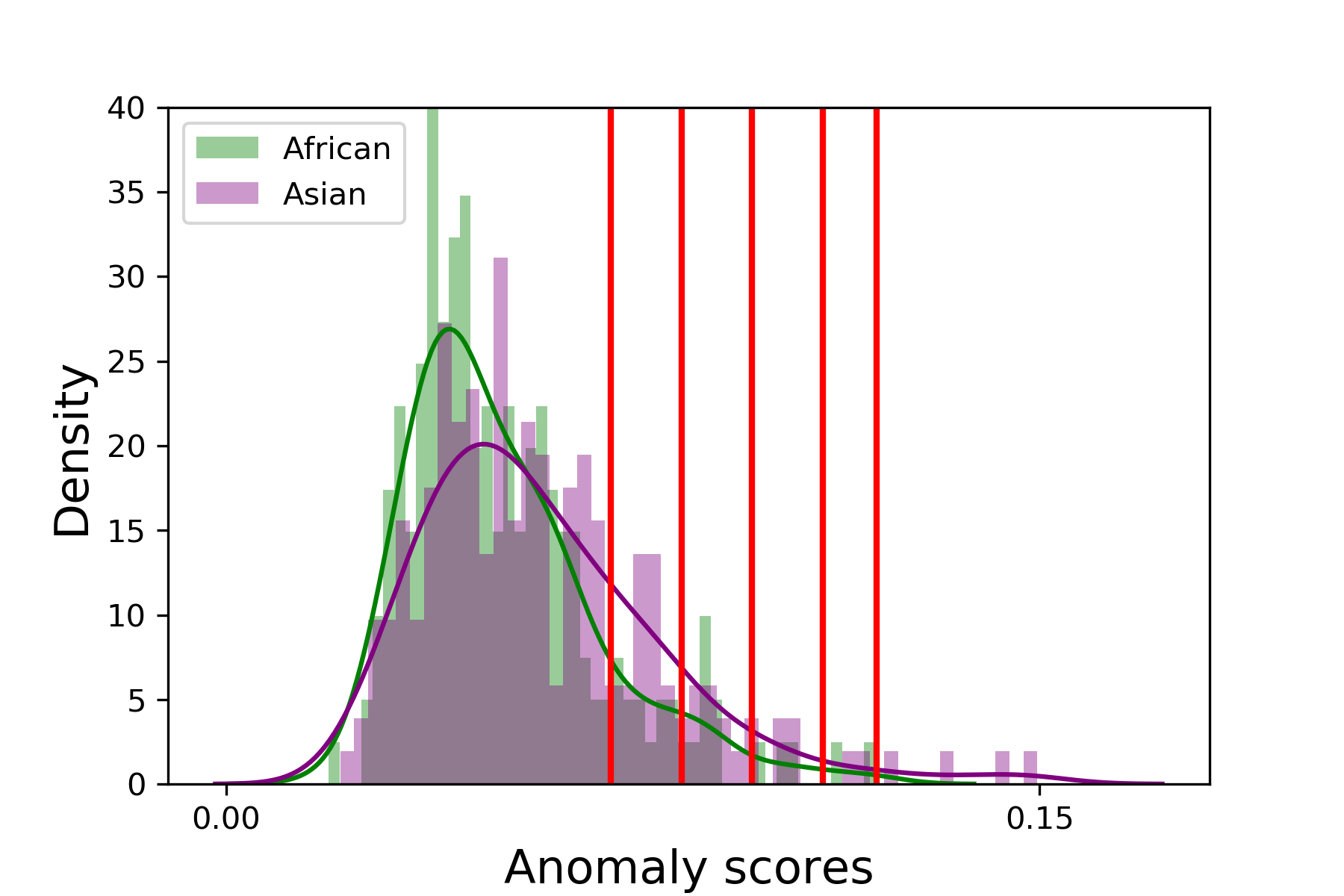} \hfill
\includegraphics[width=0.33\textwidth]{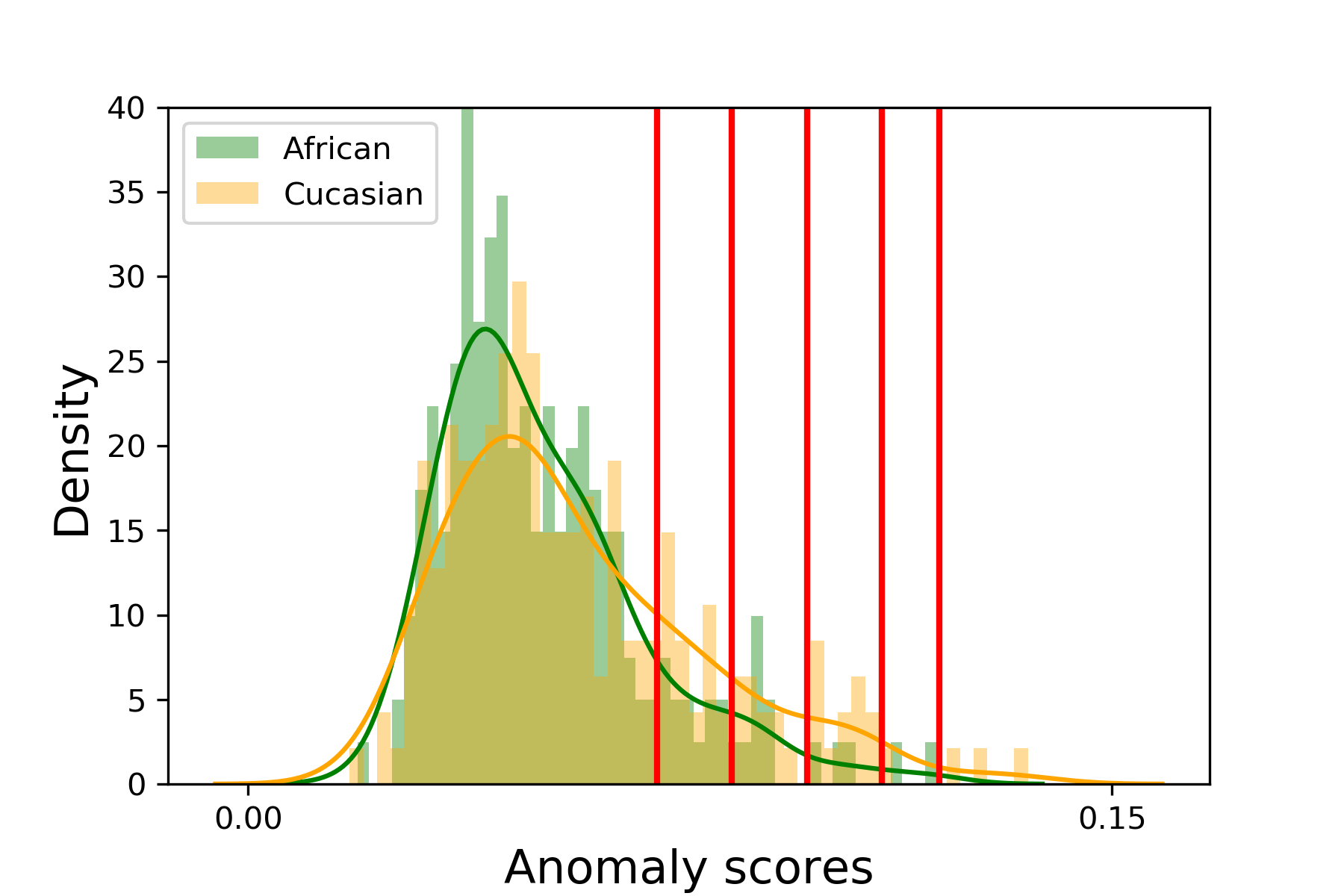}\hfill
\includegraphics[width=0.33\textwidth]{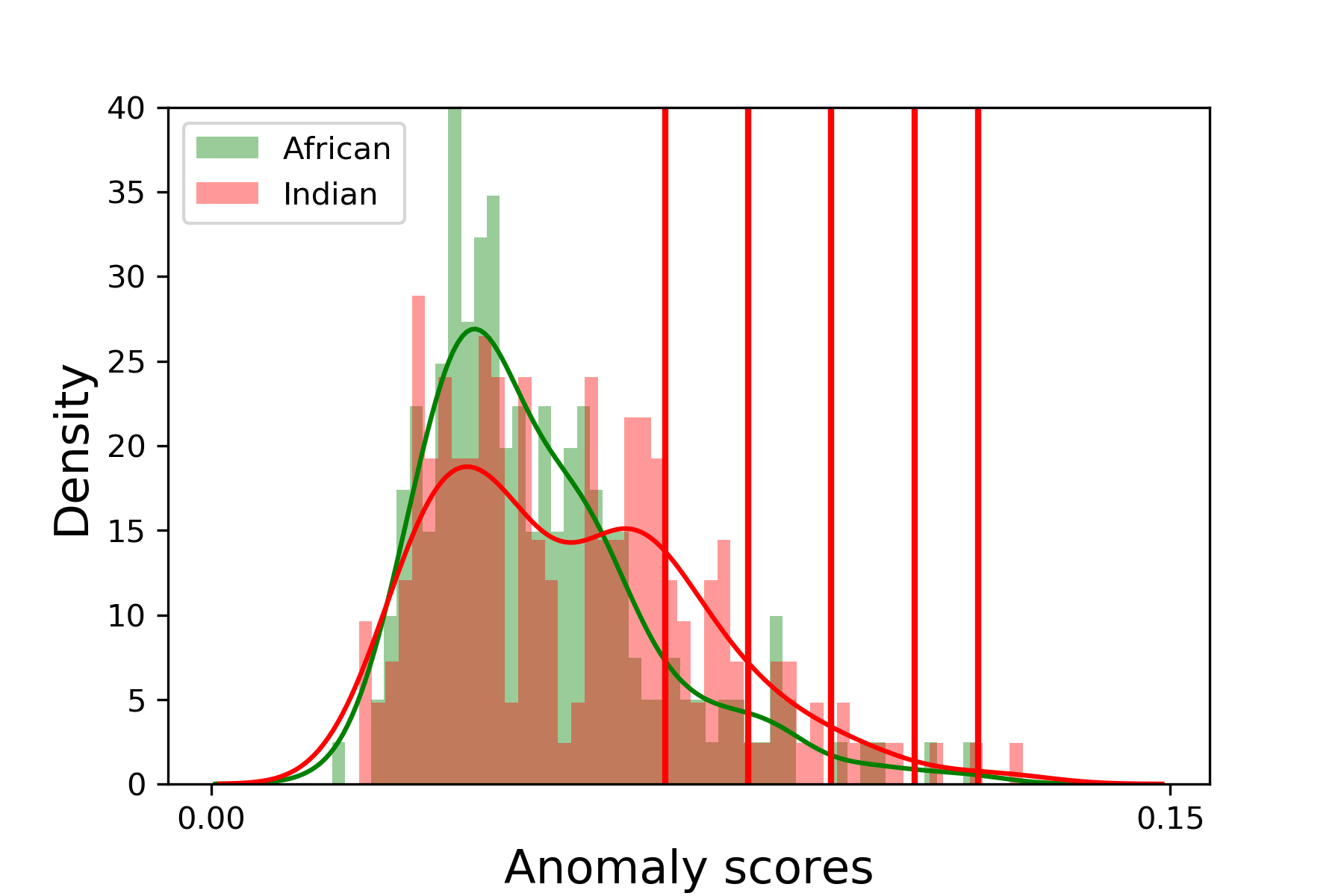}
\includegraphics[width=0.33\textwidth]{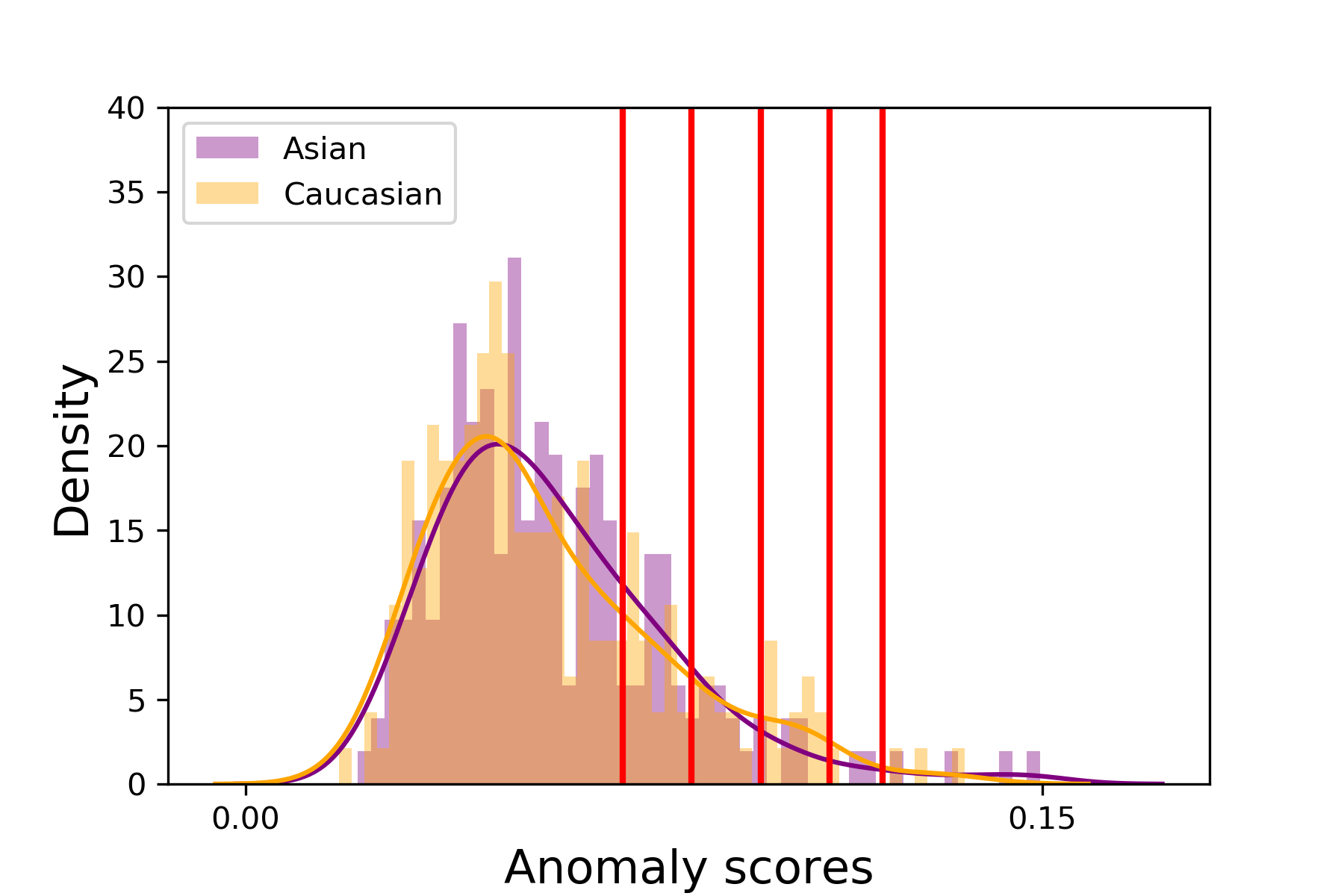} \hfill
\includegraphics[width=0.33\textwidth]{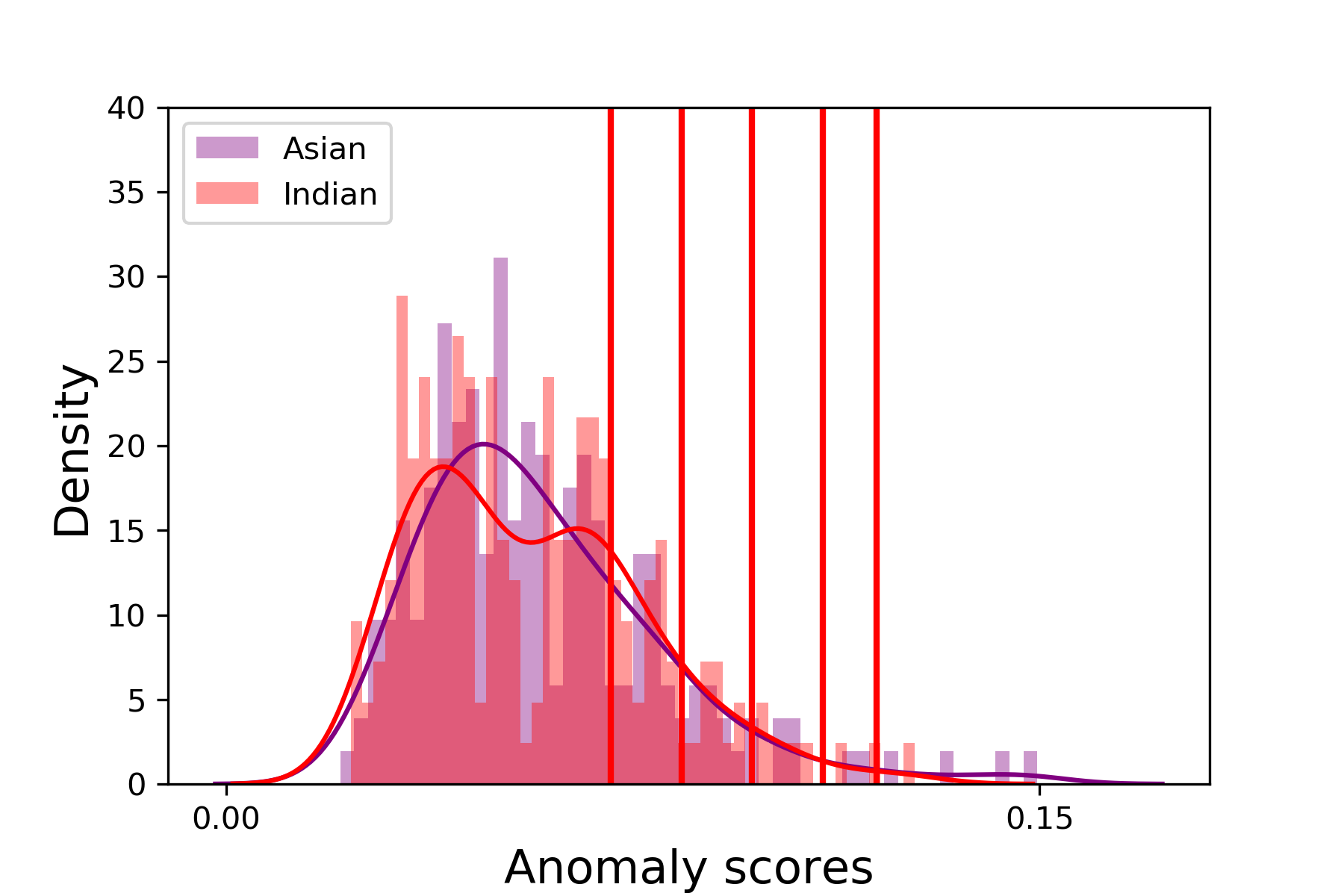}\hfill
\includegraphics[width=0.33\textwidth]{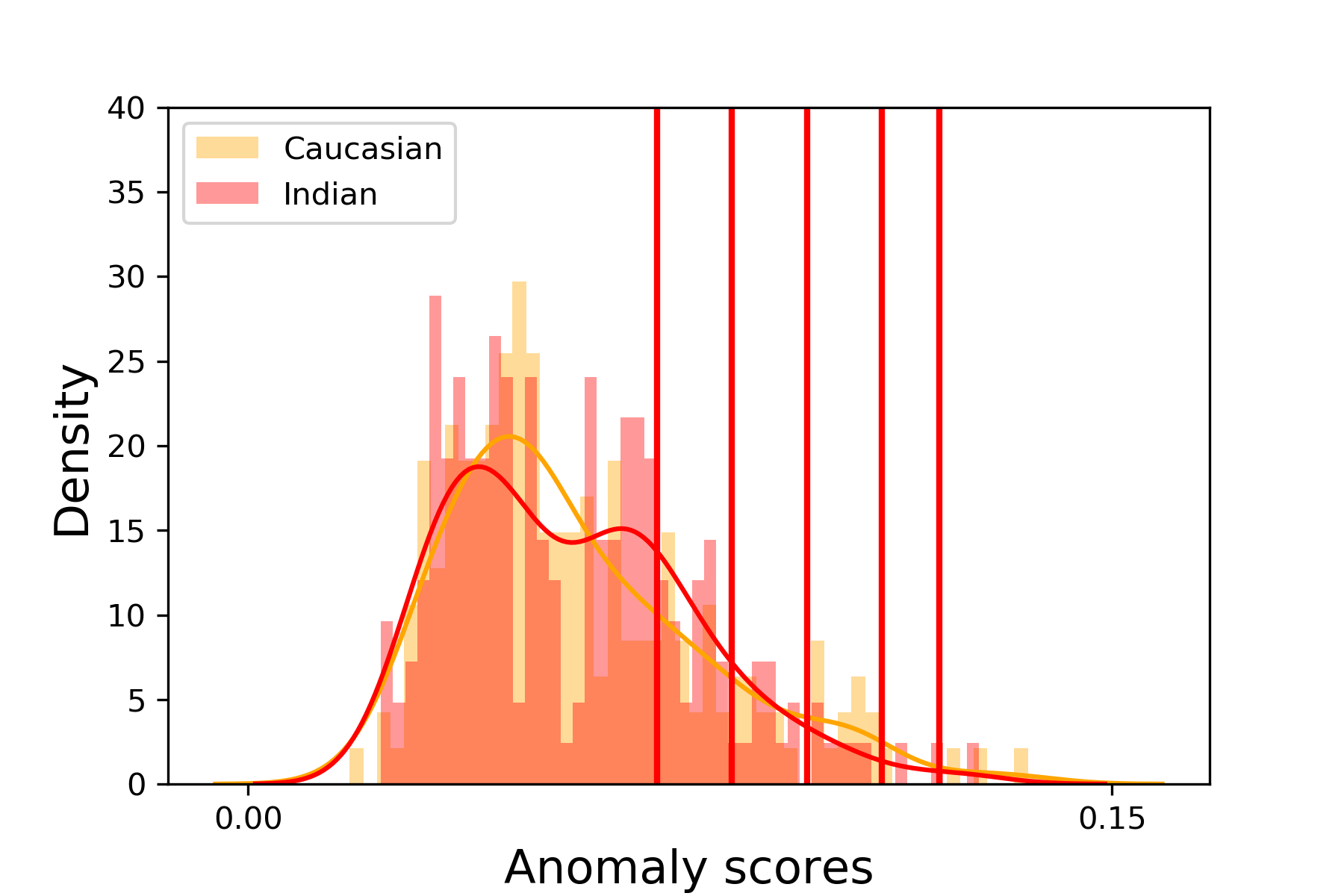}
\caption{For each pair of races in RFW, the histogram of the responses of bona-fide images. The classifier was trained on SiW.}
\label{fig:rfwHistogramsSIW}
\end{figure*}

We note that Hartigan's Dip Test detects a bimodality in the responses of the SiW trained algorithm on Indian people, having a dip value of 0.055, above the significance threshold of 0.037. This can also be verified by visual inspection of the corresponding histograms of responses, which, for each pair of races, are shown in Figs.~\ref{fig:rfwHistogramsRA}, \ref{fig:rfwHistogramsSIW}. We also note that this bimodality can be detected in the behaviour of corresponding graphs of the p-values of the chi-squared test. Indeed, in the three graphs in Fig.~\ref{fig:rfwBinarySIW} corresponding to Indian people, we can detect two distinct regions of higher bias, even though the second one does not reach the level of statistical significance.

\begin{table}[h]
\caption{p-values of the Mann–Whitney U test on each pair of races.}
\centering
\begin{tabular}{|c||c|c|c|c|c|c|} \hline 
            &  Af-As & Af-Ca & Af-In & As-Ca & As-In & Ca-In \\ \hline\hline
RAtr  & .0067 & .0253 & .3216 &  .0000 & .0137  & .0044 \\ \hline 
SiWtr & .0004 & .0058  & .0062 & .2509  & .2743 & .4805 \\ \hline 
\end{tabular}
\label{tbl:tTest-RFW}
\end{table}

\begin{table}[h]
\caption{RFW testset: means, standard deviations, and Hartigan's dip values for the responses of the RA and SiW trained classifiers.}
\centering
\begin{tabular}{|c|c|c|c|c|} \hline 
 & \textbf{Af} & \textbf{As} & \textbf{Ca} & \textbf{In} \\ \hline \hline
RAtr $\mu$ & .0217 & .0239 & .0194 & .0214 \\ \hline 
SiWtr $\mu$ & .0509 & .0579 & .0569 & .0579 \\ \hline \hline
RAtr $s.d.$ & .0094 & .0101 & .0068 & .0082 \\ \hline 
SiWtr $s.d.$ & .0175 & .0220 & .0223 & .0220\\ \hline 
RAtr dip & .0175 & .0178& .0299 & .0216 \\ \hline \hline 
SIWtr dip & .0114 & .0225 & .0149& .0550 \\ \hline 
\end{tabular}
\label{tbl:meansVar-responses-RFW}
\end{table}


\subsection{Discrete latent space}

Table~\ref{tbl:QV-SVM-RFW} shows the AUC values of SVM classifiers trained and tested, on each pair of races, with the Quantised Vectors. We note an even better performance than in the case of the SiW testset, implying an even higher potential for bias. 
\begin{table}[h]
\caption{AUC values for an SVM trained on the VQ-VAE's latent space encodings of the images. RFW testset.}
\centering
\begin{tabular}{|c||c|c|c|c|c|c|} \hline 
            &  Af-As & Af-Ca & Af-In & As-Ca & As-In & Ca-In \\ \hline\hline 
RAtr  &  .95 & .97 & .94 & .89 & .89 & .86 \\ \hline 
SiWtr &  .96 & .98  & .94 & .89  & .86 & .85 \\ \hline 
\end{tabular}
\label{tbl:QV-SVM-RFW}
\end{table}

%% file: sec6.tex
\section{Conclusion}
\label{sec:sec6} 

We conducted a systematic, largely empirical, study of race bias in face anti-spoofing, based on a series of statistical and algorithmic tests. Our premises were that we are interested in the bona fide error; that binary outcomes, scalar responses, and latent space should all be analysed for bias; and that the threshold determining the classifier's operating point should not be assumed fixed by a black-box procedure. 

Our main finding is that the behaviour of race bias depends on several characteristics of the distributions of the corresponding classifier's responses: differences in the means; differences in the variance; bimodal responses; outliers. The main implication of this finding is that the behaviour of bias can be quite complex and, for example, methods optimising for low race bias should not assume unique solutions to the problem. More generally, in our context, bias should be treated as something more complex than the difference in the means of two populations, a misconception that might be reinforced by the fact that in statistics, colloquially, the term bias is often used to describe the component of the error attributed to the difference in the means. 

Our recommendations for future work include the theoretical study of bias, especially under the reasonable assumption of log-normality in the distributions of the responses. We also propose conducting an empirical but more systematic study of the relation between the behaviour of bias detecting algorithmic tests on the latent space, and statistically significant bias in the binary outcomes of the classifier.

%% file: ms.bbl
\begin{thebibliography}{10}

\bibitem{idrd2022}
{ID R\&D whitepaper: Mitigating Demographic Bias in Facial Presentation Attack
  Detection}, 2022.

\bibitem{van2017neural}
Aaron Van Den~Oord, Oriol Vinyals, et~al.
\newblock Neural discrete representation learning.
\newblock {\em Advances in Neural Information Processing Systems}, 30, 2017.

\bibitem{chingovska2012}
Ivana Chingovska, Andr{\'e} Anjos, and S{\'e}bastien Marcel.
\newblock On the effectiveness of local binary patterns in face anti-spoofing.
\newblock In {\em The International Conference of Biometrics Special Interest
  Group (BIOSIG)}, pages 1--7. IEEE, 2012.

\bibitem{boulkenafet2015face}
Zinelabidine Boulkenafet, Jukka Komulainen, and Abdenour Hadid.
\newblock Face anti-spoofing based on color texture analysis.
\newblock In {\em International Conference on Image Processing (ICIP)}, pages
  2636--2640. IEEE, 2015.

\bibitem{albiol2008face}
Alberto Albiol, David Monzo, Antoine Martin, Jorge Sastre, and Antonio Albiol.
\newblock Face recognition using hog--ebgm.
\newblock {\em Pattern Recognition Letters}, 29(10):1537--1543, 2008.

\bibitem{jourabloo2018face}
Amin Jourabloo, Yaojie Liu, and Xiaoming Liu.
\newblock Face de-spoofing: Anti-spoofing via noise modeling.
\newblock In {\em Proceedings of the European Conference on Computer Vision
  (ECCV)}, pages 290--306, 2018.

\bibitem{nagpal2019performance}
Chaitanya Nagpal and Shiv~Ram Dubey.
\newblock A performance evaluation of convolutional neural networks for face
  anti spoofing.
\newblock In {\em International Joint Conference on Neural Networks (IJCNN)},
  pages 1--8. IEEE, 2019.

\bibitem{cai2020drl}
Rizhao Cai, Haoliang Li, Shiqi Wang, Changsheng Chen, and Alex~C Kot.
\newblock Drl-fas: A novel framework based on deep reinforcement learning for
  face anti-spoofing.
\newblock {\em IEEE Transactions on Information Forensics and Security},
  16:937--951, 2020.

\bibitem{zhang2020celeba}
Yuanhan Zhang, ZhenFei Yin, Yidong Li, Guojun Yin, Junjie Yan, Jing Shao, and
  Ziwei Liu.
\newblock Celeba-spoof: Large-scale face anti-spoofing dataset with rich
  annotations.
\newblock In {\em European Conference on Computer Vision}, pages 70--85.
  Springer, 2020.

\bibitem{yu2020fas}
Zitong Yu, Jun Wan, Yunxiao Qin, Xiaobai Li, Stan~Z Li, and Guoying Zhao.
\newblock Nas-fas: Static-dynamic central difference network search for face
  anti-spoofing.
\newblock {\em IEEE Transactions on Pattern Analysis and Machine Intelligence},
  43(9):3005--3023, 2020.

\bibitem{Yu_2020_CVPR}
Zitong Yu, Chenxu Zhao, Zezheng Wang, Yunxiao Qin, Zhuo Su, Xiaobai Li, Feng
  Zhou, and Guoying Zhao.
\newblock Searching central difference convolutional networks for face
  anti-spoofing.
\newblock In {\em Proceedings of the IEEE/CVF Conference on Computer Vision and
  Pattern Recognition (CVPR)}, June 2020.

\bibitem{yu2020auto}
Zitong Yu, Yunxiao Qin, Xiaqing Xu, Chenxu Zhao, Zezheng Wang, Zhen Lei, and
  Guoying Zhao.
\newblock Auto-fas: Searching lightweight networks for face anti-spoofing.
\newblock In {\em IEEE International Conference on Acoustics, Speech and Signal
  Processing (ICASSP)}, pages 996--1000. IEEE, 2020.

\bibitem{zhang2020face}
Ke-Yue Zhang, Taiping Yao, Jian Zhang, Ying Tai, Shouhong Ding, Jilin Li,
  Feiyue Huang, Haichuan Song, and Lizhuang Ma.
\newblock Face anti-spoofing via disentangled representation learning.
\newblock In {\em European Conference on Computer Vision}, pages 641--657.
  Springer, 2020.

\bibitem{liu2021face}
Ajian Liu, Zichang Tan, Jun Wan, Yanyan Liang, Zhen Lei, Guodong Guo, and
  Stan~Z Li.
\newblock Face anti-spoofing via adversarial cross-modality translation.
\newblock {\em IEEE Transactions on Information Forensics and Security},
  16:2759--2772, 2021.

\bibitem{yu2021revisiting}
Zitong Yu, Xiaobai Li, Jingang Shi, Zhaoqiang Xia, and Guoying Zhao.
\newblock Revisiting pixel-wise supervision for face anti-spoofing.
\newblock {\em IEEE Transactions on Biometrics, Behavior, and Identity
  Science}, 3(3):285--295, 2021.

\bibitem{wang2022}
Zhuo Wang, Qiangchang Wang, Weihong Deng, and Guodong Guo.
\newblock Face anti-spoofing using transformers with relation-aware mechanism.
\newblock {\em IEEE Transactions on Biometrics, Behavior, and Identity
  Science}, 4(3):439--450, 2022.

\bibitem{atoum2017face}
Yousef Atoum, Yaojie Liu, Amin Jourabloo, and Xiaoming Liu.
\newblock Face anti-spoofing using patch and depth-based cnns.
\newblock In {\em Proc. IJCB}, pages 319--328. IEEE, 2017.

\bibitem{liu2018}
Yaojie Liu*, Amin Jourabloo*, and Xiaoming Liu.
\newblock Learning deep models for face anti-spoofing: Binary or auxiliary
  supervision.
\newblock In {\em In Proceeding of IEEE Computer Vision and Pattern
  Recognition}, Salt Lake City, UT, June 2018.

\bibitem{wang2020deep}
Zezheng Wang, Zitong Yu, Chenxu Zhao, Xiangyu Zhu, Yunxiao Qin, Qiusheng Zhou,
  Feng Zhou, and Zhen Lei.
\newblock Deep spatial gradient and temporal depth learning for face
  anti-spoofing.
\newblock In {\em Proceedings of the IEEE/CVF Conference on Computer Vision and
  Pattern Recognition}, pages 5042--5051, 2020.

\bibitem{wu2022}
Hangtong Wu, Dan Zeng, Yibo Hu, Hailin Shi, and Tao Mei.
\newblock Dual spoof disentanglement generation for face anti-spoofing with
  depth uncertainty learning.
\newblock {\em IEEE Transactions on Circuits and Systems for Video Technology},
  32(7):4626--4638, 2022.

\bibitem{xiong2018unknown}
Fei Xiong and Wael AbdAlmageed.
\newblock Unknown presentation attack detection with face rgb images.
\newblock In {\em Proc. BTAS}, pages 1--9. IEEE, 2018.

\bibitem{jimenezdeep}
David Jim{\'e}nez-Cabello and Daniel P{\'e}rez-Cabo.
\newblock Deep anomaly detection for generalized face anti-spoofing.
\newblock In {\em Actas del IV Machine Learning Workshop}, pages 1--31.
  University of A Coru\~na, 2019.

\bibitem{Nikisins2019}
Olegs Nikisins, Anjith George, and Sébastien Marcel.
\newblock Domain adaptation in multi-channel autoencoder based features for
  robust face anti-spoofing.
\newblock In {\em 2019 International Conference on Biometrics (ICB)}, pages
  1--8, 2019.

\bibitem{Zhang2020CNNBasedAD}
Yuge Zhang, Min Zhao, Longbin Yan, Tiande Gao, and Jie Chen.
\newblock Cnn-based anomaly detection for face presentation attack detection
  with multi-channel images.
\newblock {\em IEEE International Conference on Visual Communications and Image
  Processing (VCIP)}, pages 189--192, 2020.

\bibitem{baweja2020anomaly}
Yashasvi Baweja, Poojan Oza, Pramuditha Perera, and Vishal~M Patel.
\newblock Anomaly detection-based unknown face presentation attack detection.
\newblock In {\em International Joint Conference on Biometrics (IJCB)}, pages
  1--9. IEEE, 2020.

\bibitem{mohammadi2020}
Amir Mohammadi, Sushil Bhattacharjee, and Sébastien Marcel.
\newblock Improving cross-dataset performance of face presentation attack
  detection systems using face recognition datasets.
\newblock In {\em ICASSP 2020 - 2020 IEEE International Conference on
  Acoustics, Speech and Signal Processing (ICASSP)}, pages 2947--2951, 2020.

\bibitem{abduh2021training}
L~Abduh and I~Ivrissimtzis.
\newblock Training dataset construction for anomaly detection in face
  anti-spoofing.
\newblock {\em CGVC}, 2021.

\bibitem{wang2019rfw}
Mei Wang, Weihong Deng, Jiani Hu, Xunqiang Tao, and Yaohai Huang.
\newblock Racial faces in the wild: Reducing racial bias by information
  maximization adaptation network.
\newblock pages 692--702, 10 2019.

\bibitem{tan2010face}
Xiaoyang Tan, Yi~Li, Jun Liu, and Lin Jiang.
\newblock Face liveness detection from a single image with sparse low rank
  bi-linear discriminative model.
\newblock In {\em Proc. ECCV}, pages 504--517. Springer, 2010.

\bibitem{7031384}
Di~Wen, Hu~Han, and Anil~K. Jain.
\newblock Face spoof detection with image distortion analysis.
\newblock {\em IEEE Transactions on Information Forensics and Security},
  10(4):746--761, 2015.

\bibitem{OULU_NPU_2017}
Zinelabinde Boulkenafet, Jukka Komulainen, Lei Li, Xiaoyi Feng, and Abdenour
  Hadid.
\newblock {OULU-NPU}: A mobile face presentation attack database with
  real-world variations.
\newblock In {\em 12th IEEE International Conference on Automatic Face Gesture
  Recognition (FG 2017)}, pages 612--618, 2017.

\bibitem{george2019biometric}
Anjith George, Zohreh Mostaani, David Geissenbuhler, Olegs Nikisins, Andr{\'e}
  Anjos, and S{\'e}bastien Marcel.
\newblock Biometric face presentation attack detection with multi-channel
  convolutional neural network.
\newblock {\em IEEE Transactions on Information Forensics and Security},
  15:42--55, 2019.

\bibitem{zhang2020casia}
Shifeng Zhang, Ajian Liu, Jun Wan, Yanyan Liang, Guodong Guo, Sergio Escalera,
  Hugo~Jair Escalante, and Stan~Z Li.
\newblock Casia-surf: A large-scale multi-modal benchmark for face
  anti-spoofing.
\newblock {\em IEEE Transactions on Biometrics, Behavior, and Identity
  Science}, 2(2):182--193, 2020.

\bibitem{Liu_2021_WACV}
Ajian Liu, Zichang Tan, Jun Wan, Sergio Escalera, Guodong Guo, and Stan~Z. Li.
\newblock Casia-surf cefa: A benchmark for multi-modal cross-ethnicity face
  anti-spoofing.
\newblock In {\em Proceedings of the IEEE/CVF Winter Conference on Applications
  of Computer Vision (WACV)}, pages 1179--1187, January 2021.

\bibitem{yapo2018ethical}
Adrienne Yapo and Joseph Weiss.
\newblock Ethical implications of bias in machine learning.
\newblock 2018.

\bibitem{mehrabi2021survey}
Ninareh Mehrabi, Fred Morstatter, Nripsuta Saxena, Kristina Lerman, and Aram
  Galstyan.
\newblock A survey on bias and fairness in machine learning.
\newblock {\em ACM Computing Surveys (CSUR)}, 54(6):1--35, 2021.

\bibitem{suresh2019framework}
Harini Suresh and John~V Guttag.
\newblock A framework for understanding unintended consequences of machine
  learning.
\newblock {\em arXiv preprint arXiv:1901.10002}, 2:8, 2019.

\bibitem{olteanu2019social}
Alexandra Olteanu, Carlos Castillo, Fernando Diaz, and Emre K{\i}c{\i}man.
\newblock Social data: Biases, methodological pitfalls, and ethical boundaries.
\newblock {\em Frontiers in Big Data}, 2:13, 2019.

\bibitem{singh2020robustness}
Richa Singh, Akshay Agarwal, Maneet Singh, Shruti Nagpal, and Mayank Vatsa.
\newblock On the robustness of face recognition algorithms against attacks and
  bias.
\newblock In {\em Proceedings of the AAAI Conference on Artificial
  Intelligence}, volume~34, pages 13583--13589, 2020.

\bibitem{phillips2011}
P.~Jonathon Phillips, Fang Jiang, Abhijit Narvekar, Julianne Ayyad, and
  Alice~J. O'Toole.
\newblock An other-race effect for face recognition algorithms.
\newblock {\em ACM Trans. Appl. Percept.}, 8(2), feb 2011.

\bibitem{garcia2019harms}
Raul~Vicente Garcia, Lukasz Wandzik, Louisa Grabner, and Joerg Krueger.
\newblock The harms of demographic bias in deep face recognition research.
\newblock In {\em 2019 International Conference on Biometrics (ICB)}, pages
  1--6. IEEE, 2019.

\bibitem{gluge2020not}
Stefan Gl{\"u}ge, Mohammadreza Amirian, Dandolo Flumini, and Thilo Stadelmann.
\newblock How (not) to measure bias in face recognition networks.
\newblock In {\em IAPR Workshop on Artificial Neural Networks in Pattern
  Recognition}, pages 125--137. Springer, 2020.

\bibitem{serna2021insidebias}
Ignacio Serna, Alejandro Pe{\~n}a, Aythami Morales, and Julian Fierrez.
\newblock Insidebias: Measuring bias in deep networks and application to face
  gender biometrics.
\newblock In {\em 2020 25th International Conference on Pattern Recognition
  (ICPR)}, pages 3720--3727. IEEE, 2021.

\bibitem{cavazos2020accuracy}
Jacqueline~G Cavazos, P~Jonathon Phillips, Carlos~D Castillo, and Alice~J
  O’Toole.
\newblock Accuracy comparison across face recognition algorithms: Where are we
  on measuring race bias?
\newblock {\em IEEE Transactions on Biometrics, Behavior, and Identity
  Science}, 2020.

\bibitem{pereira2021}
Tiago de~Freitas~Pereira and S{\'e}bastien Marcel.
\newblock Fairness in biometrics: a figure of merit to assess biometric
  verification systems.
\newblock {\em IEEE Transactions on Biometrics, Behavior, and Identity
  Science}, 4(1):19--29, 2021.

\bibitem{liu2020crossethnicity}
Ajian Liu, Xuan Li, Jun Wan, Sergio Escalera, Hugo~Jair Escalante, Meysam
  Madadi, Yi~Jin, Zhuoyuan Wu, Xiaogang Yu, Zichang Tan, Qi~Yuan, Ruikun Yang,
  Benjia Zhou, Guodong Guo, and Stan~Z. Li.
\newblock Cross-ethnicity face anti-spoofing recognition challenge: A review,
  2020.

\bibitem{alshareef2021study}
Norah Alshareef, Xiaohong Yuan, Kaushik Roy, and Mustafa Atay.
\newblock A study of gender bias in face presentation attack and its
  mitigation.
\newblock {\em Future Internet}, 13(9):234, 2021.

\bibitem{amini2019uncovering}
Alexander Amini, Ava~P Soleimany, Wilko Schwarting, Sangeeta~N Bhatia, and
  Daniela Rus.
\newblock Uncovering and mitigating algorithmic bias through learned latent
  structure.
\newblock In {\em Proceedings of the 2019 AAAI/ACM Conference on AI, Ethics,
  and Society}, pages 289--295, 2019.

\bibitem{fang2021demographic}
Meiling Fang, Naser Damer, Florian Kirchbuchner, and Arjan Kuijper.
\newblock Demographic bias in presentation attack detection of iris recognition
  systems.
\newblock In {\em 2020 28th European Signal Processing Conference (EUSIPCO)},
  pages 835--839. IEEE, 2021.

\bibitem{zhang2016joint}
Kaipeng Zhang, Zhanpeng Zhang, Zhifeng Li, and Yu~Qiao.
\newblock Joint face detection and alignment using multitask cascaded
  convolutional networks.
\newblock {\em IEEE Signal Processing Letters}, 23(10):1499--1503, 2016.

\bibitem{10.1214/aoms/1177730491}
H.~B. Mann and D.~R. Whitney.
\newblock {On a Test of Whether one of Two Random Variables is Stochastically
  Larger than the Other}.
\newblock {\em The Annals of Mathematical Statistics}, 18(1):50 -- 60, 1947.

\bibitem{hartigan1985dip}
John~A Hartigan and Pamela~M Hartigan.
\newblock The dip test of unimodality.
\newblock {\em The annals of Statistics}, pages 70--84, 1985.

\end{thebibliography}
